\documentclass[10pt,twocolumn,letterpaper]{article}
\usepackage{iccv}

\def\paperID{6731} 
\def\confName{ICCV}
\def\confYear{2025}

\usepackage[dvipsnames]{xcolor}

\usepackage{multirow}
\usepackage{tabularx}
\usepackage{adjustbox}
\usepackage{bbm}

\usepackage{pifont}
\newcommand{\xmark}{\ding{55}}
\definecolor{vlightgray}{gray}{0.87}
\newcommand{\myxmark}{{\color{vlightgray}\xmark{}}}

\definecolor{tblblue}{RGB}{31, 119, 180}
\definecolor{tblorange}{RGB}{255, 127, 14}
\definecolor{tblgreen}{RGB}{44, 160, 44}
\definecolor{tblred}{RGB}{214, 39, 40}
\newcommand{\blue}[1]{\textcolor{tblblue}{#1}}
\newcommand{\orange}[1]{\textcolor{orange}{#1}}
\newcommand{\green}[1]{\textcolor{tblgreen}{#1}}
\newcommand{\purple}[1]{\textcolor{violet}{#1}}
\newcommand{\pink}[1]{\textcolor{pink}{#1}}

\newcommand{\newcat}[1]{\textcolor{tblgreen}{#1}}
\newcommand{\storage}[1]{\textcolor{tblblue}{#1}}
\newcommand{\tablecat}[1]{\textcolor{blue}{#1}}

\newcommand{\CAnobase}{$\text{CA}_{\text{nb}}$}

\newcommand{\ourtasklong}{static to openable\xspace}
\newcommand{\ourtask}{S2O\xspace}
\newcommand{\papertitle}{S2O: Static to Openable Enhancement for Articulated 3D Objects\xspace}
\newcommand{\pmdataset}{PartNet-Mobility\xspace}
\newcommand{\pmopen}{PM-Openable\xspace}
\newcommand{\pmopenext}{PM-Openable-ext\xspace}
\newcommand{\acddata}{Articulated Containers Dataset\xspace}
\newcommand{\acd}{ACD\xspace}
\newcommand{\pmopenshort}{PM-O\xspace}
\newcommand{\pmopenextshort}{PM-OE\xspace}
\newcommand{\ourdata}{\acddata}
\newcommand{\ourdatashort}{\acd}

\newcommand{\pointgroup}{\textsc{PointGroup}\xspace}
\newcommand{\pg}{\textsc{PG}\xspace}

\newcommand{\pointnext}{\textsc{PointNeXt}\xspace}
\newcommand{\pointnextshort}{\textsc{PX}\xspace}

\newcommand{\maskthreed}{\textsc{Mask3D}\xspace}
\newcommand{\swinthreed}{\textsc{Swin3D}\xspace}

\newcommand{\meshwalker}{\textsc{MeshWalker}\xspace}

\newcommand{\stom}{\textsc{Shape2Motion}\xspace}
\newcommand{\stomacro}{\textsc{S2M}\xspace}
\newcommand{\opdformer}{\textsc{OpdFormer}\xspace}
\newcommand{\unet}{\textsc{U-Net}\xspace}
\newcommand{\adam}{\textsc{Adam}\xspace}
\newcommand{\adamw}{\textsc{AdamW}\xspace}
\newcommand{\opd}{\textsc{OPD}\xspace}
\newcommand{\opdmulti}{\textsc{OPDMulti}\xspace}
\newcommand{\gammamotion}{\textsc{GAMMA}\xspace}
\newcommand{\gammagroup}{\textsc{FPNGroupMot}\xspace}
\newcommand{\ogroup}{\textsc{FPNGroup}\xspace}
\newcommand{\heurmot}{\textsc{HeurMot}\xspace}

\newcommand{\imnet}{\textsc{IM-NET}\xspace}
\newcommand{\tsne}{\textsc{t-SNE}\xspace}
\newcommand{\occost}{\textsc{OC-cost}\xspace}

\newcommand{\drawer}{\texttt{drawer}\xspace}
\newcommand{\door}{\texttt{door}\xspace}
\newcommand{\lidd}{\texttt{lid}\xspace}
\newcommand{\mttrans}{\texttt{prismatic}\xspace}
\newcommand{\mtrot}{\texttt{revolute}\xspace}

\newcommand{\indicator}{\ensuremath{\mathbbm{1}}}

\newcommand{\mypara}[1]{\noindent\textbf{#1}}

\newcolumntype{Y}{>{\centering\arraybackslash}X}
\newcommand\imgclip[2]{\adjincludegraphics[Clip={#1\width} {#1\height} {#1\width} {#1\height}]{#2}}

\definecolor{cvprblue}{rgb}{0.21,0.49,0.74}
\usepackage[pagebackref,breaklinks,colorlinks,citecolor=cvprblue]{hyperref}

\title{\papertitle}
\author{
Denys Iliash$^{1}$\hspace{.4cm}
Hanxiao Jiang$^{2}$\hspace{.4cm}
Yiming Zhang$^{1}$\hspace{.4cm}
Manolis Savva$^{1}$\hspace{.4cm}
Angel X. Chang$^{1, 3}$
\\
$^{1}$Simon Fraser University\hspace{.3cm}
$^{2}$Columbia University\hspace{.3cm}
$^{3}$Canada-CIFAR AI Chair, Amii
\\
\texttt{\href{https://3dlg-hcvc.github.io/s2o/}{https://3dlg-hcvc.github.io/s2o/}}
}
\begin{document}

\twocolumn[{%
\maketitle
\captionsetup{type=figure}
\includegraphics[width=\textwidth]{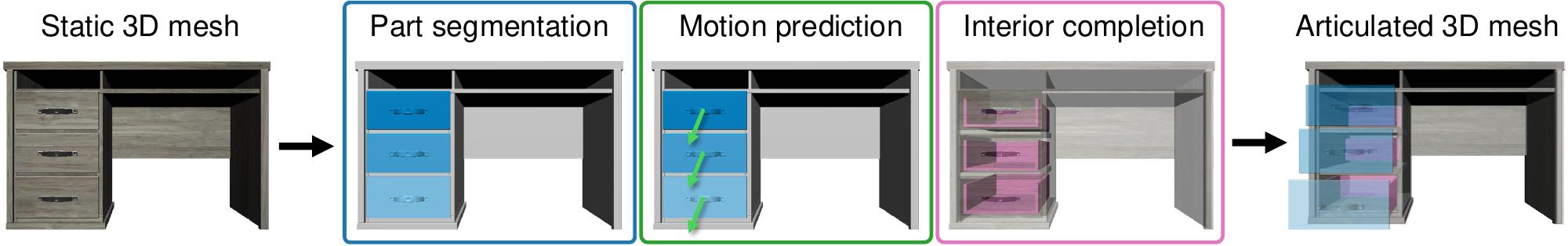}
\captionof{figure}{We introduce the \emph{static to openable} (\ourtask) task: enhancing a static 3D mesh into an articulated openable object. Our \ourtask pipeline consists of three stages: openable part segmentation (blue), motion prediction (green), and interior completion (pink). We benchmark several methods implementing this pipeline across multiple modalities: images, point clouds, and meshes. Our experiments demonstrate enhancement of static 3D meshes with plausible openable part motions and highlight open challenges and directions for future work.}
\label{fig:task-overview}
\vspace{1em}

}]

\begin{abstract}
Despite much progress in large 3D datasets there are currently few interactive 3D object datasets, and their scale is limited due to the manual effort required in their construction.
We introduce the static to openable (S2O) task which creates interactive articulated 3D objects from static counterparts through openable part detection, motion prediction, and interior geometry completion.
We formulate a unified framework to tackle this task, and curate a challenging dataset of openable 3D objects that serves as a test bed for systematic evaluation.
Our experiments benchmark methods from prior work, extended and improved methods, and simple yet effective heuristics for the S2O task.
We find that turning static 3D objects into interactively openable counterparts is possible but that all methods struggle to generalize to realistic settings of the task, and we highlight promising future work directions.
Our work enables efficient creation of interactive 3D objects for robotic manipulation and embodied AI tasks.
\end{abstract}  
\section{Introduction}

Leveraging 3D environments to train embodied agents for navigation~\cite{shen2020igibson,szot2021habitat}, and for object manipulation and rearrangement~\cite{xiang2020sapien,batra2020rearrangement,ehsani2021manipulathor,lemke2024spotcompose} is increasingly popular.
Everyday tasks such as opening a drawer remain challenging for embodied agents, partially due to training data unavailability and lack of realism.
This research direction relies on interactive 3D environments.
However, there are few such datasets due to the need for laborious real-world acquisition or manual annotation, limiting scale and diversity.

We focus on the \textit{\ourtasklong} (\ourtask) task: enhancing static 3D object meshes to make their parts openable, thus converting them to articulated 3D objects (e.g., making drawers in \cref{fig:task-overview} openable).
Our motivation is the increasing availability of large-scale static 3D object datasets~\cite{chang2015shapenet,deitke2023objaverse,deitke2024objaverse}.
By automatically enhancing static 3D objects to be openable, we can significantly expand the scale and diversity of available interactive 3D objects.
In turn, we can increase the volume of challenging, realistic data beyond that in current datasets of articulated objects.

\begin{figure}
\centering
\setkeys{Gin}{width=\linewidth}
\begin{tabularx}{\linewidth}{Y Y Y | Y Y Y}
\toprule
\multicolumn{3}{c}{\pmopen} & \multicolumn{3}{c}{\ourdatashort (ours)}\\
\cmidrule(l{0pt}r{2pt}){0-2} \cmidrule(l{2pt}r{2pt}){4-6}
Val &\multicolumn{2}{c}{Train} &\multicolumn{3}{c}{Val}\\
\cmidrule(l{0pt}r{2pt}){0-0} 
\cmidrule(l{2pt}r{2pt}){2-3} \cmidrule(l{2pt}r{2pt}){4-6}

\includegraphics[trim={10 5 10 10},clip]{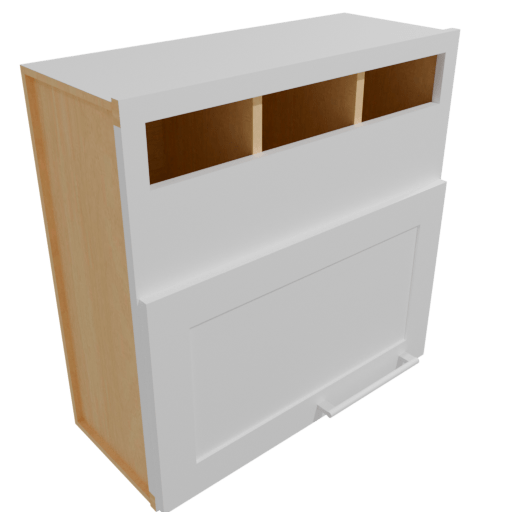} & 
\includegraphics[trim={10 5 10 10},clip]{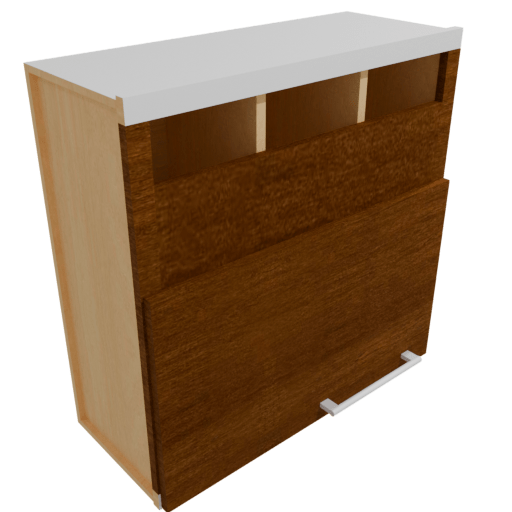} &
\includegraphics[trim={10 5 10 10},clip]{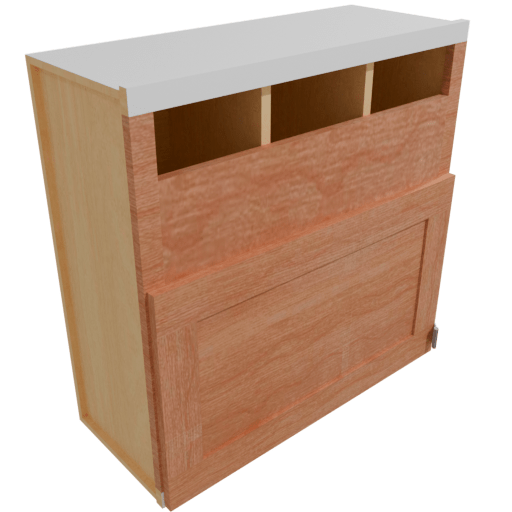} & 

\includegraphics[trim={60 10 60 10},clip]{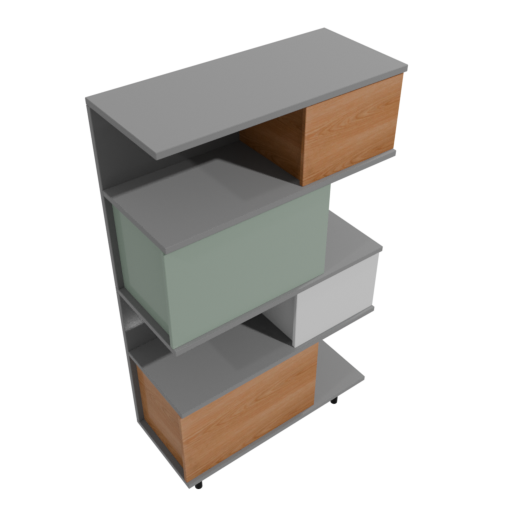} & 
\includegraphics[trim={10 10 20 10},clip]{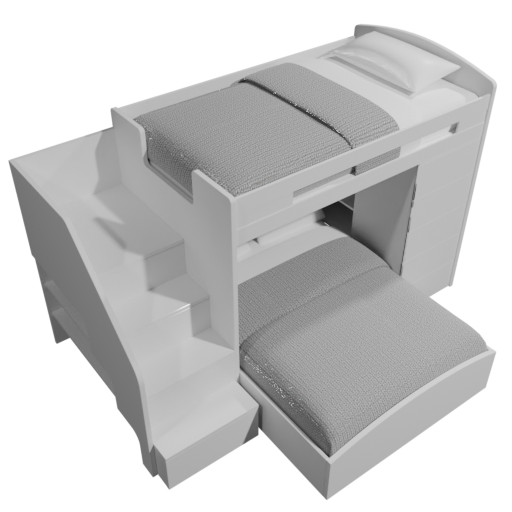} &
\includegraphics[trim={20 20 20 20},clip]{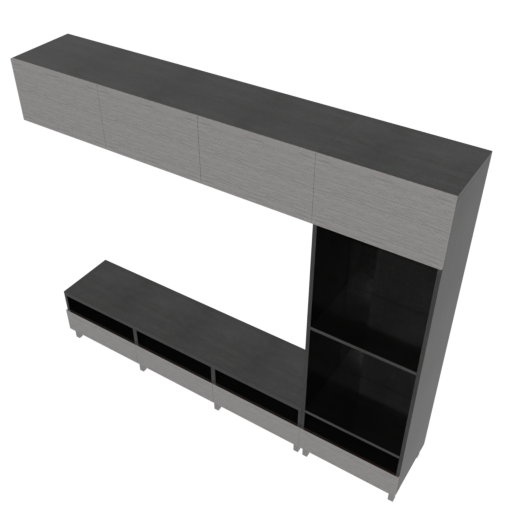}\\

\includegraphics[trim={10 10 10 10},clip]{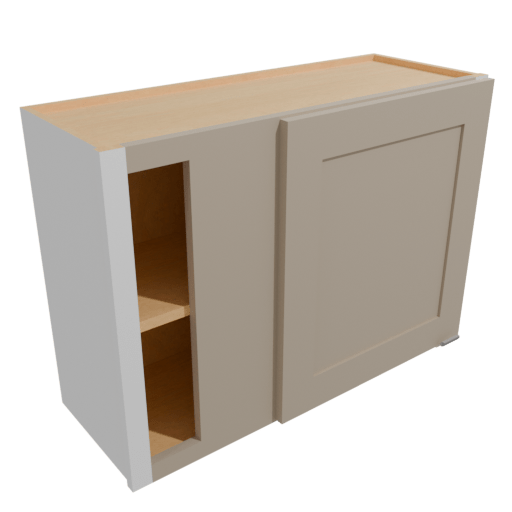} & 
\includegraphics[trim={10 10 20 10},clip]{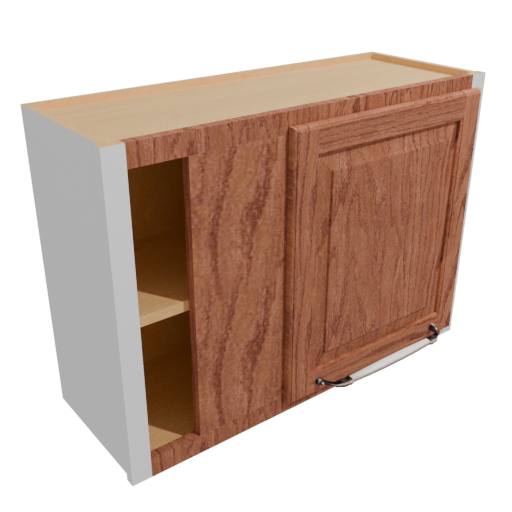} &
\includegraphics[trim={10 10 10 10},clip]{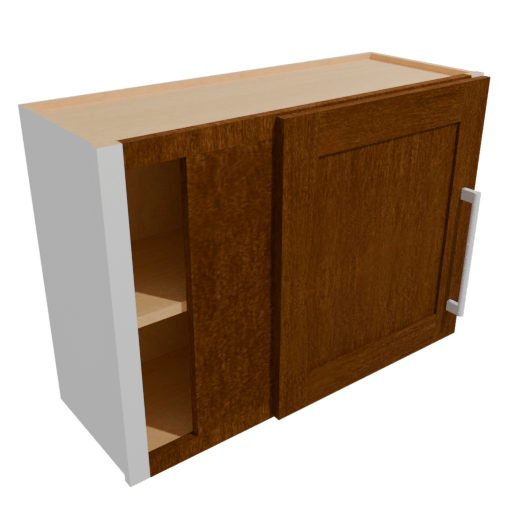} & 

\includegraphics[trim={10 10 10 10},clip]{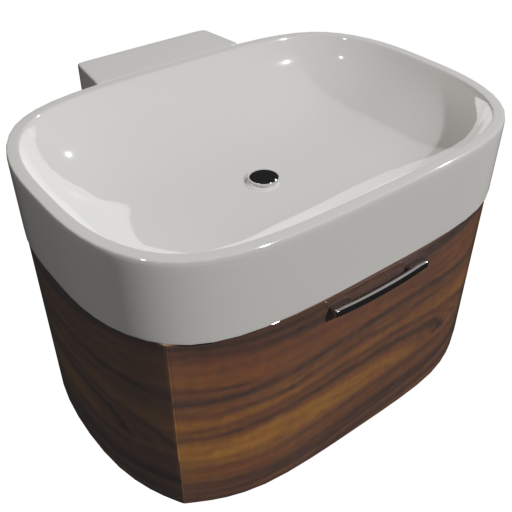} & 
\includegraphics[trim={10 20 10 20},clip]{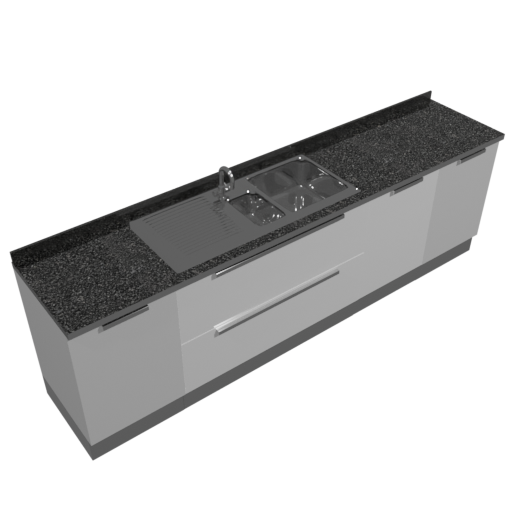} &
\includegraphics[trim={15 10 20 10},clip]{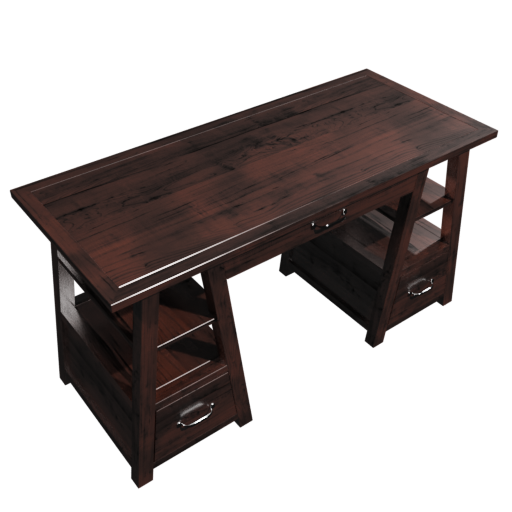}\\

\includegraphics[trim={10 5 10 10},clip]{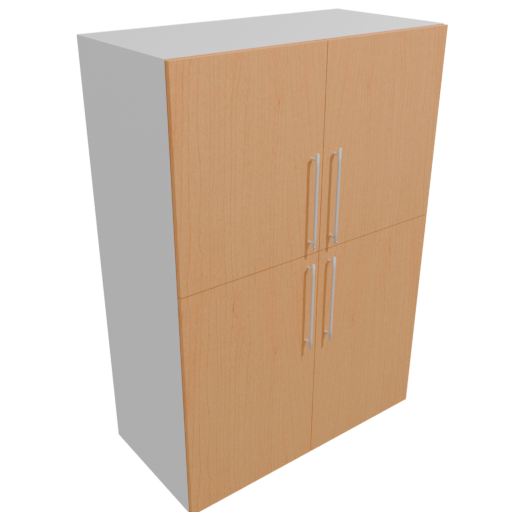} & 
\includegraphics[trim={10 10 10 10},clip]{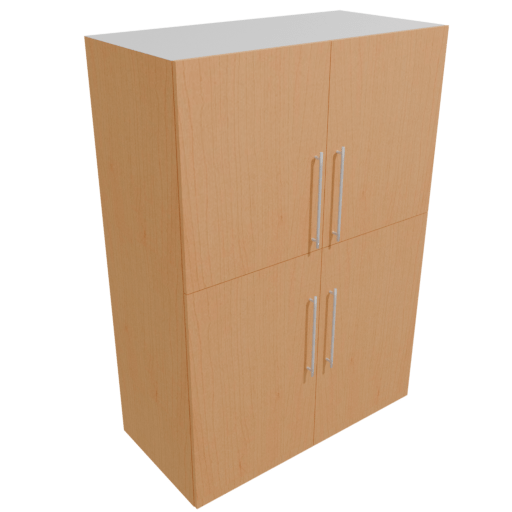} &
\includegraphics[trim={10 10 10 10},clip]{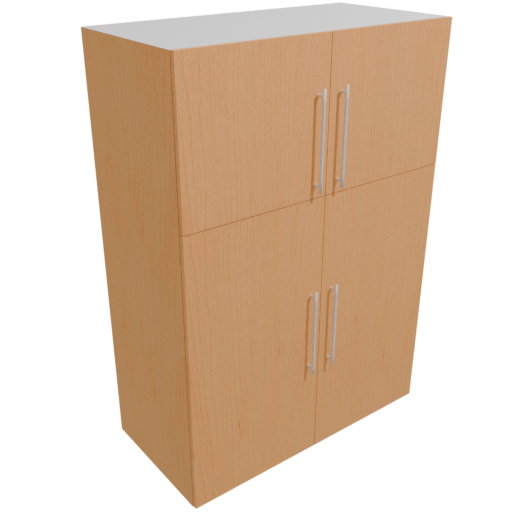} & 

\includegraphics[trim={15 20 15 20},clip]{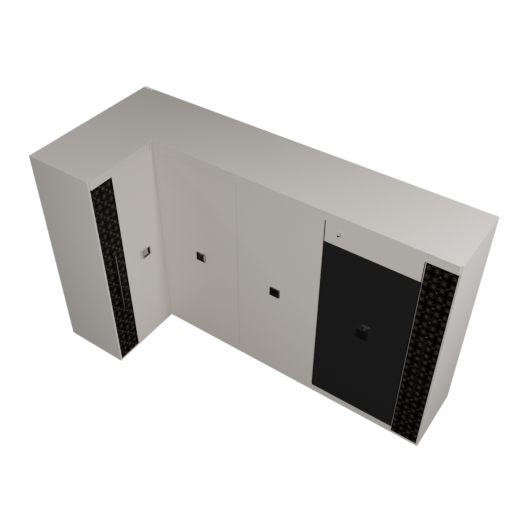} & 
\includegraphics[trim={10 30 10 30},clip]{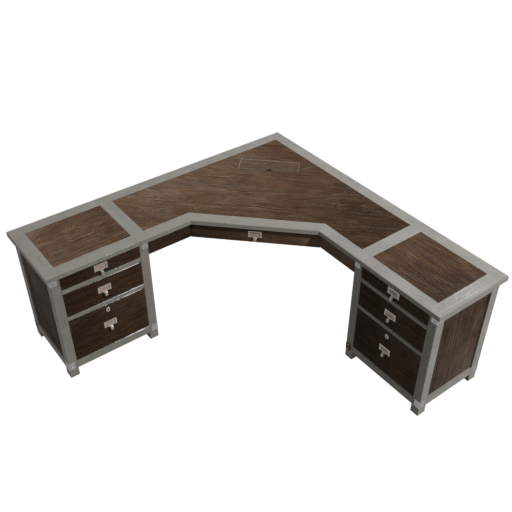} &
\includegraphics[trim={10 20 10 20},clip]{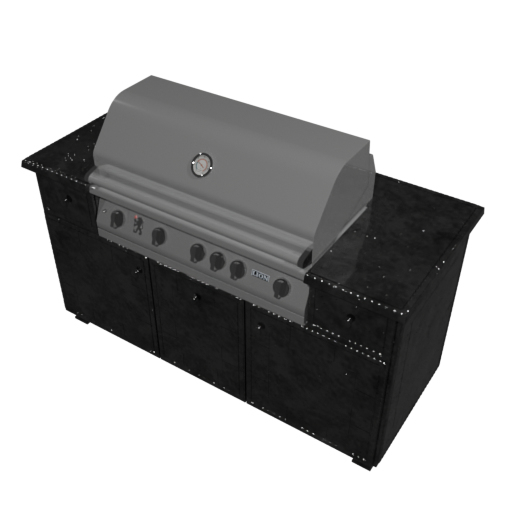}\\

\bottomrule
\end{tabularx}
\vspace{-5pt}
\caption{
Examples of objects from \pmopen (left) and our \ourdata (right) dataset.
\pmopen contains highly similar or identical objects tainting the train-val-test split separation, and is also biased towards objects with easy to annotate motions and complete interior geometry.
In contrast, \ourdata offers a more challenging and realistic dataset with more diverse and complex objects.
}
\label{fig:pm-acd-example-comparison}
\vspace{-5pt}
\end{figure}

\begin{figure}
\centering
\includegraphics[width=0.8\linewidth]{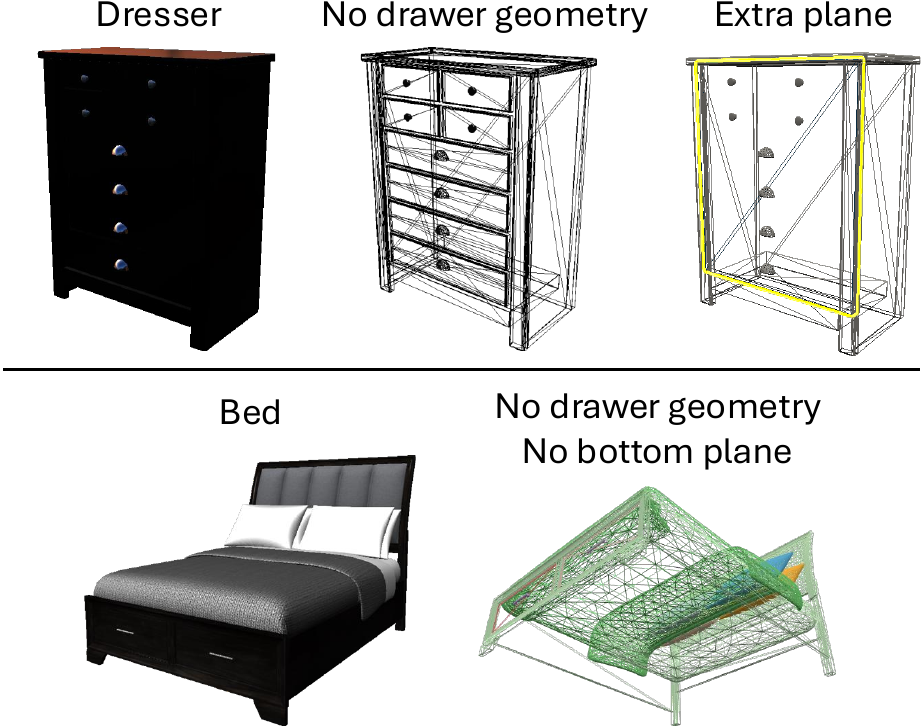}
\caption{
Examples from our \ourdata dataset illustrating challenges in our task.
From left: missing part geometry inside the object (drawer bodies), need to cut geometry when completing interior (flat plane behind drawer front faces), and missing surfaces to support openable parts (bottom of bed).
}
\vspace{-5pt}
\label{fig:task-challenges}
\end{figure}

We specifically target openable containers (cabinets, refrigerators, storage chests, etc.) as they are both prevalent in real-world interiors and widely available as static 3D meshes. 
For example, in the recent HSSD~\cite{khanna2023habitat} dataset, 68\% of approximately 3400 articulated objects are openable containers. The rest are doors, windows, chairs, picture frames, mirrors, lamps, etc, most of which have just one or two articulatable parts.
Compared to objects where the articulation is determined purely by the object category (e.g. scissors, laptop, ceiling fan), containers exhibit complex variations on the number of articulatable parts and how they articulate (e.g., \cref{fig:pm-acd-example-comparison}).
Container objects make up a large part of indoor environments and are also important for many household tasks (e.g., put bowl in cabinet) and having a diverse set of such objects will allow for the exploration of diverse skills and tasks.
While articulating container objects may seem simple, there are in fact many challenges, as \cref{fig:task-challenges} shows.
Segmenting parts is challenging as there is often little geometric detail and few visual cues to separate drawers and doors from the rest of the object.
Furthermore, most static 3D assets are missing interior structures, or have extraneous geometry internally.

To address these challenges, we propose a pipelined approach that tackles three inter-related sub-problems on 3D meshes: openable part segmentation, motion prediction, and interior completion (blue, green, pink boxes in \cref{fig:task-overview}).
Though there has been some work on some of these sub-problems independently, to our knowledge there has been no prior work on the end-to-end \ourtask task or a systematic benchmark of methods to solve the overall task.

In this work, we formalize the \ourtask task and investigate different approaches to the three subtasks.  By analyzing the performance on the subtasks, we determine that part segmentation is the most challenging component.
For part segmentation, we extend prior work on segmenting point clouds and images to obtain a mesh-level segmentation and compare against directly segmenting the 3D mesh.
We establish an evaluation protocol where we train on one set and evaluate on a separate set of diverse shapes.
To do so, we use PartNet-Mobility for training and curate a new Articulated Containers Dataset (ACD) with 354 objects that offer a more challenging testbed for the \ourtask task.

In summary, we:
1) Introduce the \ourtasklong (\ourtask) task to enable creation of interactive 3D objects from static counterparts;
2) Curate the \ourdata (\acd): a challenging dataset enabling experiments to identify open challenges in the \ourtask task;
3) Propose a framework for \ourtask to benchmark part segmentation, motion prediction, and interior completion;
4) Develop the \ogroup, \gammagroup, and \heurmot methods for openable part segmentation and motion prediction; and 
5) Show that part segmentation is a key challenge and that our simple yet effective feature adapter module for segmentation improves F1 score by $33\%$ relative over prior work.

\section{Related work}

We discuss work on the sub-tasks in our problem statement (part segmentation, motion prediction, interior completion), and the broader 3D articulated object generation problem.

\mypara{Part segmentation.}
We focus on \textit{openable} part segmentation which is less studied.
\citet{mao2022multiscan} benchmarked part segmentation for objects in real-world scenes.
Other work adapted low-shot transfer from 2D backbones for 3D point cloud part segmentation~\cite{Liu2022PartSLIPLP, Zhou2023PartSLIPEL} or focused on open-vocabulary manipulable object segmentation~\cite{lemke2024spotcompose}.
Closely related to our work are two methods for openable part detection in images: \opd~\cite{jiang2022opd} and \opdmulti~\cite{sun2023opdmulti}.
Both segment openable parts and predict motion parameters, but neither is focused on producing 3D objects.

\mypara{Motion prediction.}
Part motion prediction has been addressed for single~\cite{wang2019shape2motion,yan2019rpm} or multiple state point clouds~\cite{yi2019deep,huang2021multibodysync}.
Early work trained separate models for each object category and part structure class~\cite{li2020category}.
More recent work has shifted to a category-agnostic regime by focusing on general openable part detection and motion prediction~\cite{jiang2022opd,sun2023opdmulti}, and a wider set of articulated parts~\cite{geng2023gapartnet}.
Other work focuses on mobility prediction from videos, reconstructing a single part that is observed to be moving by fitting a 3D plane~\cite{qian2022understanding}.
\gammamotion~\cite{yu2024gamma} is a recent approach that repurposes the \pointgroup segmentation model to also output motion parameters.
Our work incorporates methods from this literature to tackle the motion prediction sub-problem and to benchmark its impact on our \ourtask task.

\mypara{Interior completion.}
Recent work has looked at completing object interiors given two or more articulation states (open and closed).
\citet{patil2023rosi} take images of different states, and \citet{liu2023paris} take multiview images from two states.
However, assuming multiple input states is not realistic as openable parts and their motions are not typically available for 3D meshes.
Some work tackles geometric completion given partial observation in one state for objects~\cite{dai2017shape,cheng2023sdfusion} and for scenes~\cite{song2017semantic,dai2018scancomplete,wang2023slice3d}, but interior geometry completion for container objects has not been studied in detail despite many issues due to missing interiors (see \cref{fig:task-challenges}).
Very recent work targets interior completion of openable parts~\cite{luo2025physpart} but can suffer from low reconstruction quality, and does not preserve the original mesh geometry, a requirement of our \ourtask task.

Thus, we introduce a simple but effective approach to complete interior geometry.

\mypara{Articulated 3D object generation.}
Early work assumed a static mesh in one state as well as part segmentation were provided as input~\cite{hu2017learning}.
Related work predicts the kinematic chain and motion parameters for a given 3D mesh, again assuming part segmentation as an input~\cite{xu2022unsupervised}.
Other work approaches the problem from a 3D reconstruction perspective, estimating 3D geometry and part motion parameters given RGB-D observations of the objects~\cite{jiang2022ditto,hsu2023ditto,abbatematteo2019learning}.
Other reconstruction methods take multiple images, videos~\cite{jain2020screwnet}, or multiple point clouds~\cite{yi2019deep, huang2021multibodysync}) to identify which parts move and how they move.
A related line of work generates articulated 3D objects, both unconditionally~\cite{lei2023nap} and given input conditions~\cite{liu2023cage}, but both rely on a dataset of annotated articulated objects with geometrically complete interiors.
There is work on interactive interfaces to annotate 3D object datasets with parts and motion parameters~\cite{xu2020motion}, but in general such data is limited in scale.
Other work~\cite{le2024articulate} creates articulated objects from images by retrieving and connecting part meshes, and predicting joint motion parameters.
This work relies on a database of URDF models (e.g., PartNet-Mobility~\cite{xiang2020sapien}) to retrieve parts.
In contrast, our method creates articulated objects from static objects, including segmentation into appropriate parts.
The output of our system can be used as a source of data for such work.

Our work is the first to investigate the complete \ourtasklong task going from a static 3D object mesh to an articulated 3D object, without assuming part segmentations or other annotations in the input.
We systematically benchmark approaches to this challenging problem statement.

\section{Task}

The input to our \ourtasklong (\ourtask) task is a static 3D triangle mesh of a furniture model $M$, and the output is an articulated 3D mesh where each openable part can be interactively moved (see \cref{fig:task-overview}).
The openable parts of the object $P = \{p_1 \ldots p_k\}$ (drawers, doors and lids) are specified through: i) semantic label $l_i$ in \{\drawer, \door, \lidd\} and part mesh segmentation $m_i$ for each part $p_i$; ii) a set of motion parameters $\Phi = \{\phi_1 \ldots \phi_k\}$; and iii) completed part geometries $G = \{g_1 \ldots g_k\}$.
The three sub-problems in our task (part segmentation, motion prediction, interior completion) can be formulated as inferring these three outputs.
We formalize each of these outputs in more detail below.

For the part segmentation outputs, we leverage methods that may operate on images, point clouds, or directly on 3D meshes.
In all three cases, we project predicted semantic labels and segmentation masks onto the original 3D mesh to obtain the $l_i$ labels and segmentations $m_i$.

For the motion prediction outputs, we follow prior work by \citet{jiang2022opd} and define the motion parameters $\phi_i$ of each openable part by specifying the motion type $c_i \in \{\mttrans, \mtrot\}$, motion axis direction $a_i \in R^3$ and motion origin $o_i \in R^3$.
Specifically, we have $\phi_i=[c_i, a_i, o_i]$ for revolute joints (e.g., door rotating around a hinge), and $\phi_i=[c_i, a_i]$ for prismatic joints (e.g., drawer sliding out).

For the interior completion outputs, we produce the geometry $g_i$ by merging the set of triangles corresponding to the segmentation of the part $m_i$ (e.g., drawer front face), with additional geometry representing the interior (e.g., drawer body).
Vertex colors, texture coordinates, and texture maps are similarly extended.

\section{Datasets}

We curate two datasets: \pmopen and \ourdata (\acd).
\pmopen is adapted from \pmdataset ~\cite{xiang2020sapien} objects with openable parts.
\pmdataset is commonly used for tasks related to articulated objects.
\acd is a dataset we introduce with more challenging objects, constructed from objects in ABO~\cite{collins2022abo}, 3D-FUTURE~\cite{fu20213d}, and HSSD~\cite{khanna2023habitat}.
\Cref{fig:pm-acd-example-comparison} shows that \pmopen objects are highly similar across train and val splits while \acd offers more diverse object categories, geometric structures, and part motions (see \cref{tab:data-stats}).

\begin{table}
\centering
\caption{
Summary statistics for the 3D object datasets used in our experiments. The \ourdata dataset is a more challenging test bed with a more diverse set of object categories, and more parts per object on average.}
\vspace{-5pt}
\resizebox{\linewidth}{!}{
\begin{tabular}{@{} l rr rrr rrr @{}}
\toprule
Dataset & obj & cat & part & part/obj & \%drawer & \%door & \%lid \\
\midrule
\pmopen  & 548 & 9 & 1277 & 1.97 & 40.4 & 54.7 & 4.9 \\
\ourdatashort  & 354 & 21 & 1350 & 3.81 & 56.0 & 43.1 & 0.9 \\
\bottomrule
\end{tabular}
}
\vspace{-8pt}
\label{tab:data-stats}
\end{table}

\mypara{\pmopen.} 
The dataset contains 460 training, 95 validation and 93 test objects.
Our experiments report results on the val split from \citet{jiang2022opd} excluding the box and suitcase categories.
We discovered similar or identical objects varying only in textures or minor geometric details, spanning across splits (see \cref{fig:pm-acd-example-comparison}).
To quantify this we measured the L2 distance of \imnet~\cite{chen2019learning} embeddings.
We found 10\% of objects in the train split have a corresponding similar (L2 $\leq$ 0.5) object in the train split, as well as 7.4\% in val and 8.6\% in test split respectively, with corresponding similar objects in the train split.
This suggests that the dataset is self-repetitive and breaks the train vs val/test separation assumption.
Moreover, we noted that this dataset is not representative of commonly used static object datasets such as ABO, 3D-FUTURE, HSSD, where the majority of assets are modelled without interiors.
These findings motivated the curation of \acd.
To investigate the impact of the distribution gap between \pmopen (with interiors) and datasets without interiors we create \textbf{\pmopenext}, a version of \pmopen without interiors (see supplement).

\mypara{\ourdata.}
We create \acd by selecting openable container objects from three commonly used datasets: 3D-FUTURE~\cite{fu20213d} (185), HSSD~\cite{khanna2023habitat} (147), and ABO~\cite{collins2022abo} (22).
\acd consists of 354 objects for which we annotated openable parts and motion parameters using the annotation pipeline from \citet{mao2022multiscan}.
The objects in \acd exhibit more diversity and are more complex than \pmopen, with L-shaped and corner cabinets, beds with storage units etc. (see \cref{fig:pm-acd-example-comparison}).
Overall, the objects have significantly more openable parts and exhibit characteristics common for static 3D mesh objects (missing interior geometry, missing bottom/top surfaces, partially floating parts).
These characteristics create a more \textit{realistic and challenging} test bed for our task, and enable out-of-distribution evaluation on unseen part arrangements.
To investigate training using this dataset, we split \acd into \acd-train (3D-Future assets) and \acd-val (HSSD and ABO assets).
See the supplement for statistics and details.

\section{Approach}

Our framework breaks down the \ourtasklong (\ourtask) task into three stages: 1) identify and segment openable parts; 2) predict motion parameters for openable parts; and 3) complete interior geometry.
For each stage, we compare different families of approaches.

\subsection{Part segmentation}

In this stage, we take a static mesh and produce a segmentation that allows openable parts to be separated and articulated.
Compared to recent work in 3D segmentation from images and point clouds, our problem differs in that we segment \textit{meshes} where: 1) parts are often not spatially separated; and 2) the 3D meshes are non-manifold.
Thus, we need to adapt prior approaches to our scenario.

We consider segmenting from image, point cloud (PC), and mesh.
The output is mesh segmentation, so we need to preprocess the mesh to obtain image and PC data and apply postprocessing to project the predictions back onto the mesh.
In preliminary experiments (see supplement), we find that point cloud methods are the most effective at segmenting the parts, likely due to the focus on point-based method in recent years.
Thus, in the main paper we describe our PC-based method, \ogroup.

\mypara{Point sampling.}
For training we sample 200K points from each part, then we apply farthest point sampling (FPS) to downsample to 20K points total per shape. During inference we sample 1M points total to ensure good coverage and downsample to 20K points to match training. We use k nearest neighbor lookup to propagate to full point clouds.

\mypara{Segmentation.}
We base our segmentation approach (\ogroup) on \pointgroup~\cite{jiang2020pointgroup}, a widely-used PC segmentation approach with good performance.
We introduce two key modifications to the original \pointgroup to enhance its performance on openable-part segmentation: 1) a feature pyramid network (FPN) and 2) a stronger backbone, \pointnext~\cite{qian2022pointnext} and a topology-aware propagation procedure, injecting mesh-based inductive bias.

\begin{figure}
\includegraphics[width=\linewidth]{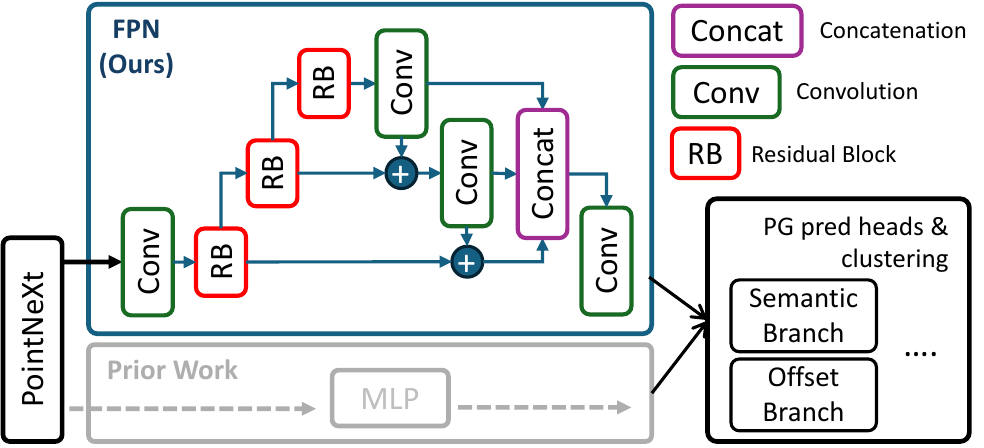}
\caption{
Our \ogroup enhances PointGroup to improve segmentation by 1) introducing an FPN feature adapter block to process point features, 2) using PointNext as the network backbone.
}
\vspace{-8pt}
\label{fig:fpn}
\end{figure}

\pointgroup (\pg) was originally designed for scene segmentation, where objects are spatially separated.
In part segmentation, parts are not necessarily spatially separated and thus more challenging to segment.
We find that \pointgroup's semantic and offset branches, originally simple 2-layer MLPs, are not sufficiently powerful for part segmentation.
Thus we enhance these branches in \ogroup with a feature adapter layer, a Feature Pyramid Network (FPN)~\cite{Lin2016FeaturePN} that adapts backbone features at different resolutions before the semantic and offset branches (see \cref{fig:fpn}). 

We also consider different backbones for \pointgroup.
The original \pg uses a \unet backbone with sparse convolutions. 
We experiment with newer backbones (\pointnext~\cite{qian2022pointnext}, \swinthreed~\cite{yang2023swin3d}), and find that \pointnext performs best. Thus we select it as the basis of our \ogroup.
Finally, we introduce motion prediction heads to \ogroup to create \gammagroup (see \cref{sec:motion-methods}).

\mypara{Point cloud to mesh propagation.}
We design two procedures for propagating predictions from point clouds to meshes. Triangle-based propagation propagates masks using point-to-triangle correspondences while handling overlapping masks. Our second algorithm is topology-aware propagation uses over-segmentations rather than individual triangles. Our initial experiments showed that this procedure is beneficial only for good-performing methods as it can propagate errors in case of bad predictions. Hence we use it only for \ogroup, omitting the baselines.
See the supplement for more details and ablations.

\subsection{Motion prediction}
\label{sec:motion-methods}
In the motion prediction stage, we predict motion parameters (i.e. motion type, motion axis direction, and motion axis origin) for each openable part.
We compare a heuristic method against two learning-based methods on point clouds: 
\stom~\cite{wang2019shape2motion} and our \gammagroup.

\mypara{\heurmot (heuristic).}
Our heuristic method predicts motion parameters based on the \textit{part category}, \textit{3D bounding box}, and the \textit{front and up directions} of the part.
We first select the most frequent motion type (prismatic vs revolute) given the part category based on the statistics from the train set.
For prismatic joints, we directly use the given front direction of the openable part as the motion axis direction.
For revolute joints, we assume that the motion axis lies along one of the edges of the part's 3D bounding box. 
We first determine whether the edge is on the front or back face by checking for alignment with the object base part, and assuming the front is aligned with the front face of the base part.
Typically, the axis line is on the opposite edge of the handle position.
We estimate the handle position using a geometric heuristic that assumes handle geometry is more complex compared to the whole openable part.
See the supplement for more details.

\mypara{\gammagroup (learned).}
\citet{yu2024gamma} introduces \gammamotion which builds upon \pointgroup by adding motion type, motion direction and offset to axis heads.
Following this work, we add motion prediction heads to our \ogroup.
We call our model with motion prediction heads \gammagroup.
For \gammagroup, we use two separate FPNs, one for segmentation and one for motion prediction.
Compared to \gammamotion, we simplify the losses for motion type down to a single cross-entropy loss.
In addition, we find that predicting offsets to origin of the motion axes, rather than to axes themselves, leads to better segmentation performance while maintaining comparable motion prediction.
As predictions are per-point, we aggregate the point predictions using majority voting to obtain the motion type and motion axis per part. For the motion origin, we use the median point after predicted offsets are applied.
Similarly to \ogroup, we use topology-aware propagation and design post-processing heuristic for motion parameters. See supplement for details and ablations. 

\subsection{Interior completion}

\mypara{Heuristic.} For interior completion, we focus on completion of the movable part so that when articulated the part is complete.
For openable objects, the most important part to complete is the drawer for which we use a set of heuristics to build the sides (see \Cref{fig:interior-completion} for results).
We assume that the drawer front is complete and is represented by an oriented bounding box.
We then determine the depth of the drawer by ray tracing from the center and two sides of the drawer inward toward the back of the object until we hit a surface (or reach the bounding box of the object).   
If the distance from the center toward the back is larger than from the sides, then we create a corner drawer.
Otherwise, we build a rectangular box for the bottom and each of the three missing sides of the drawer by extending inward from the front side.
We note that these heuristics are limited, and cannot complete other types of interiors (e.g., dishracks in dishwashers).
In addition, while these heuristics work well for GT segmentation, they are likely to be insufficient for predicted segmentations and atypically shaped drawers.

\mypara{Baselines.}
We use the image-conditioned generation method SINGAPO~\cite{liu2025singapo}, and articulated object reconstruction method URDFormer~\cite{chen2023urdformer} to study their interior completion and object reconstruction capabilities.

\section{Experiments}

We benchmark different approaches for the three stages to determine how far we are from automatically constructing an articulated container from a static mesh.
We focus on answering the following questions:
1) \textit{Can methods trained on PM-Openable generalize to other datasets with more diverse shapes? Will extra data help in generalization?}
2) \textit{How difficult is each stage?  Is a learned method always necessary and better?}
3) \textit{Do our proposed methods result in better segmentation and motion prediction?}

We train on \pmopen (train split), and evaluate on \pmopen (val split) and \acd.
We also study the impact of including \pmopenext into the training split, to assess generalization and the complexity of \acd.

\begin{figure*}
\vspace{-5pt}
\centering
\setkeys{Gin}{width=\linewidth}
\begin{tabularx}{\textwidth}{@{} p{2cm}  Y @{\hskip 0pt} Y @{\hskip 0pt} Y @{\hskip 0pt} Y @{\hskip 0pt} Y @{\hskip 0pt} Y @{\hskip 0pt} Y @{\hskip 0pt} Y @{}}

\toprule

\textbf{Method} & \multicolumn{4}{c}{\textbf{\pmopen}} & \multicolumn{4}{c}{\textbf{\ourdatashort}}\\
\cmidrule(l{0pt}r{2pt}){2-5} \cmidrule(l{2pt}r{0pt}){6-9}

GT
& \includegraphics[trim=50 30 50 140,clip]{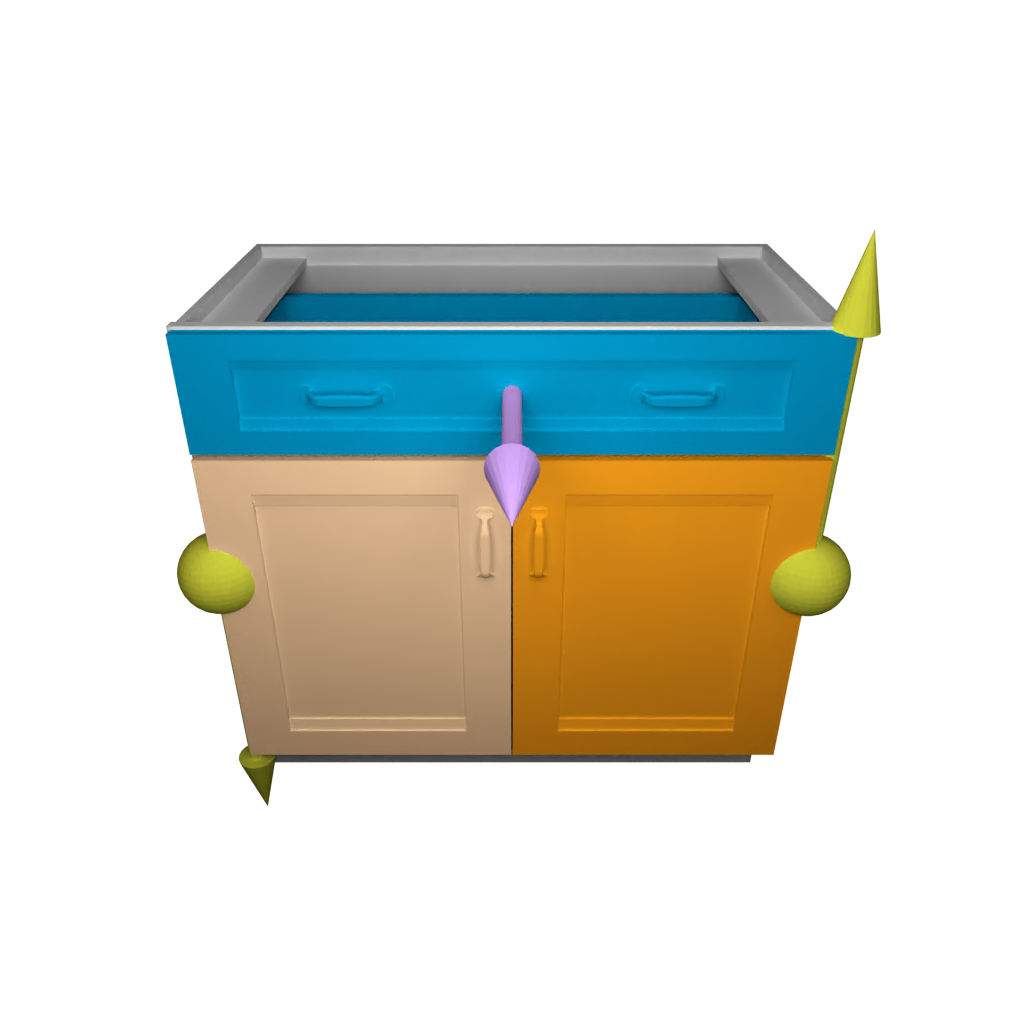}
& \includegraphics[trim=50 30 50 140,clip]{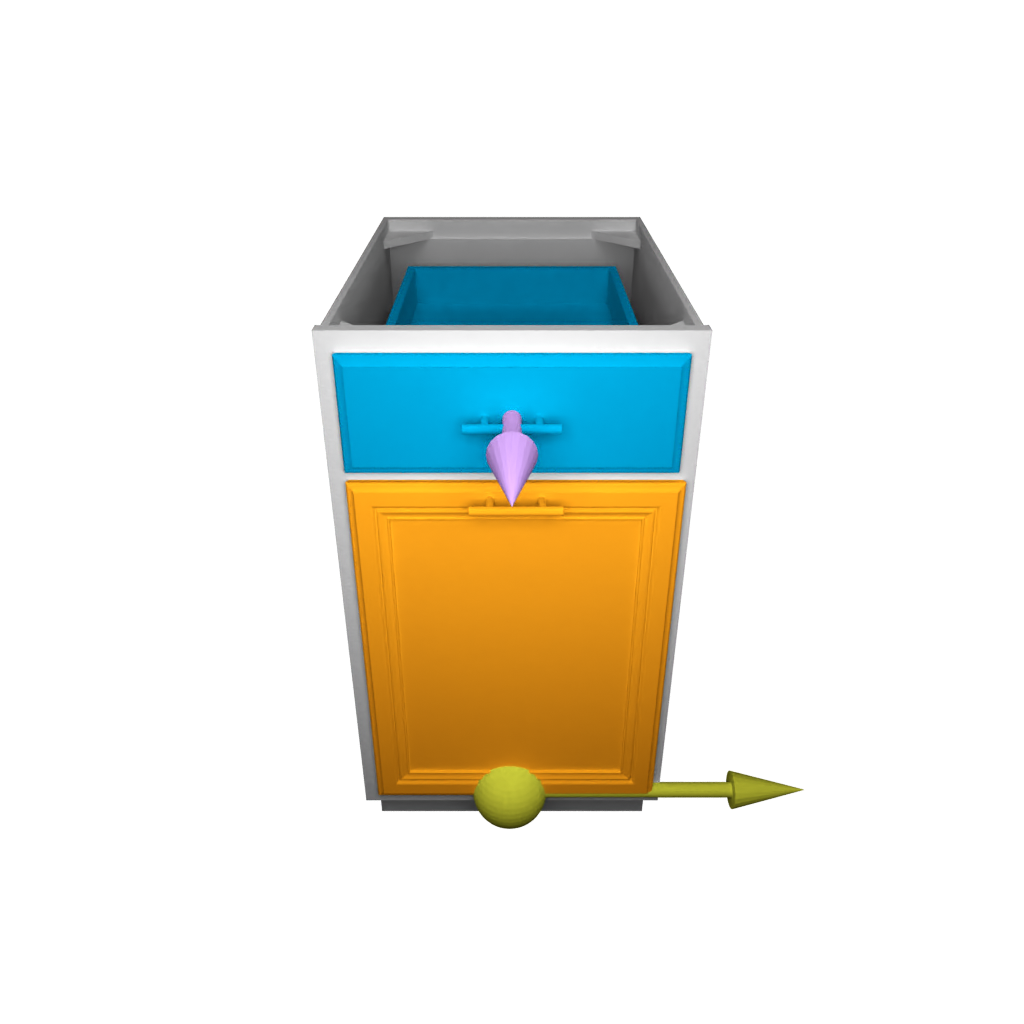}
& \includegraphics[trim=50 30 50 140,clip]{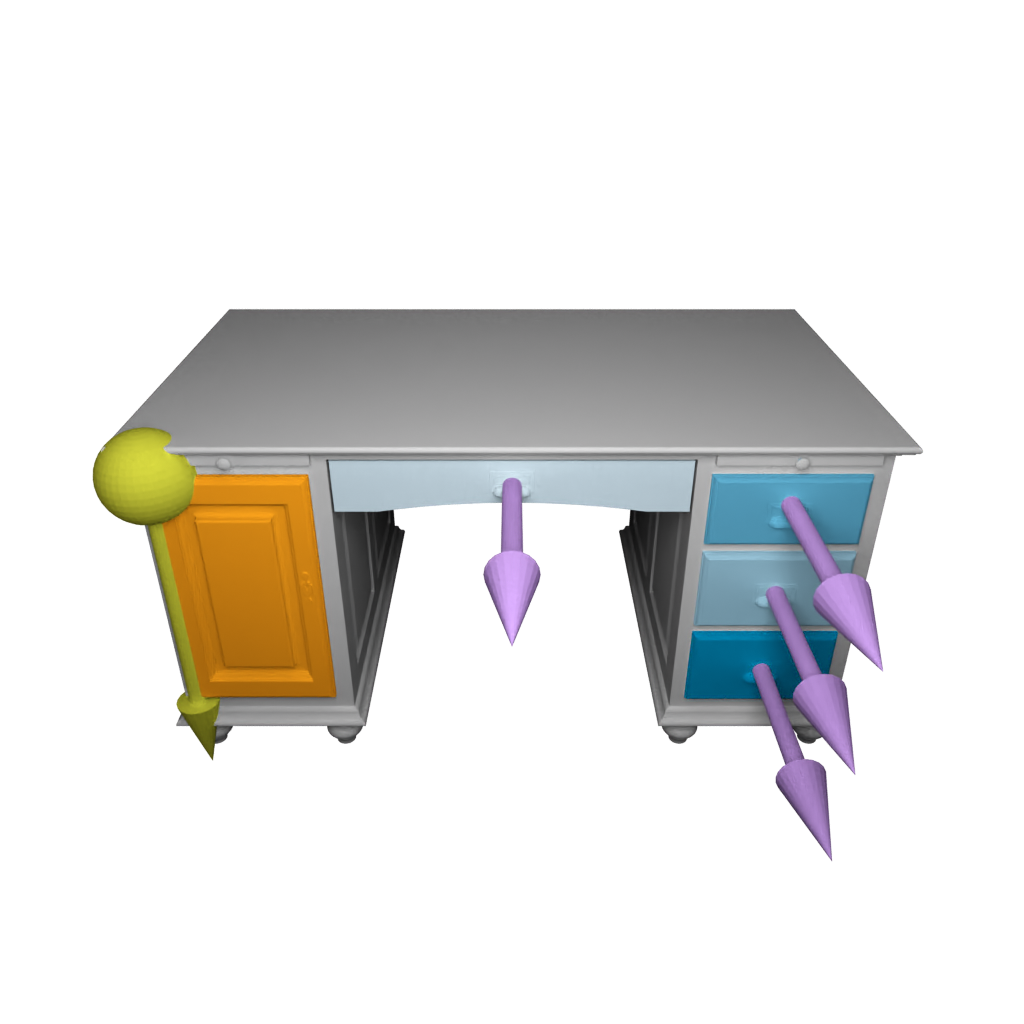}
& \includegraphics[trim=50 30 50 140,clip]{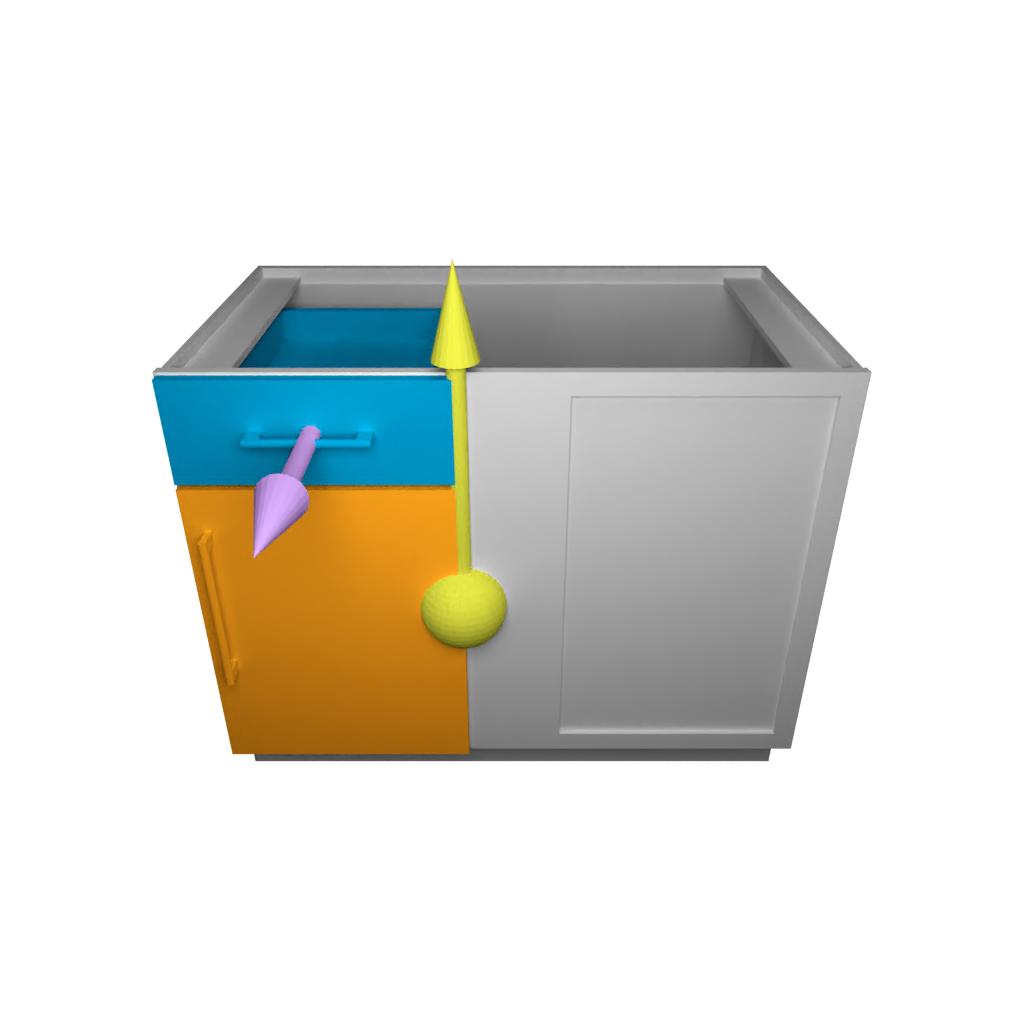}
& \includegraphics[trim=50 30 50 140,clip]{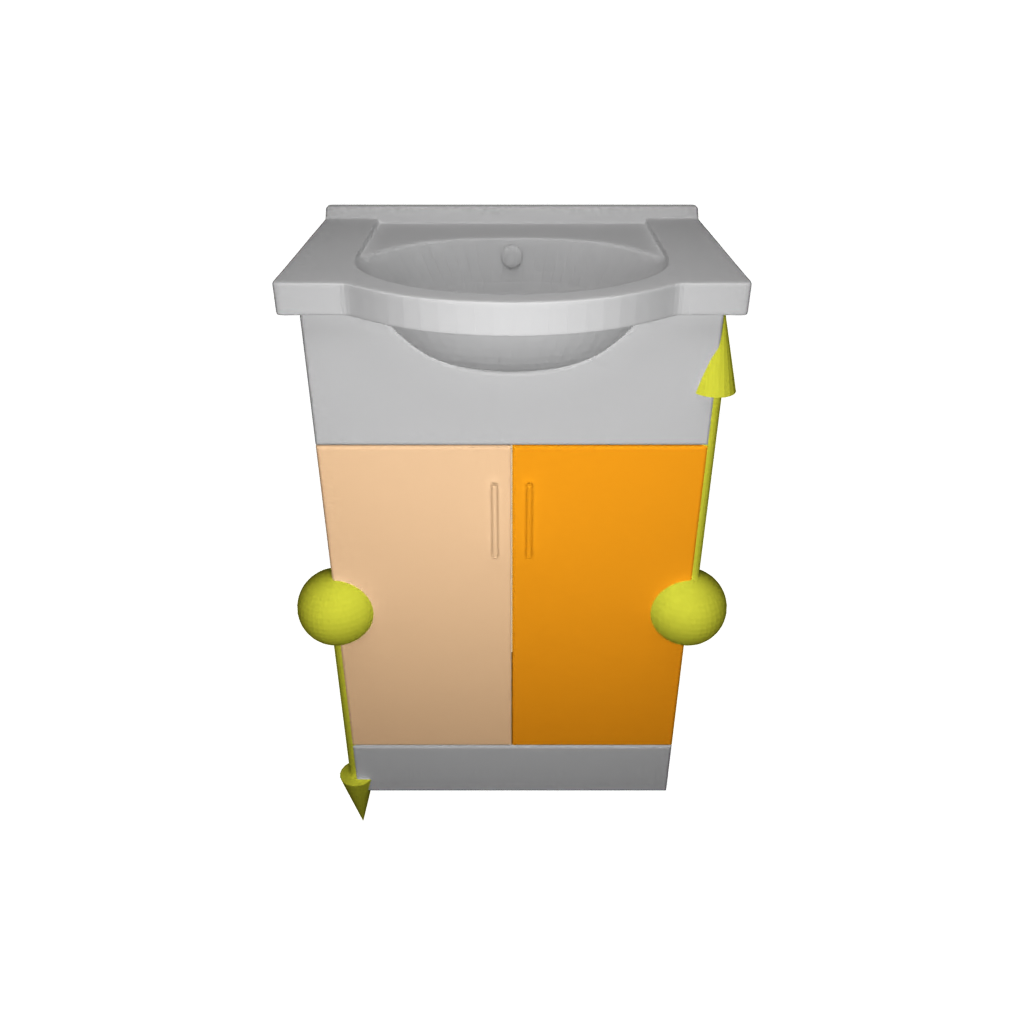}
& \includegraphics[trim=50 30 50 140,clip]{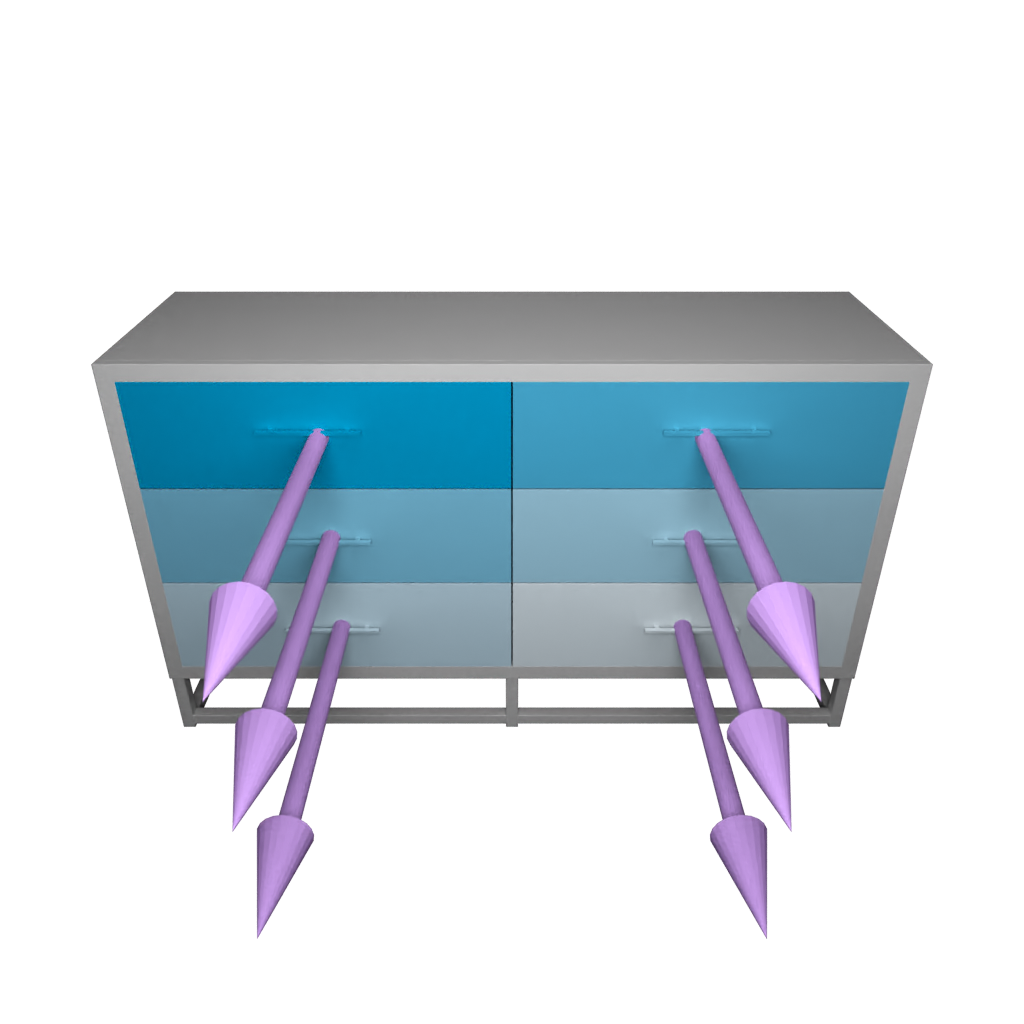}
& \includegraphics[trim=50 30 50 140,clip]{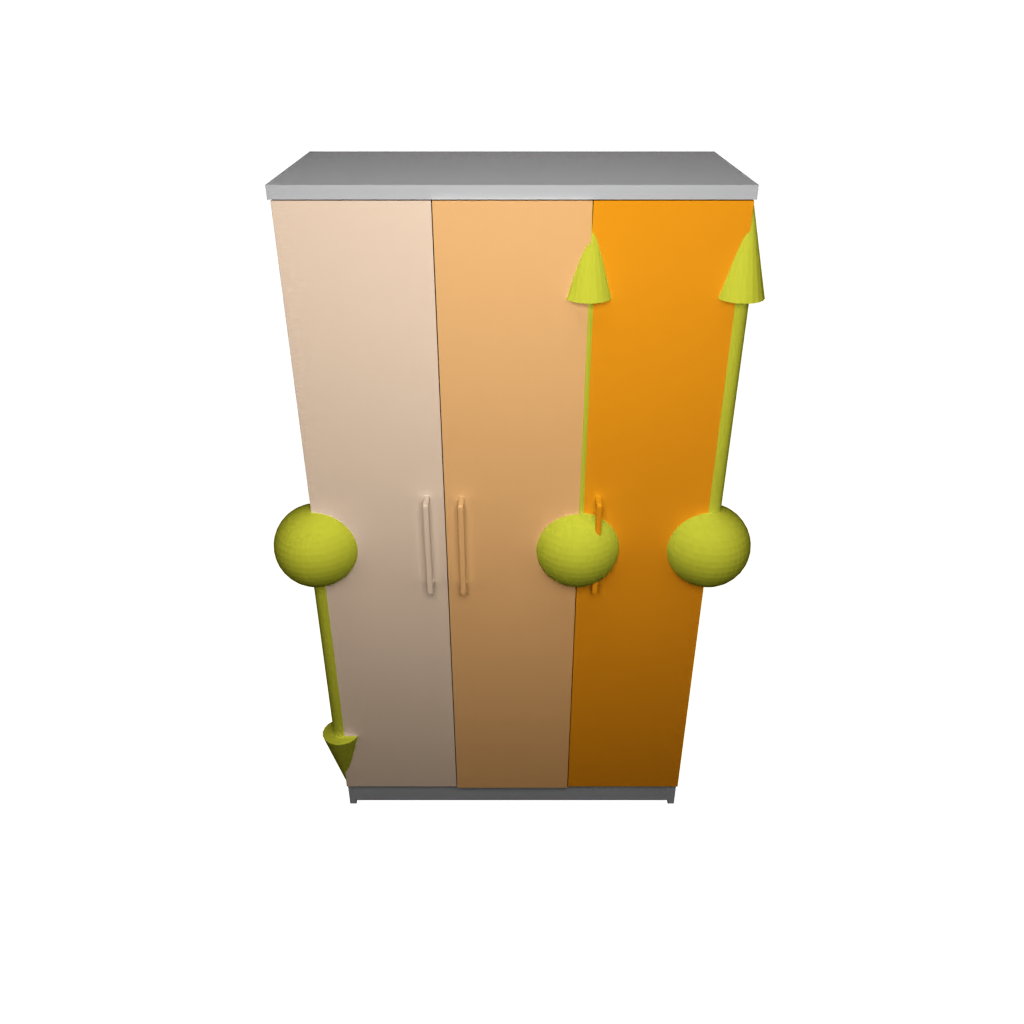}
& \includegraphics[trim=50 30 50 140,clip]{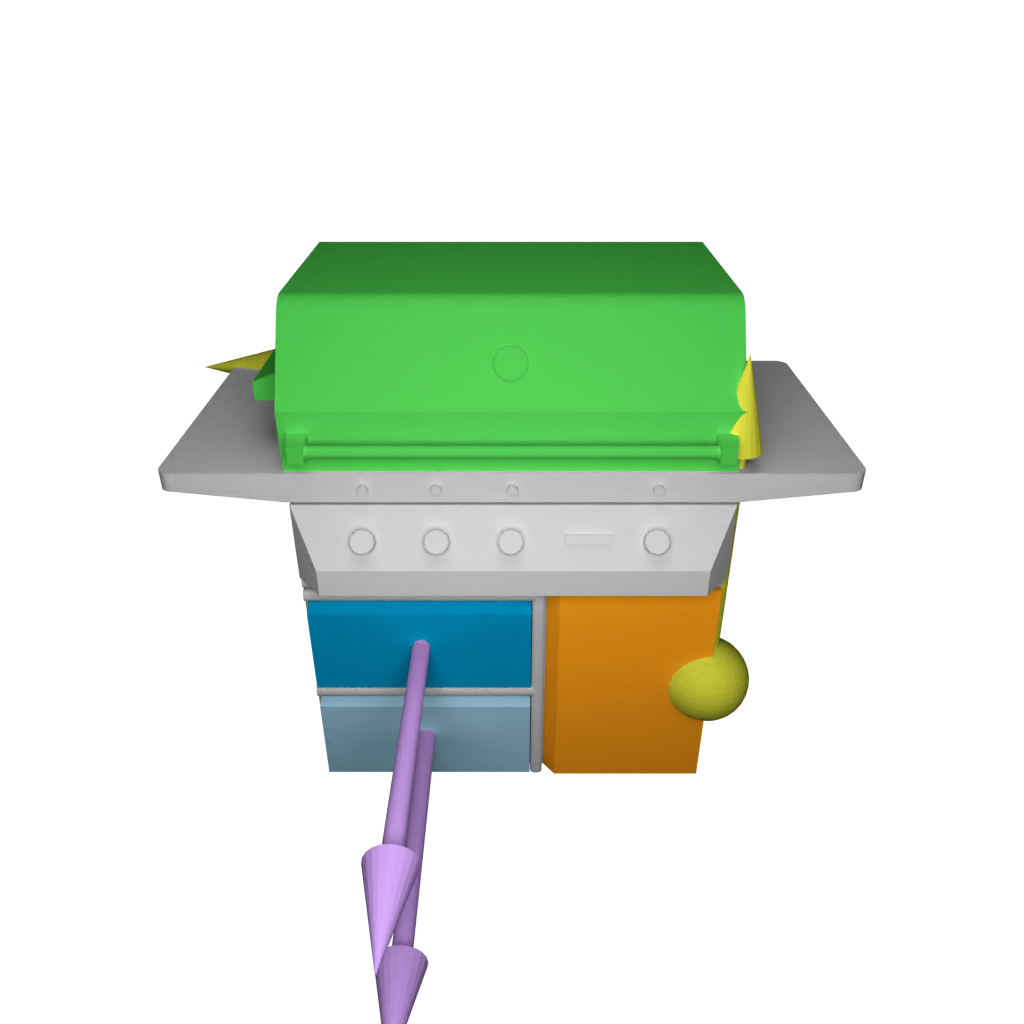}\\

\maskthreed
& \includegraphics[trim=50 140 50 140,clip]{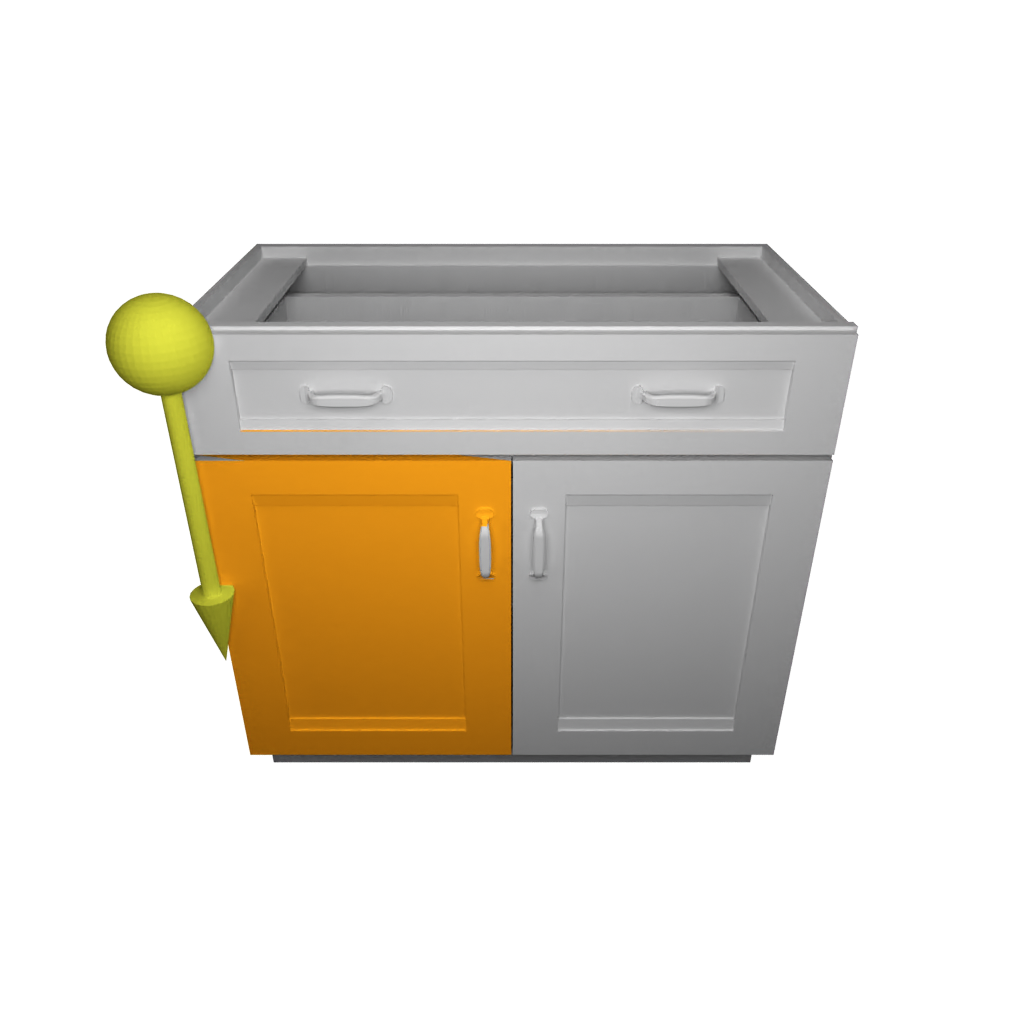}
& \includegraphics[trim=50 140 50 140,clip]{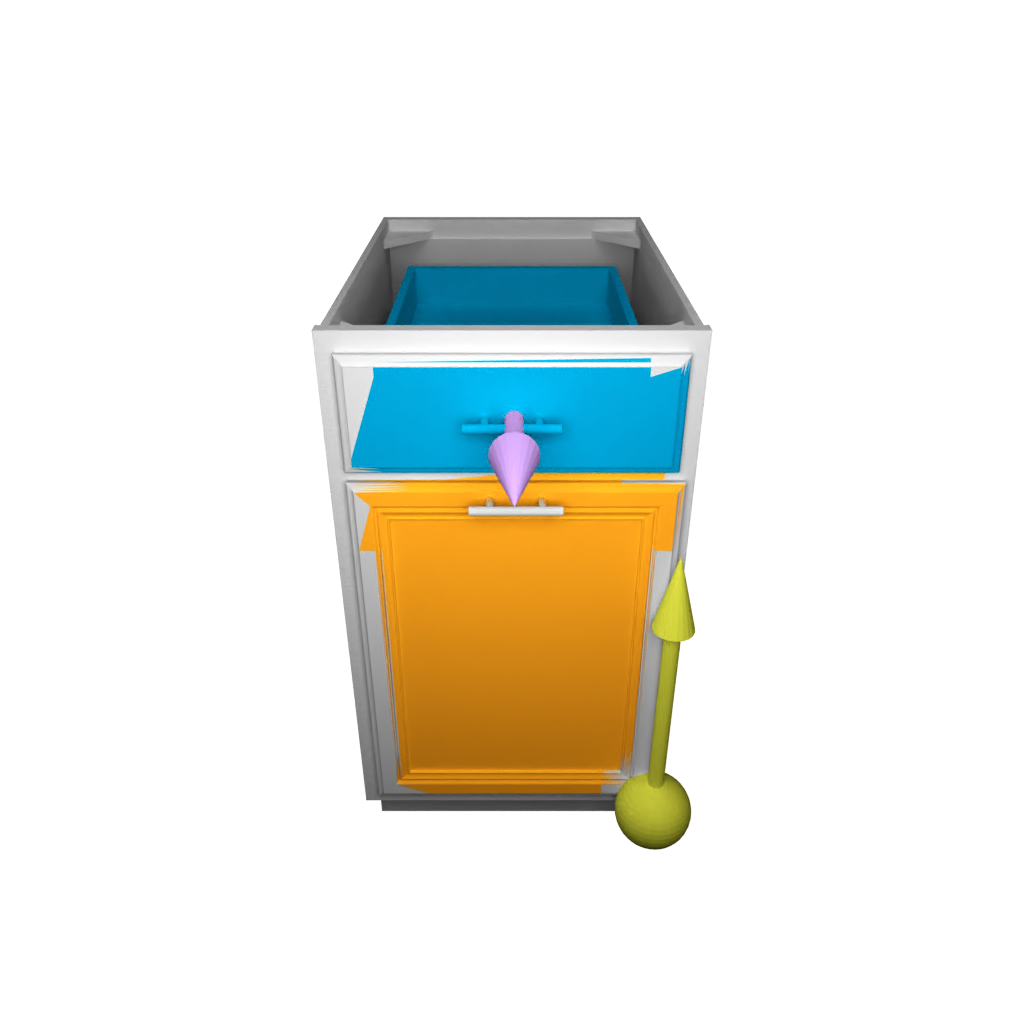}
& \includegraphics[trim=50 140 50 140,clip]{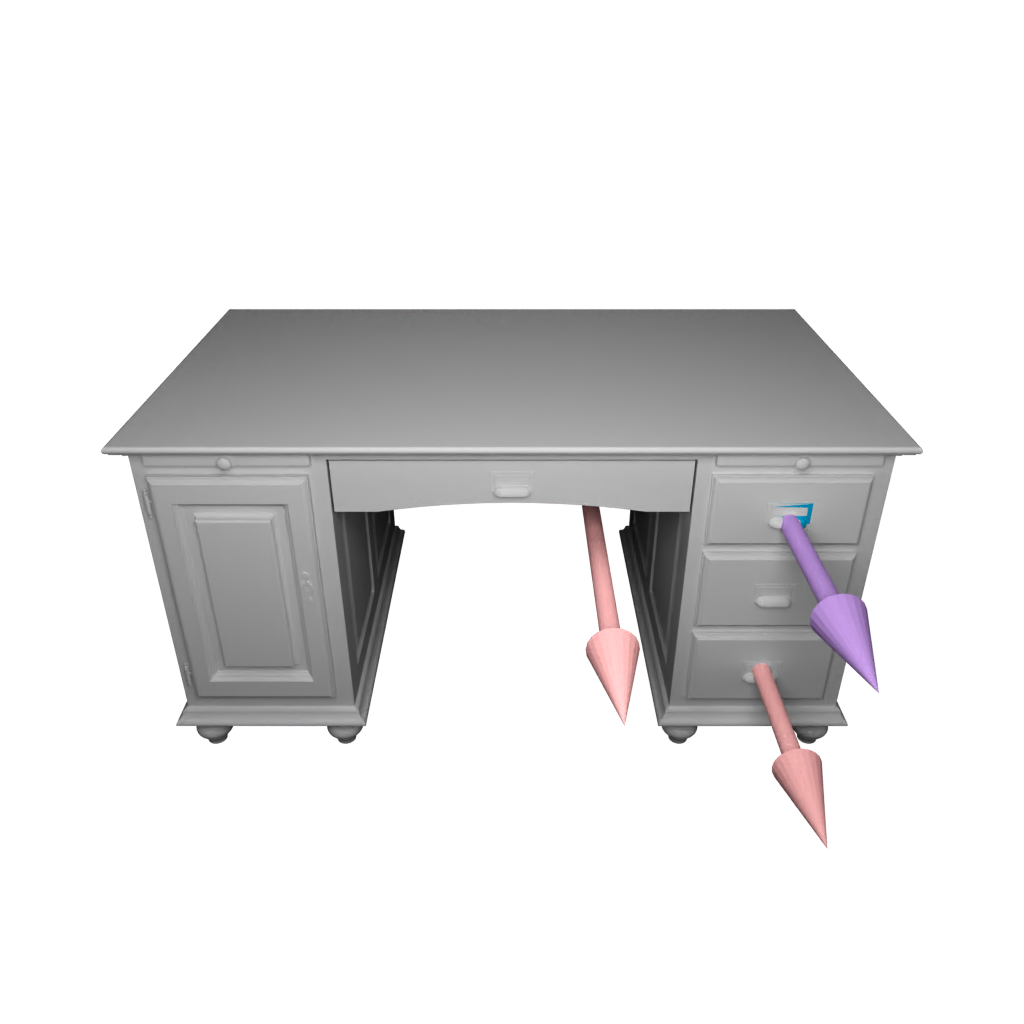}
& \includegraphics[trim=50 140 50 140,clip]{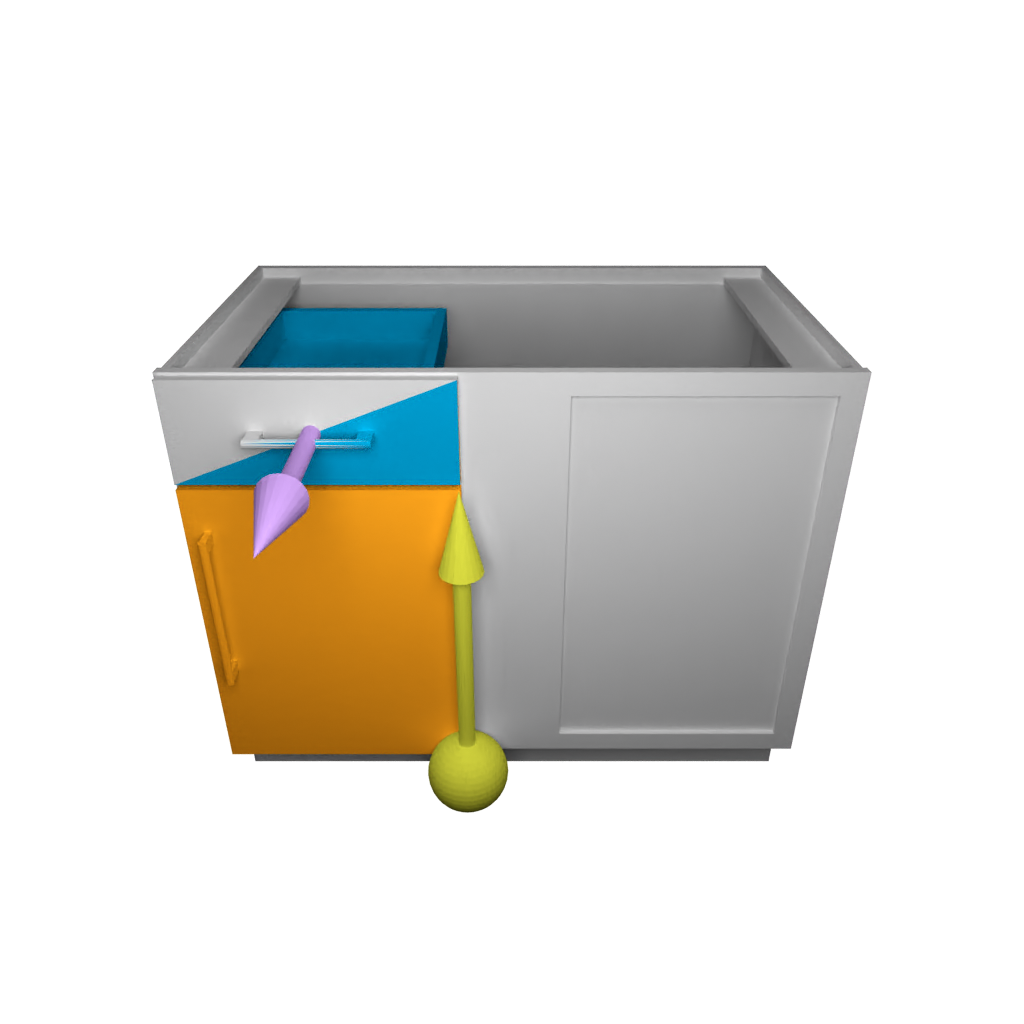}
& \includegraphics[trim=50 140 50 140,clip]{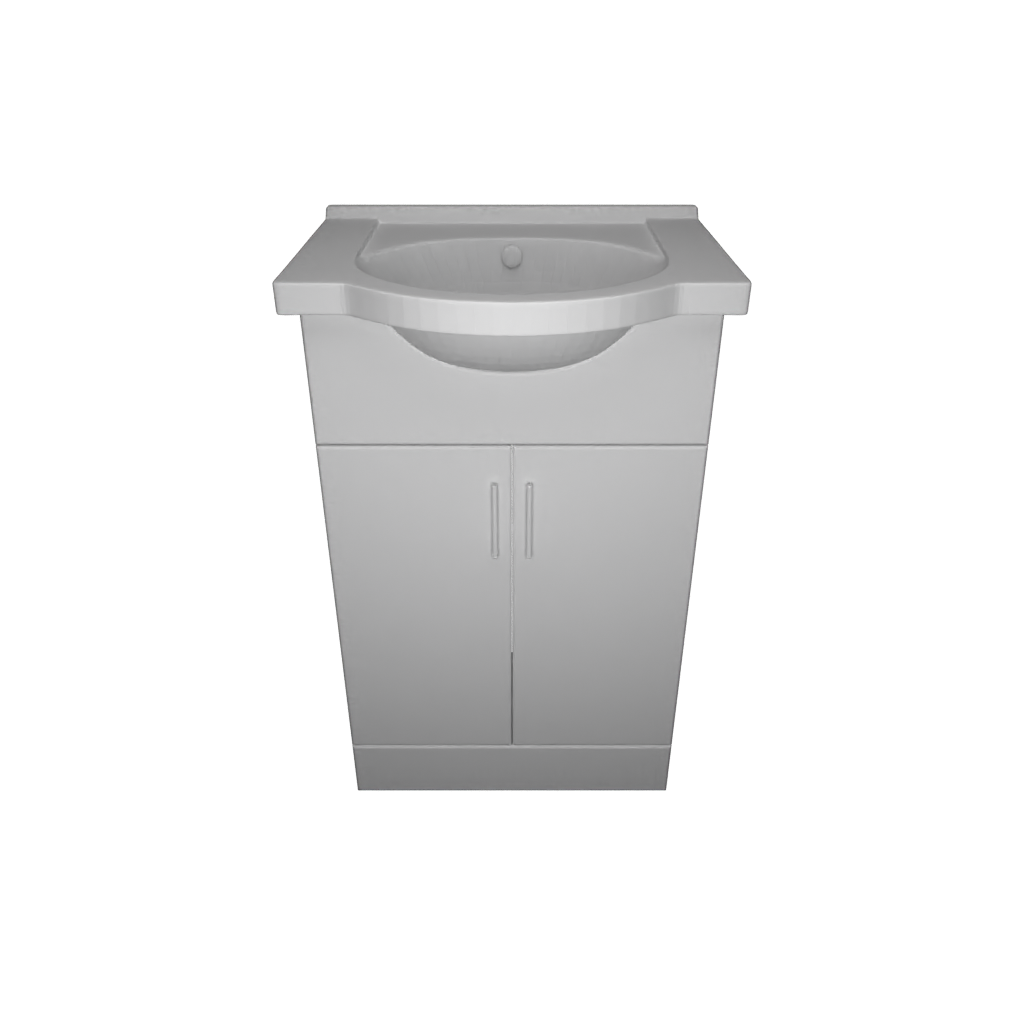}
& \includegraphics[trim=50 140 50 140,clip]{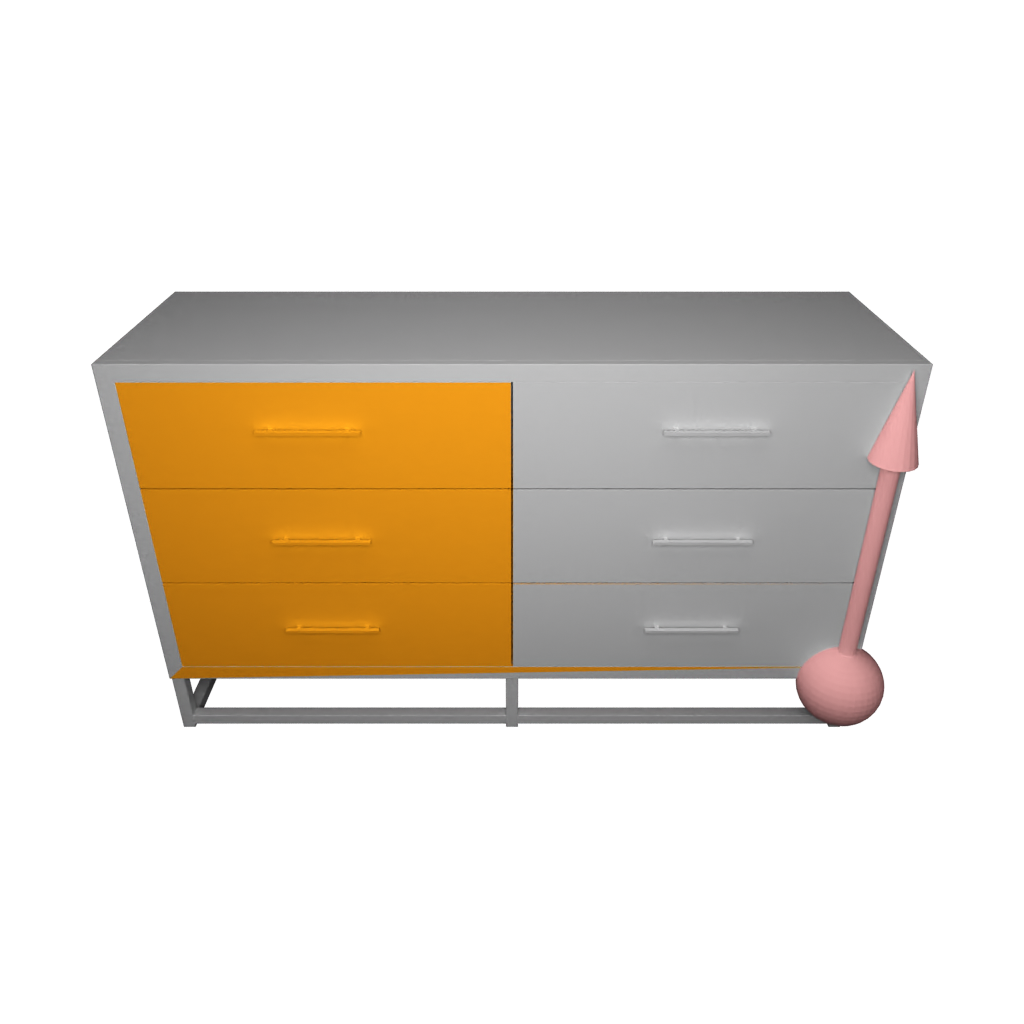}
& \includegraphics[trim=50 140 50 140,clip]{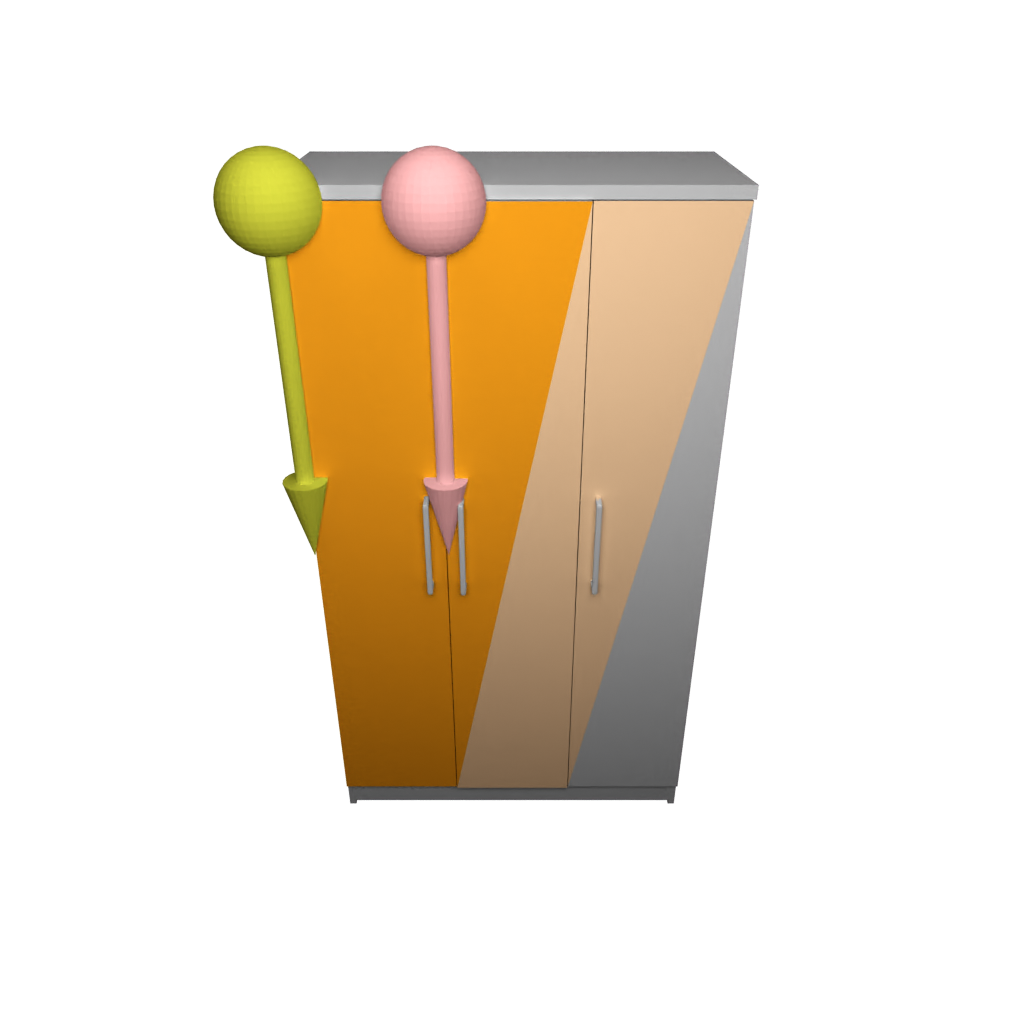}
& \includegraphics[trim=50 140 50 140,clip]{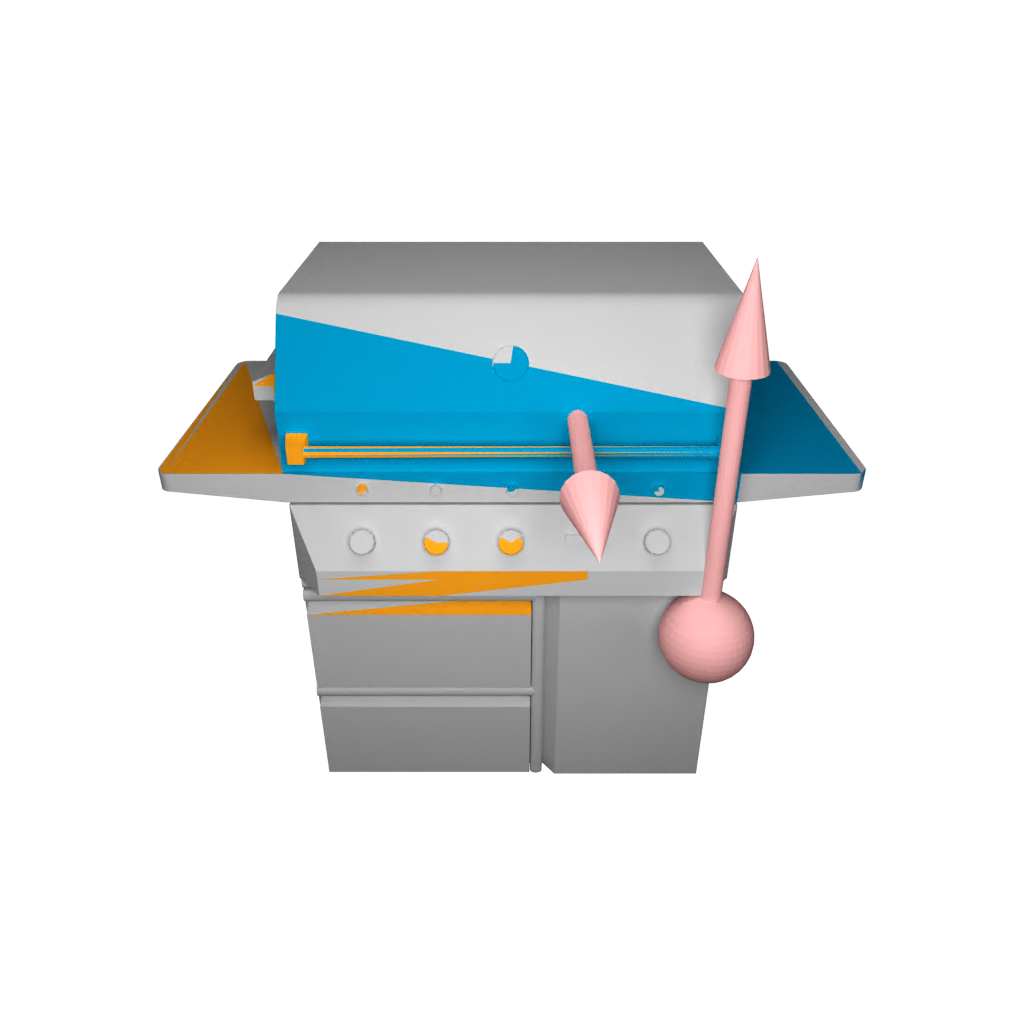}\\

\ogroup
& \includegraphics[trim=50 140 50 140,clip]{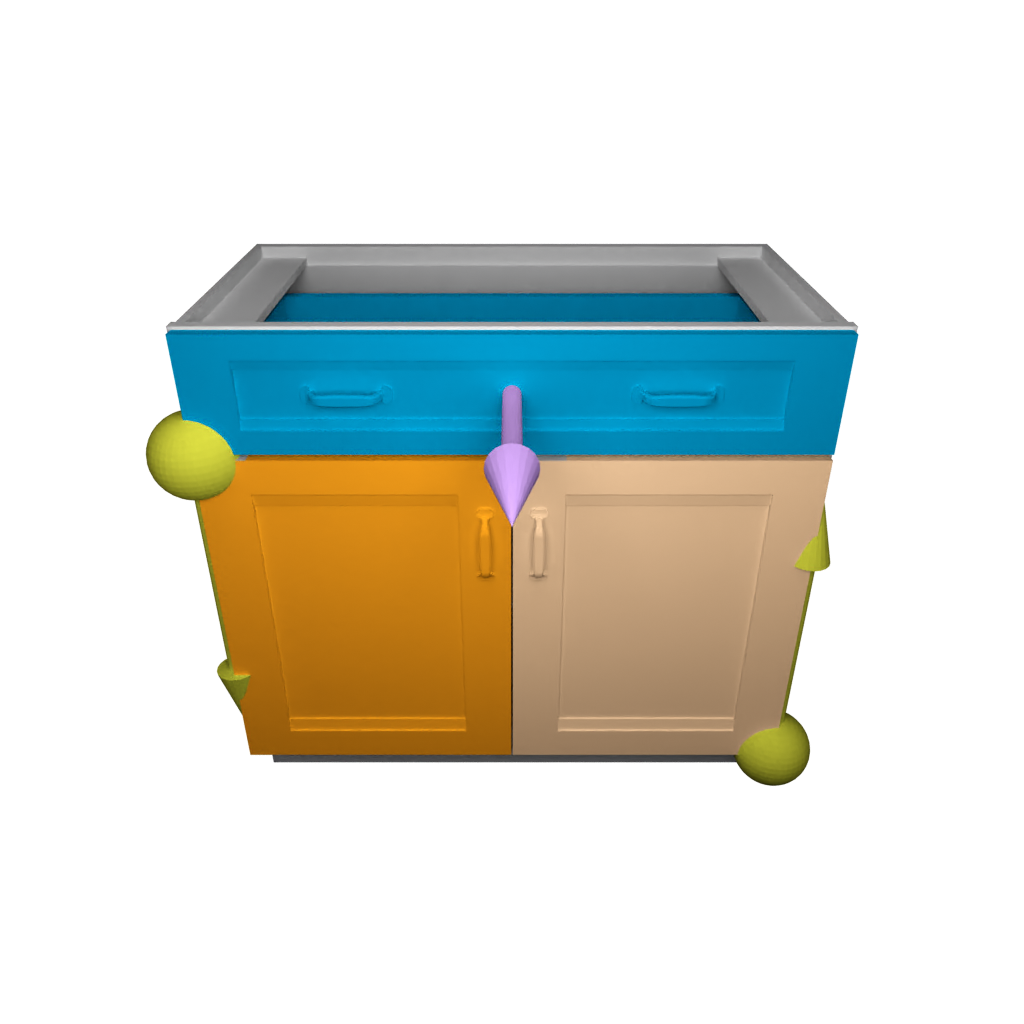}
& \includegraphics[trim=50 140 50 140,clip]{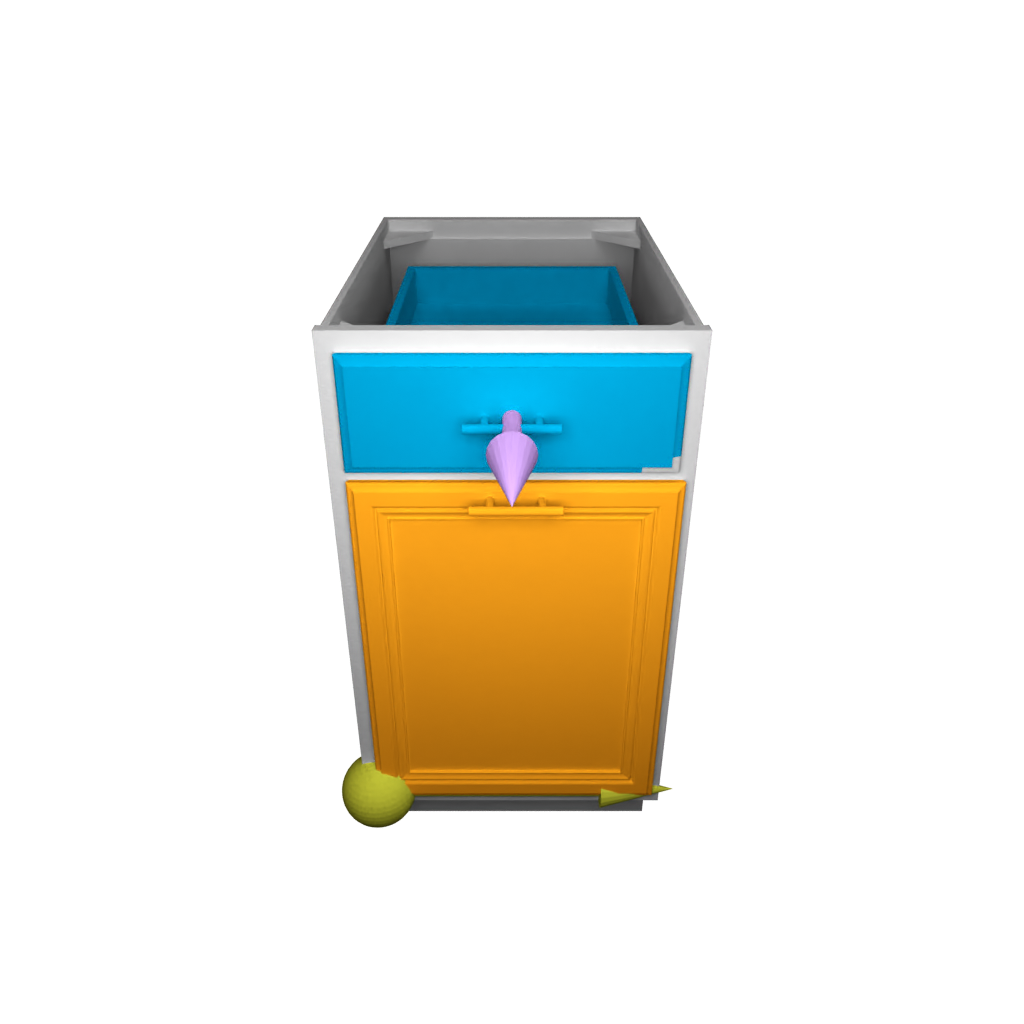}
& \includegraphics[trim=50 140 50 140,clip]{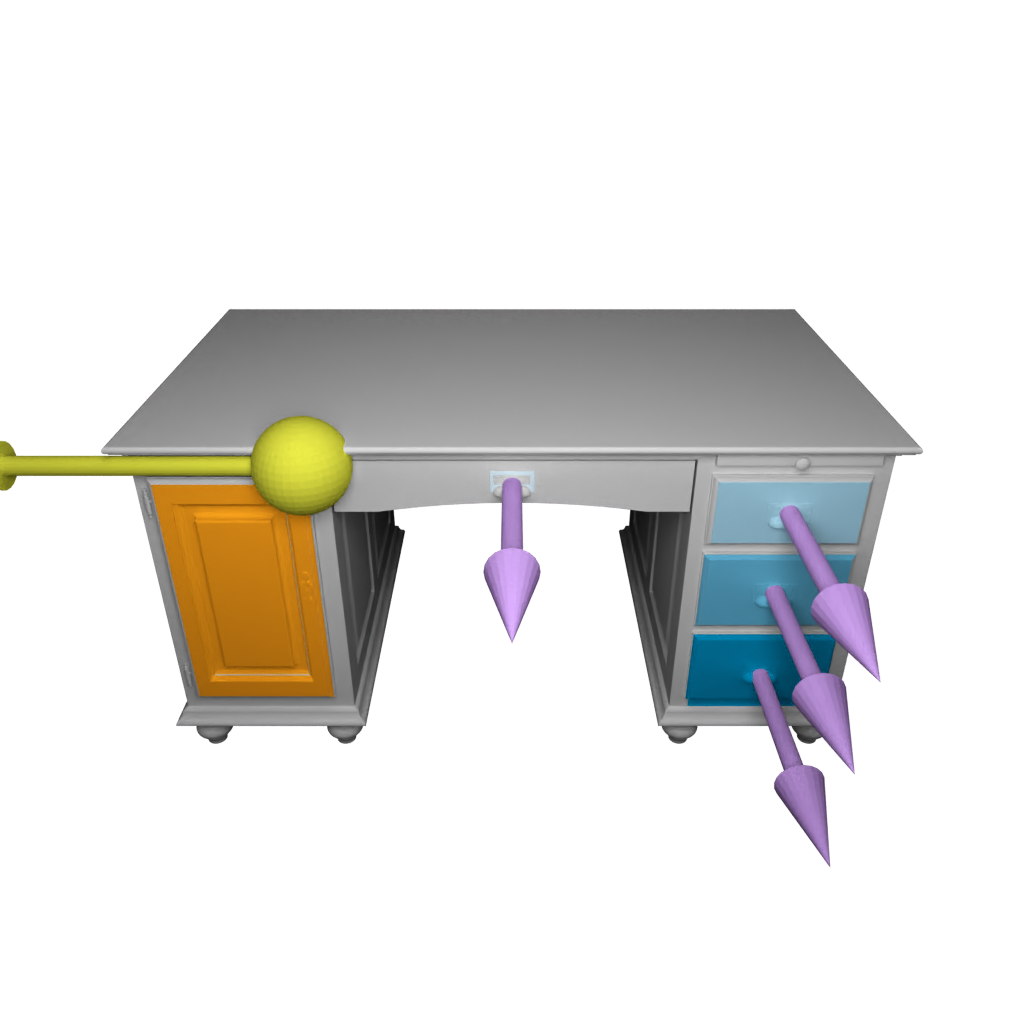}
& \includegraphics[trim=50 140 50 140,clip]{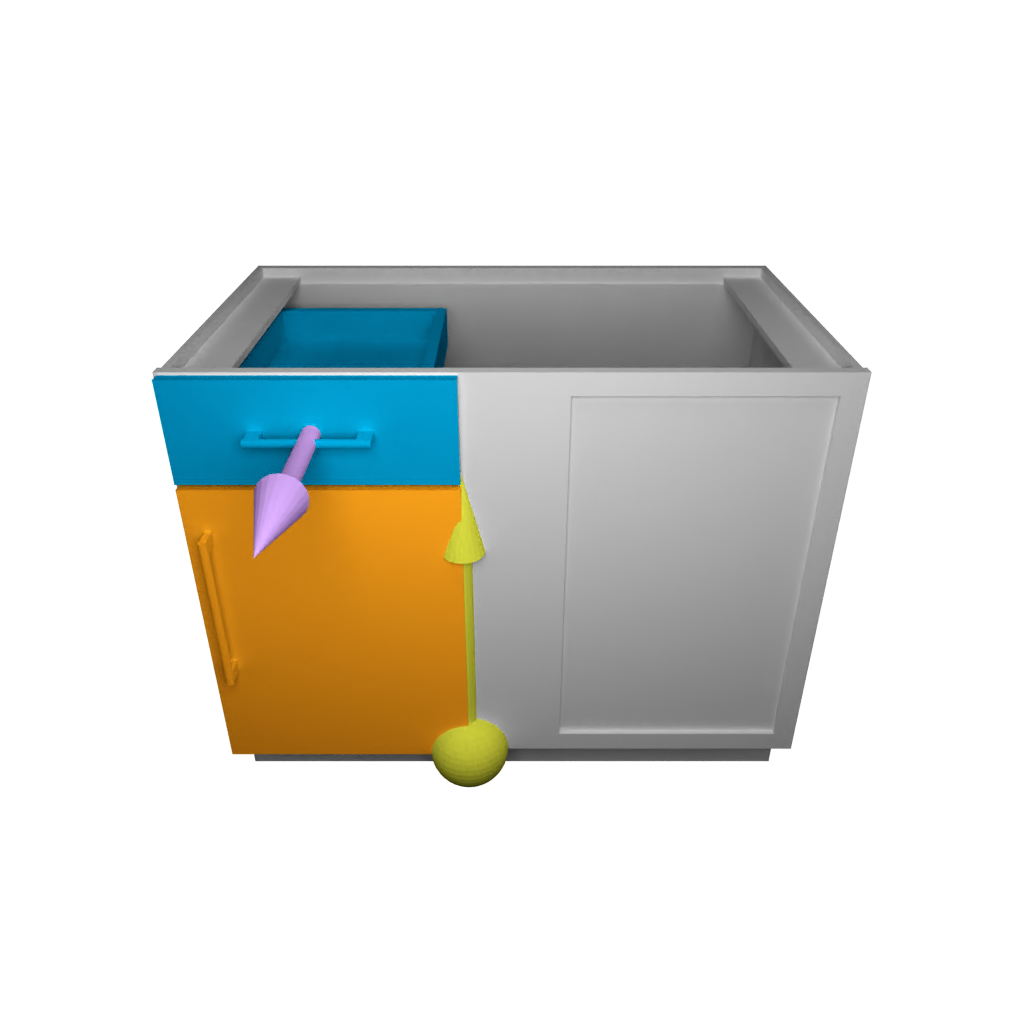}
& \includegraphics[trim=50 140 50 140,clip]{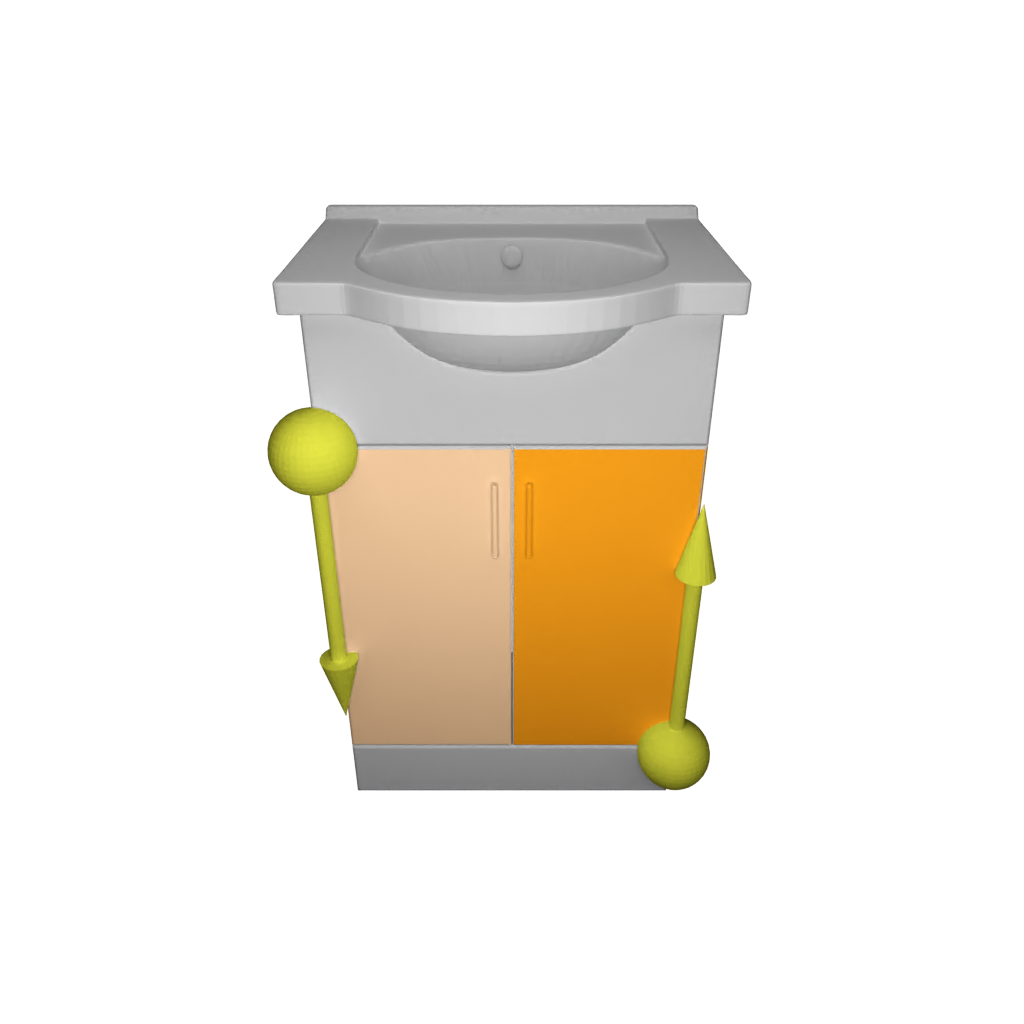}
& \includegraphics[trim=50 140 50 140,clip]{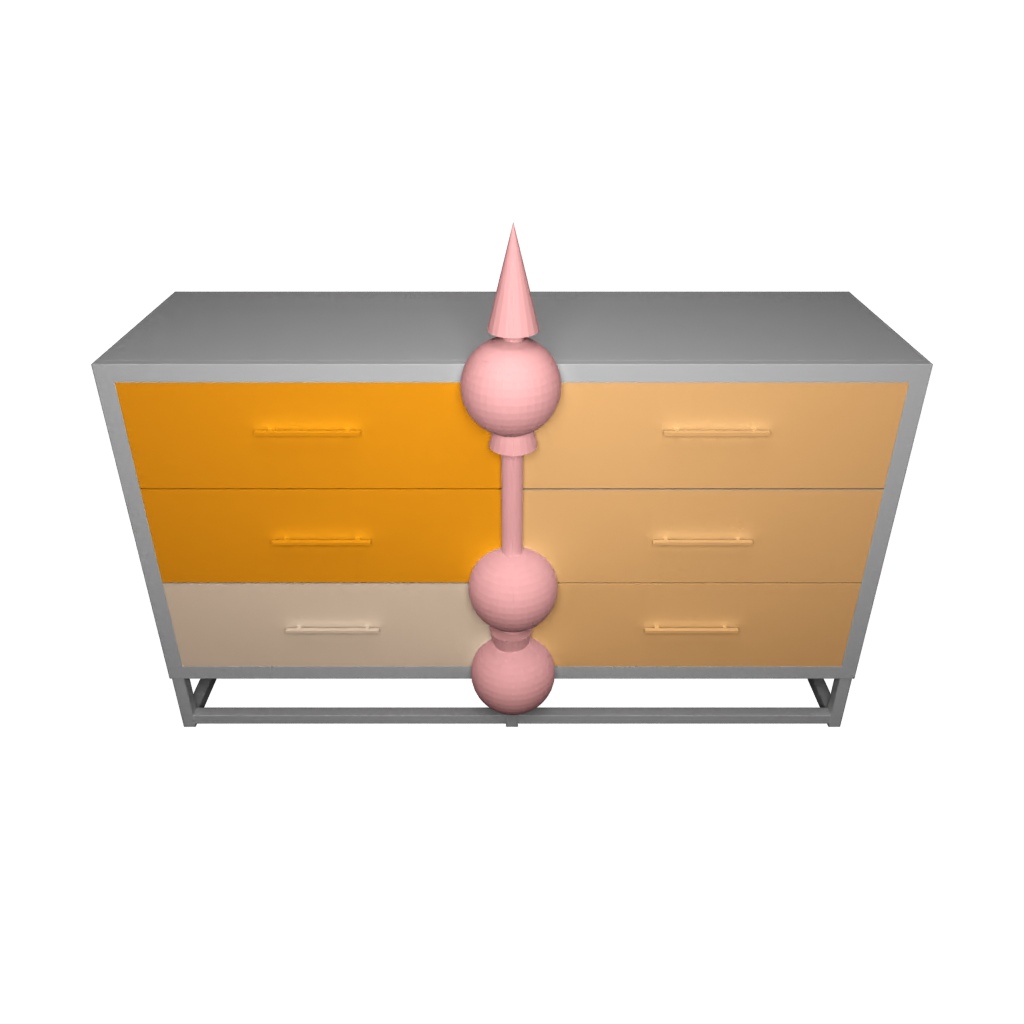}
& \includegraphics[trim=50 140 50 140,clip]{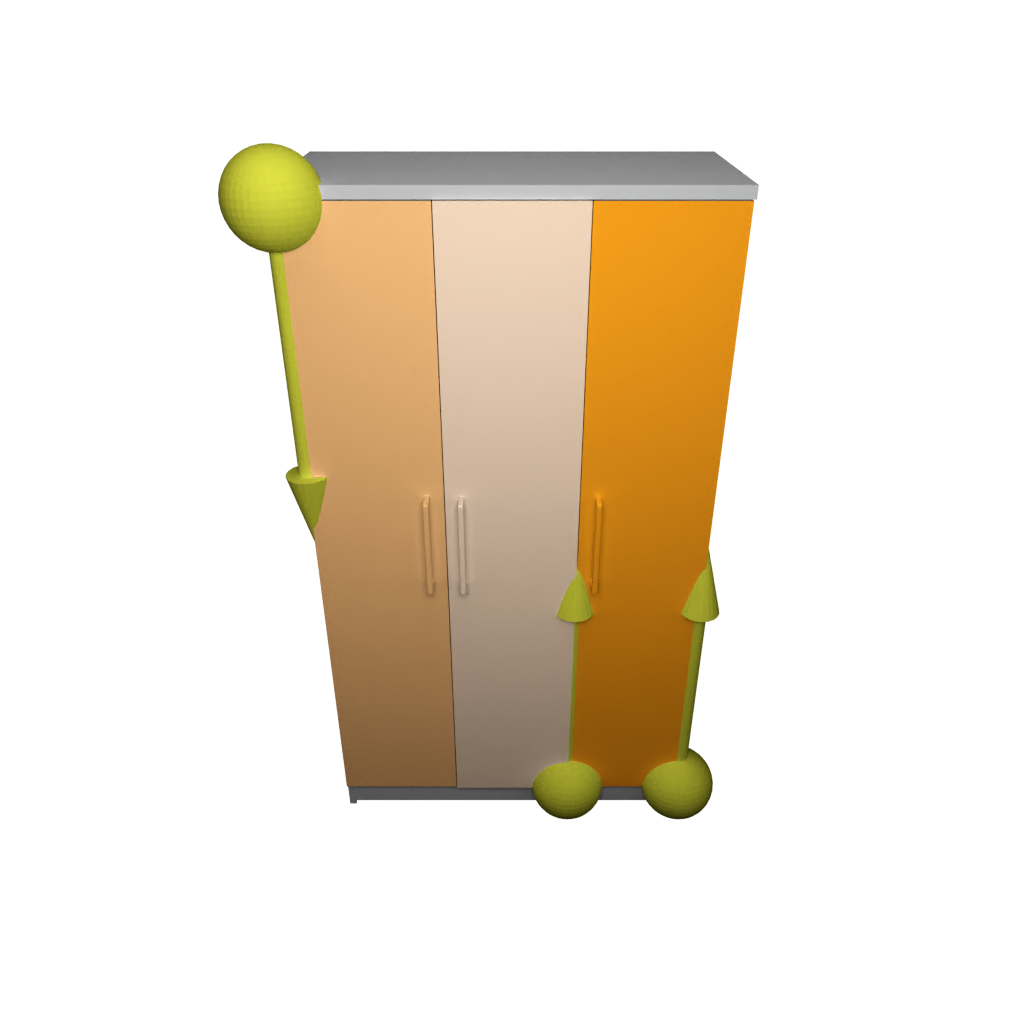}
& \includegraphics[trim=50 140 50 140,clip]{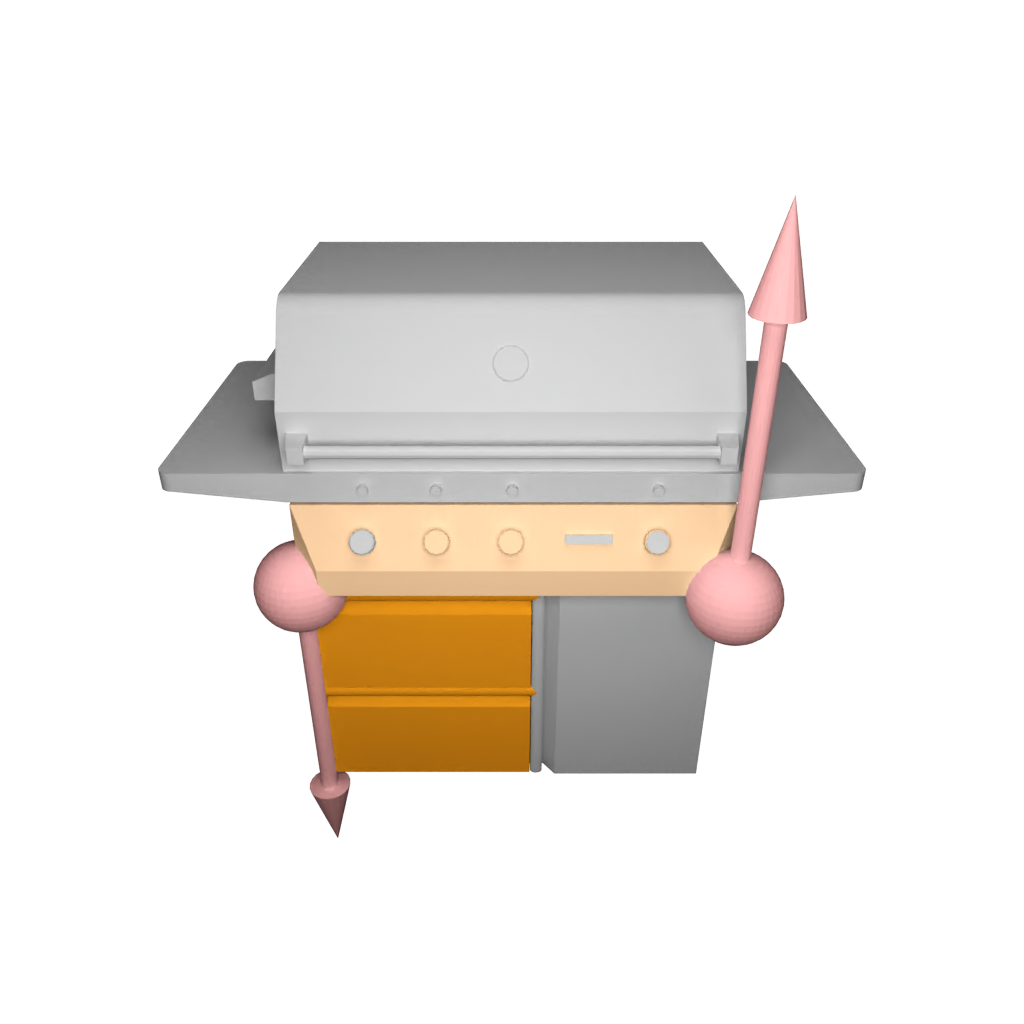}\\

\resizebox{!}{0.6em}{\gammagroup}
& \includegraphics[trim=50 110 50 140,clip]{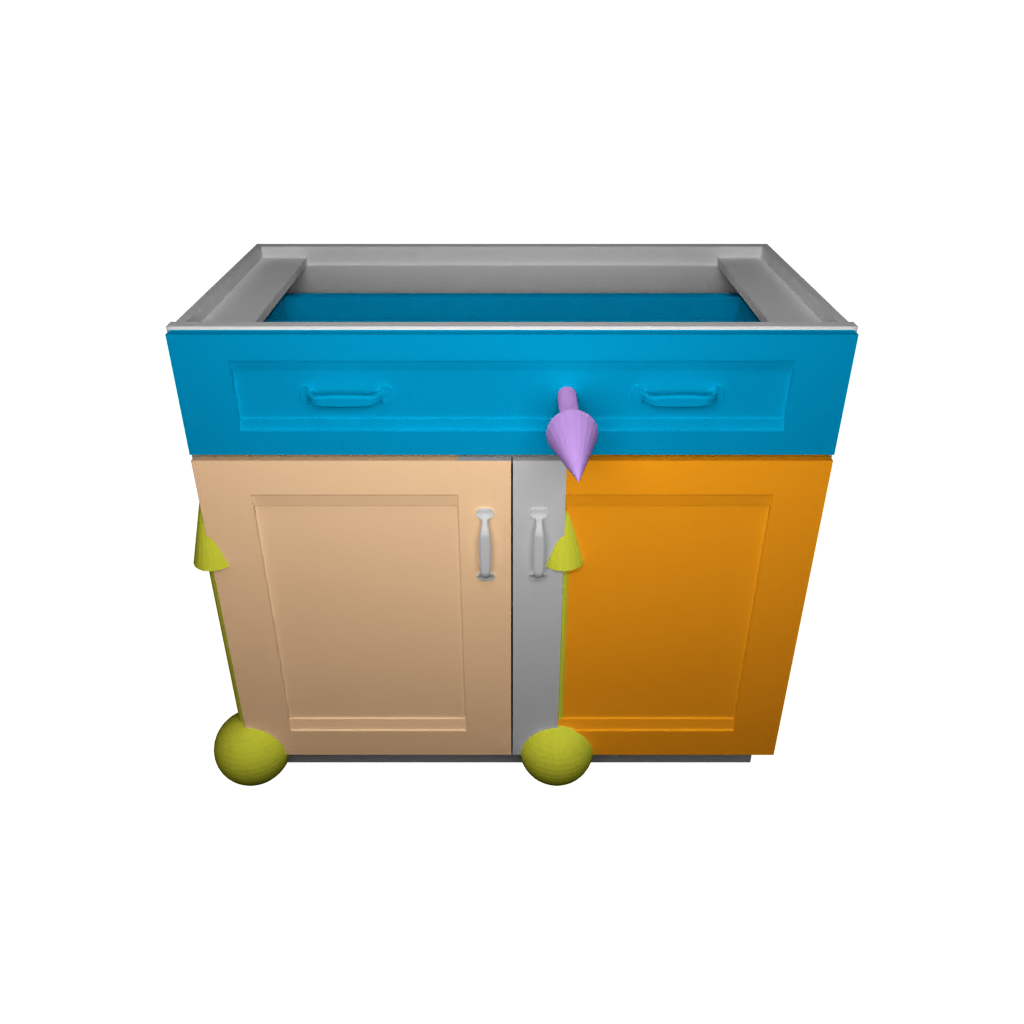}
& \includegraphics[trim=50 110 50 140,clip]{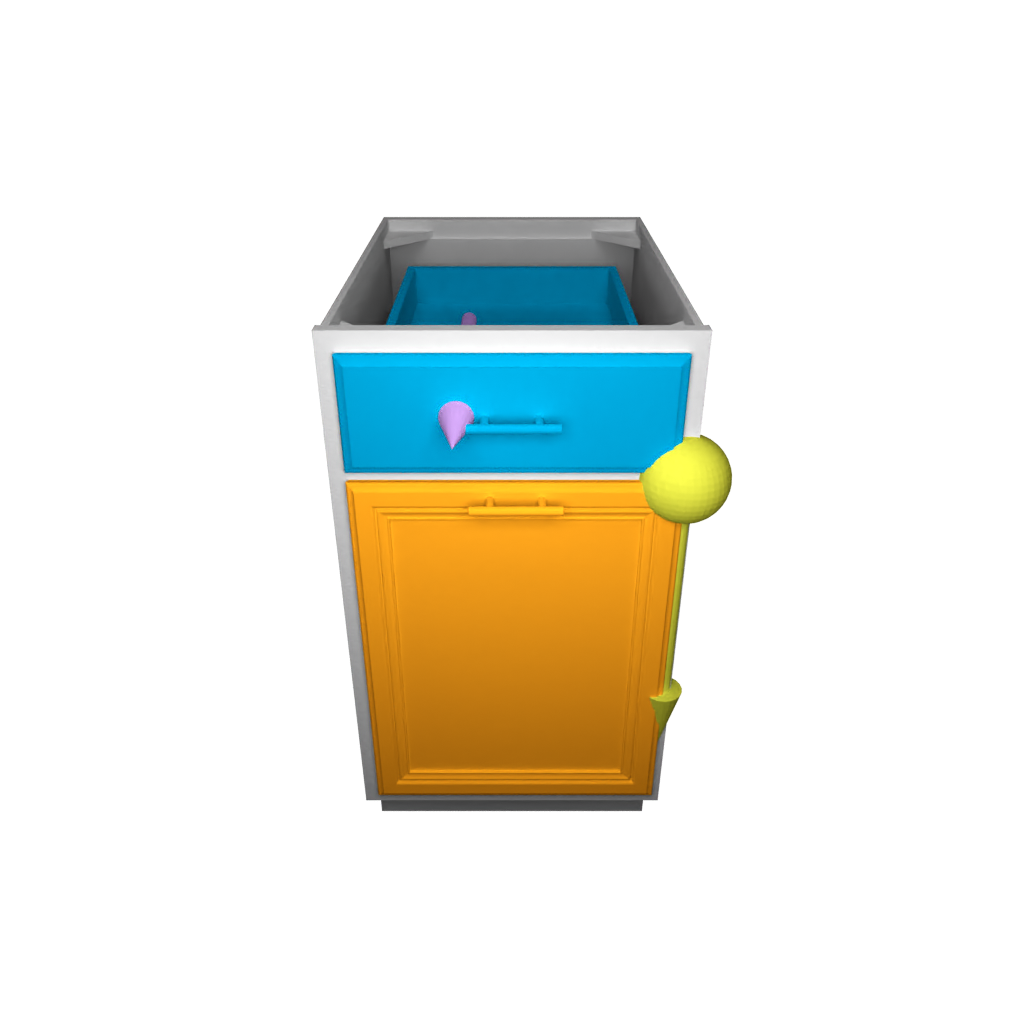}
& \includegraphics[trim=50 110 50 140,clip]{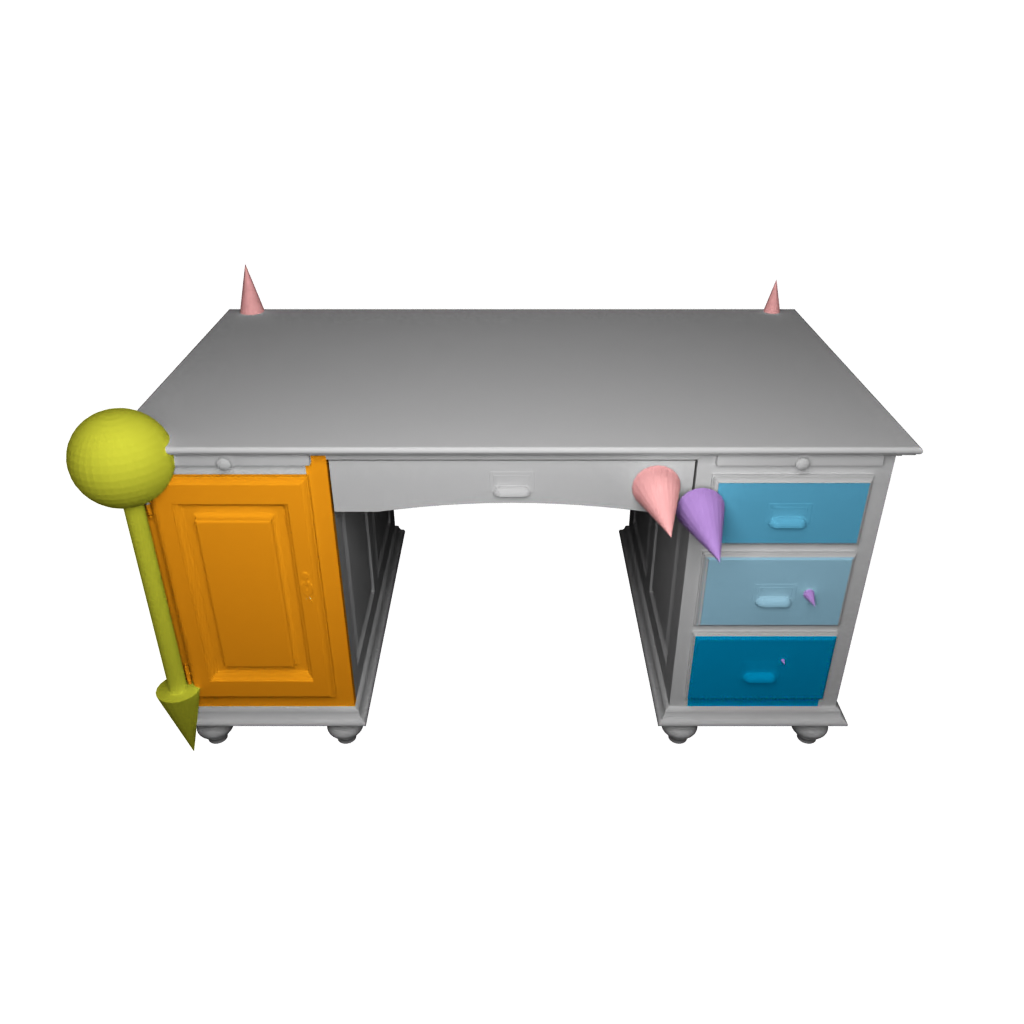}
& \includegraphics[trim=50 110 50 140,clip]{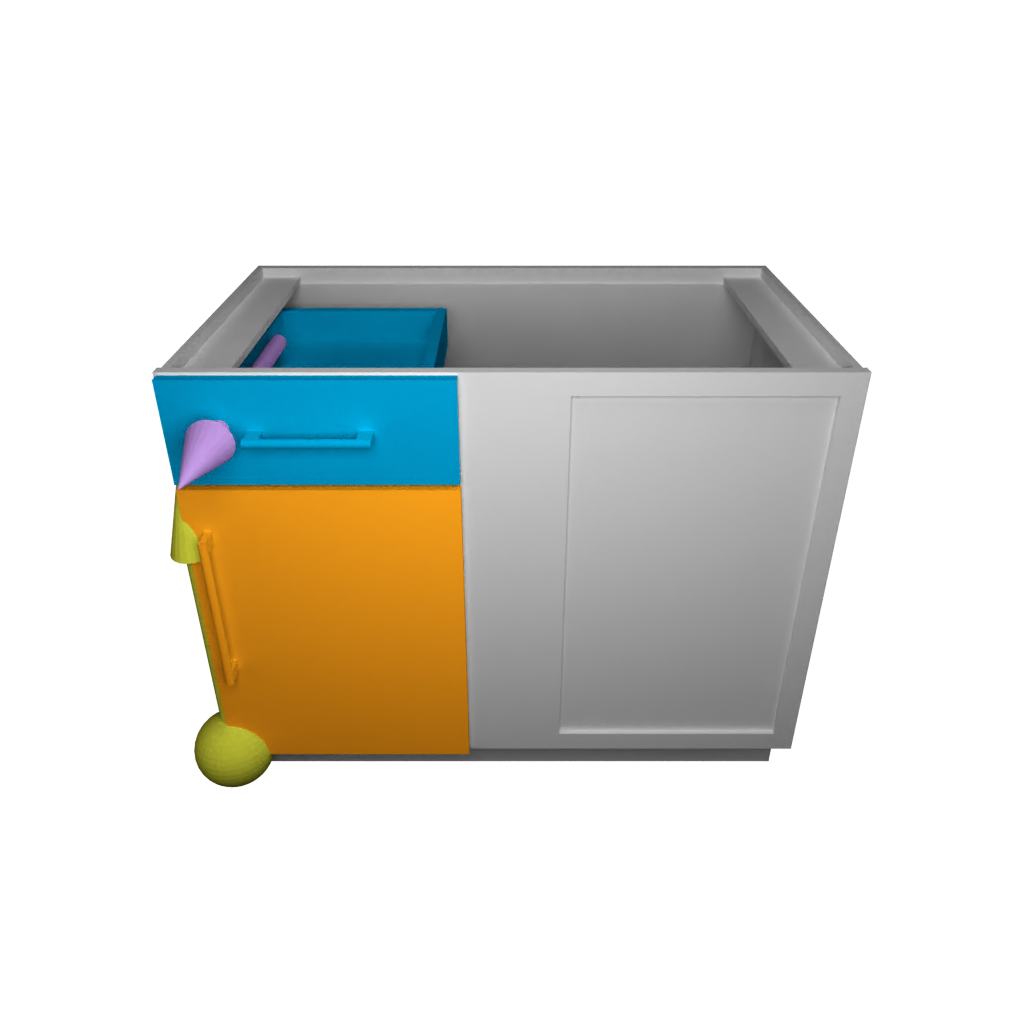}
& \includegraphics[trim=50 110 50 140,clip]{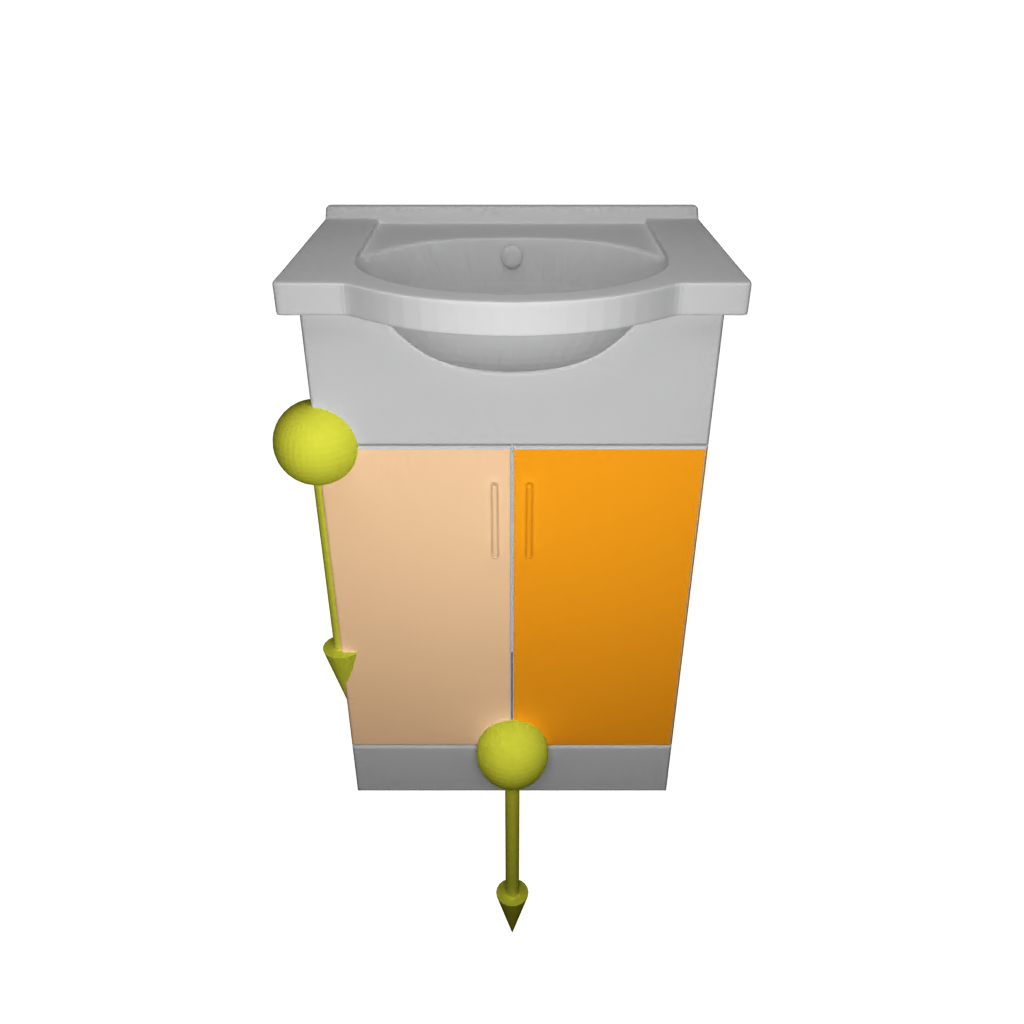}
& \includegraphics[trim=50 110 50 140,clip]{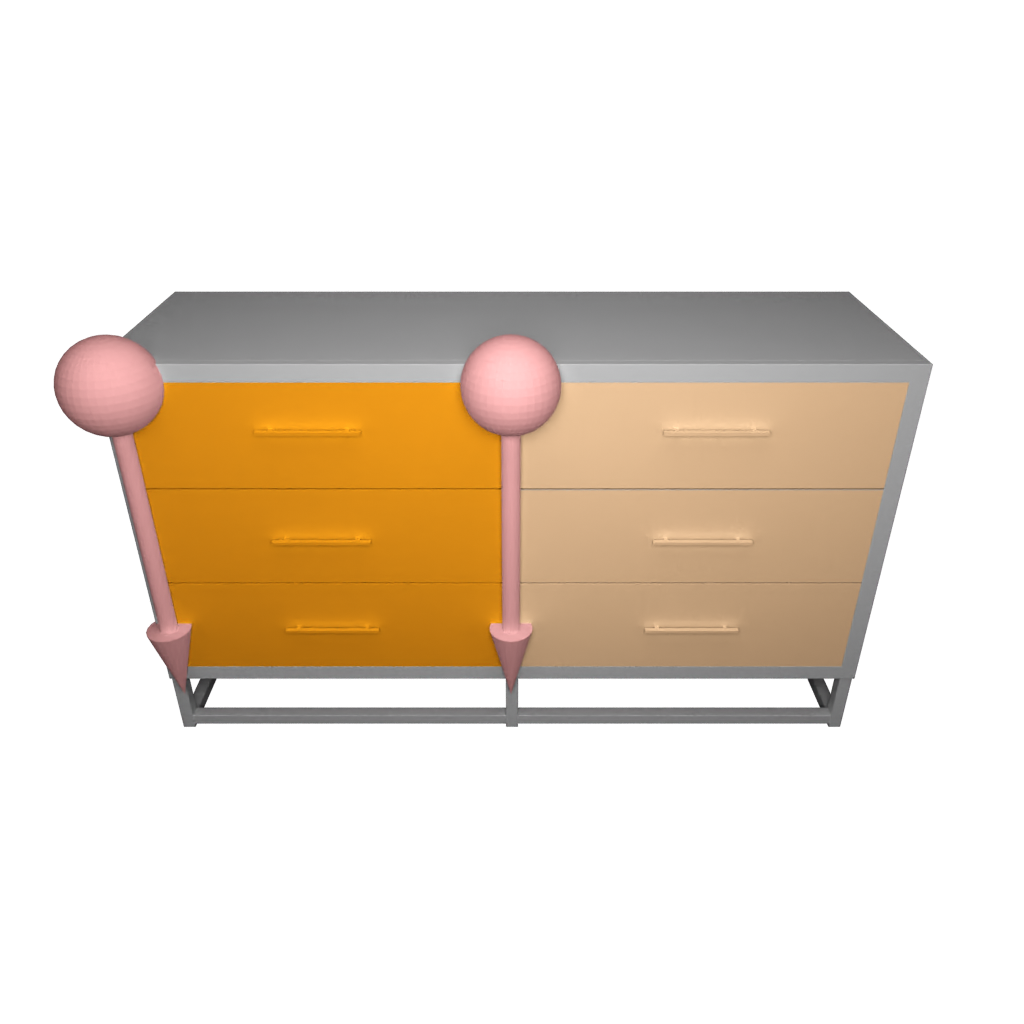}
& \includegraphics[trim=50 110 50 140,clip]{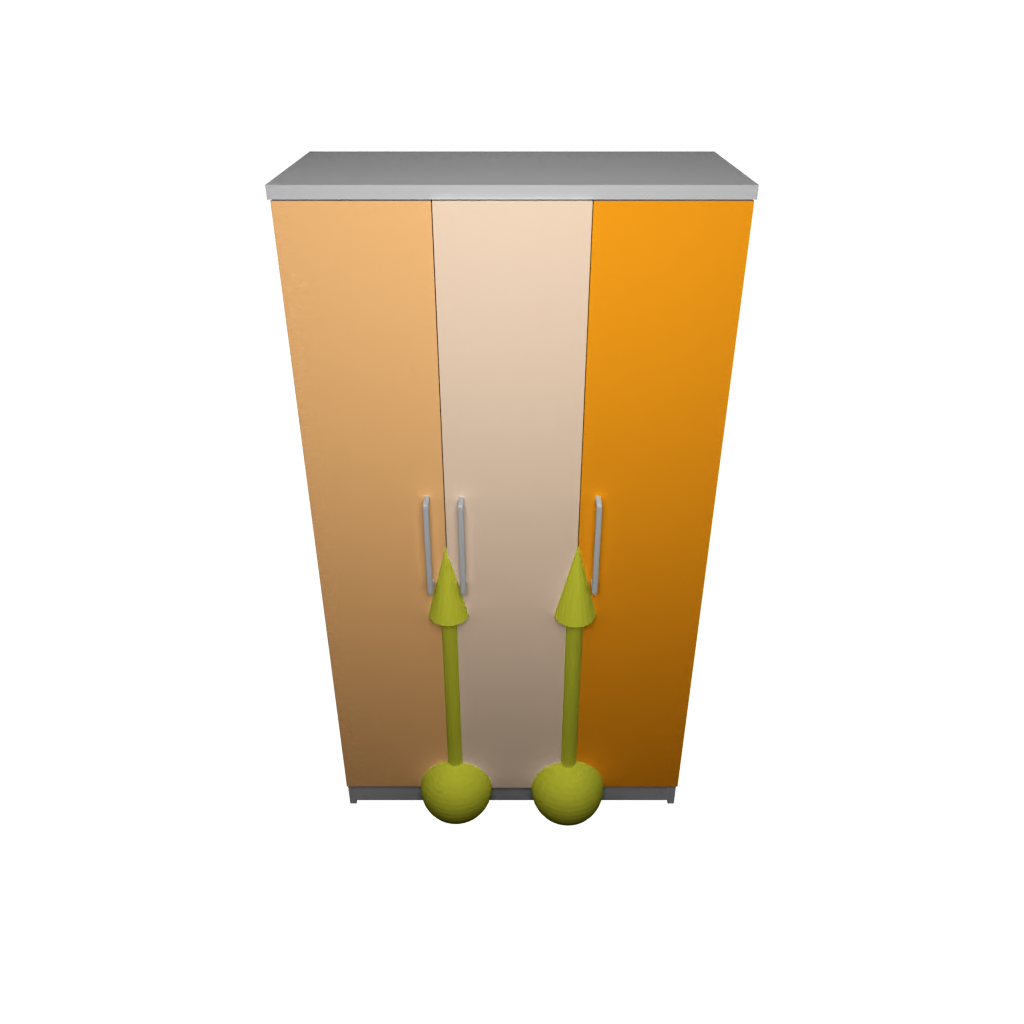}
& \includegraphics[trim=50 110 50 140,clip]{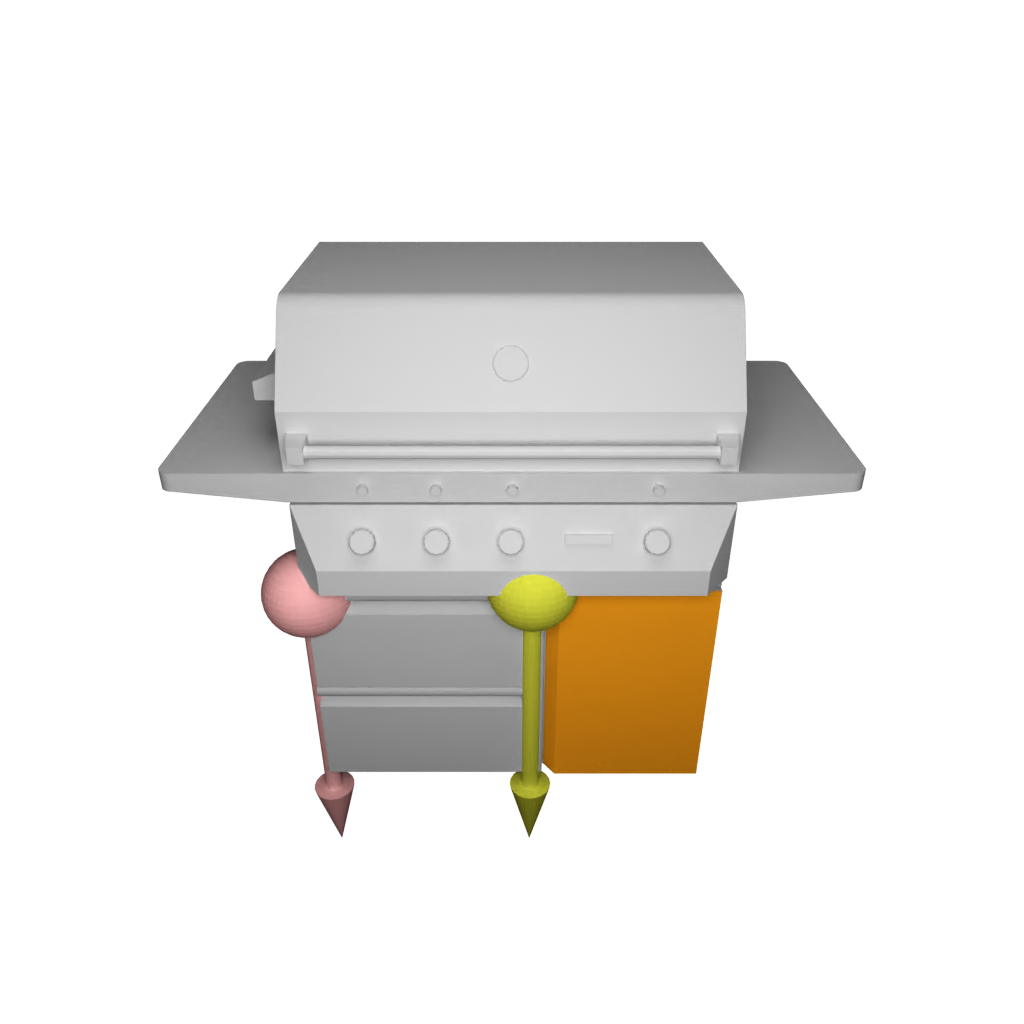}\\

\bottomrule
\end{tabularx}
\vspace{-5pt}
\caption{Segmentation and motion prediction results. For \maskthreed and \ogroup, motion prediction uses \heurmot while \gammagroup used the trained motion prediction heads.
Detected instances in shades of \blue{blue} for drawers, \orange{orange} for doors and \green{green} for lids.
For motion prediction, predicted axis in \green{green} for matched revolute joints, \purple{purple} for matched prismatic, and \pink{pink} for parts unmatched to any GT parts.
We find that \heurmot works well, but relies on high quality segmentation which \ogroup can provide for \pmopenshort. On \acd, all methods struggle to generalize, especially on drawers.
}
\vspace{-4pt}
\label{fig:qual-motion-combined}
\end{figure*}

\subsection{Part segmentation}
\label{sec:expr-mesh-segmentation}

We compare how well different segmentation methods detect and segment openable parts, focusing on mesh segmentation performance in the main paper.
In the supplement, we report results for evaluation directly on point clouds, and performance breakdown by part type.

\mypara{Metrics.}
We report the precision, recall and F1 scores based on matches of predicted to ground-truth openable parts at an IoU=0.5 threshold based on triangle area.
We follow the metric computation from prior work~\cite{jiang2022opd} to greedily match predicted parts to ground truth parts.
In the main paper, we report P/R/F1 macro-averaged over objects, which gives us an estimate of the percentage of complete objects that are well-segmented and can be articulated for use.
The supplement reports results micro-averaged over parts, and discusses limitations of other mesh segmentation metrics.

\begin{table}
\caption{
Evaluation of openable part mesh segmentation.
We report precision (P), recall (R), and F1 for openable parts (base part is excluded) at IoU=0.5.
Reported values are macro-averaged over objects.
See supplement for results including part micro-averages. $*$ denotes adding \pmopenext to the training split.
}
\vspace{-5pt}

\centering
{
\small
\begin{tabular}{@{} l rrr rrr @{}}
\toprule
& \multicolumn{3}{c}{\pmopen} & \multicolumn{3}{c}{\ourdatashort} \\
\cmidrule(l{0pt}r{2pt}){2-4} \cmidrule(l{0pt}r{2pt}){5-7}
Method & P & R & F1 & P & R & F1 \\
\midrule
\stom & 1.2 & 0.5 & 0.7 & 2.2 & 0.5 & 0.8\\
\pointgroup & 47.6 & 40.0 & 42.1 & 10.4 & 2.7 & 4.3\\
\maskthreed & 52.6 & 33.9 & 41.1 & 19.2 & 6.0 & 9.2\\
\midrule
\ogroup & \textbf{86.3} & \textbf{77.4} & \textbf{81.5} & 30.0 & 7.4 & 11.9\\
\ogroup$\!^{*}$ & 85.2 & 74.8 & 79.4 & \textbf{37.0} & \textbf{15.5} & \textbf{21.9}\\
\gammagroup & 77.3 & 66.9 & 71.6 & 24.5 & 4.5 & 7.5\\
\gammagroup$\!^{*}$ & 74.5 & 55.1 & 63.2 & 24.9 & 3.7 & 6.5\\
\bottomrule
\end{tabular}
}
\label{tab:results-mesh-precision-no-base-macro}
\vspace{-4mm}
\end{table}
\begin{table}
\caption{
Impact of training \ogroup with data from PM-Openable-Ext and ACD-train (in addition to PM-Openable).
Including additional data improves performance on more realistic \acd-val shapes significantly.
}
\vspace{-5pt}
\resizebox{\linewidth}{!}
{
\begin{tabular}{@{} l cc rrr rrr @{}}
\toprule
& \multicolumn{2}{c}{Data} & \multicolumn{3}{c}{\pmopen} & \multicolumn{3}{c}{\ourdatashort-val} \\
\cmidrule(l{0pt}r{2pt}){2-3} 
\cmidrule(l{0pt}r{2pt}){4-6} \cmidrule(l{0pt}r{2pt}){7-9}
Method & \pmopenextshort & ACD & P & R & F1 & P & R & F1 \\
\midrule
\ogroup &  \myxmark & \myxmark & 86.3 & 77.4 & 81.5 & 47.4 & 19.8 & 27.9 \\
\ogroup &  \checkmark & \myxmark & 85.2 & 74.8 & 79.4 & 47.1 & 28.0 & 34.8\\
\ogroup & \myxmark & \checkmark & \textbf{90.0} & \textbf{85.8} & \textbf{87.6} & 66.6 & 52.8 & 58.8 \\
\ogroup &  \checkmark & \checkmark & 75.4 & 69.2 & 71.8 & \textbf{75.3} & \textbf{58.5} & \textbf{65.7} \\
\midrule
\gammagroup &  \myxmark & \myxmark & 77.3 & 66.9 & 71.6 & 42.5 & 16.7 & 23.8 \\
\gammagroup &  \checkmark & \myxmark & 74.5 & 55.1 & 63.2 & 49.3 & 14.2 & 21.9\\
\gammagroup &  \myxmark & \checkmark & 72.4 & 60.4 & 65.6 & 62.6 & 44.2 & 51.8\\
\gammagroup &  \checkmark & \checkmark & 79.3 & 64.7 & 71.1 & 55.0 & 35.3 & 42.9\\
\bottomrule
\end{tabular}
}
\label{tab:results-mesh-precision-no-base-acdtrain}
\vspace{-4mm}
\end{table}

\mypara{Results.}
In \cref{tab:results-mesh-precision-no-base-macro}, we compare our proposed \ogroup against several prior point-clouded based segmentation methods: \stom~\cite{wang2019shape2motion}, \pointgroup~\cite{jiang2020pointgroup}, and \maskthreed~\cite{schult2023mask3d}. 
The results show the task is challenging overall, especially with our more diverse data (\acd).
\ogroup is the top performer in segmentation.
We find that \ogroup achieves very good performance on \pmopen and almost doubles the F1 score of \maskthreed and \pointgroup. While \gammagroup still outperforms the baselines, it lags behind \ogroup likely due to the additional motion prediction losses. Finally, \stom is much less stable and likely requires more heuristics to obtain better performance as it does not output instances natively.

\textit{Generalization to \acd.}
On \acd, we see a significant decrease across metrics, showing the challenge of our dataset and the inability of models trained on \pmopen to generalize. Shelf geometry behind door parts is typically present in \pmopen but mostly missing in \acd, explaining lower performance on doors. As \acd assets typically do not have complete interior geometry, the signal of a drawer box attached behind the front is lost. By using a mix of the original \pmopen and \pmopenext (that exhibits characteristics of ACD described above) for training we boost \ogroup's performance from 11.9 to 21.9 F1 points.
From \cref{fig:qual-motion-combined}, we see that segmentation methods trained only on \pmopen struggle to generalize on \acd. This happens mostly for drawers, as they lack the interior geometry present in the training data and for more complex shapes in general. Frequently, the regions with openable drawers are identified as an openable part.  However, models struggle to separate them into correct instances or incorrectly label them as \texttt{door}.

\textit{Can training with more varied data help generalization?}
In \cref{tab:results-mesh-precision-no-base-acdtrain}, we compare the impact of using more data for training by including PM-Openable-ext and ACD-train (3D-FUTURE) in training and evaluating on ACD-val (HSSD + ABO).
We find that adding \acd-train benefits performance on \pmopen, improving by 8.2 F1 points. 
A significant improvement is made on \acd-val, where previously the best performing model trained on the mix of \pmopen and \pmopenext achieves an improvement from 34.8 to 65.7 F1 score points when \acd-train is added to training. This shows that \acd is beneficial not only for benchmarking but also pushes the boundary of generalization. This finding may benefit the embodied AI community as it is related to the longstanding challenge of the sim-to-real gap. Qualitative results are in the supplement.

\begin{table}
\centering
\caption{
We ablate different \pointgroup backbones and the use of the FPN and over-segmentation voting (OSV) for \ogroup. 
}
\vspace{-5pt}
\resizebox{\linewidth}{!}
{
\begin{tabular}{@{} lccc rrr rrr @{}}
\toprule
& & & & \multicolumn{3}{c}{\pmopen} & \multicolumn{3}{c}{\ourdatashort} \\
\cmidrule(l{0pt}r{2pt}){5-7} \cmidrule(l{0pt}r{2pt}){8-10}
Backbone & FPN & OSV & \pmopenextshort & P & R & F1 & P & R & F1 \\
\midrule
\unet & \myxmark & \myxmark & \myxmark & 47.6 & 40.0 & 42.1 & 10.4 & 2.7 & 4.3\\
\swinthreed & \myxmark & \myxmark & \myxmark & 65.2 & 49.1 & 55.7 & 23.7 & 7.2 & 11.0\\
\pointnextshort & \myxmark & \myxmark & \myxmark & 72.8 & 53.5 & 61.2 & 9.5 & 3.1 & 4.7\\
\midrule
\pointnextshort & \checkmark & \myxmark & \myxmark & 86.3 & 77.4 & 81.5 & 30.0 & 7.4 & 11.9\\
\pointnextshort & \checkmark & \checkmark & \myxmark & \textbf{92.2} & \textbf{83.0} & \textbf{87.2} & 30.0 & 7.0 & 11.3\\
\pointnextshort & \checkmark & \myxmark & \checkmark & 85.2 & 74.8 & 79.4 & 34.8 & 15.1 & 21.0\\
\pointnextshort & \checkmark & \checkmark & \checkmark & 85.2 & 74.8 & 79.4 & \textbf{37.0} & \textbf{15.5} & \textbf{21.9}\\
\bottomrule
\end{tabular}
}
\label{tab:results-pg-backbone-ablate}
\vspace{-4mm}
\end{table}

\textit{Ablations.}
We ablate the choice of \pointnext as our backbone, and our design choices in \Cref{tab:results-pg-backbone-ablate}.  On \pmopen, we find that \pointnext gives the best performance, while \swinthreed performs better for \acd.  
By replacing the MLP with FPN for feature processing, we are able to considerably improve the performance to $81.5$ F1 on \pmopen and $11.9$ F1 on ACD.
By propagating to the over-segmentation (OSV) instead of per-triangle voting, we further improve performance.
Finally, incorporating \pmopenext enables better generalization on \acd.

\subsection{Motion prediction}
\label{sec:expr-motion}

\mypara{Metrics.}
For motion prediction, we follow \citet{jiang2022opd} and report part-averaged mAP for matching motion type, axis and origin (M, MA, and MAO), as well as axis and origin errors (AE and OE).
For computing metrics with predicted segmentation, we need to match the predicted parts to the ground truth parts.
We use a similar matching algorithm when evaluating the part segmentations.
For the matched pairs, we calculate the axis and origin error, the accuracy of the motion type, motion axis (within 5 degrees), and MAO (within 0.1 times the diagonal length of the openable part).

\begin{table}[t]
\centering
\caption{Motion prediction evaluation on \pmopen.
Learned baselines such as \stom struggle to predict part motion, partially due to segmentation. Our \gammagroup produces meaningful part motions but underperforms compared to \ogroup with heuristic.
}
\vspace{-5pt}
\resizebox{\linewidth}{!}
{
\begin{tabular}{@{} llc rrr rr @{}}
\toprule 
& & & \multicolumn{3}{c}{F1 \% $\uparrow$} & \multicolumn{2}{c}{Error $\downarrow$} \\
\cmidrule(l{0pt}r{2pt}){4-6} 
\cmidrule(l{0pt}r{2pt}){7-8}
Motion & Segmentation  & \# & {+}M & {+}MA & {+}MAO & AE & OE \\
\midrule
Learned & \stom & 3 & 0.9 & 0.9 & 0.9 & 30.0 & \textbf{0.06} \\
Learned & \gammagroup & 133 & 71.2 & 61.9 & 50.6 & 9.5 & 0.18 \\
Heur & \maskthreed & 78 & 39.2 & 34.1 & 27.1 & 9.2 & 0.44 \\
Heur  & \gammagroup & 133 & 71.1 & 62.0 & 57.5 & 10.8 & 0.12\\
Heur & \ogroup & \textbf{148} & \textbf{79.0} & \textbf{70.1} & \textbf{60.0} & \textbf{9.1} & 0.16\\
\midrule
Heur & GT & 182 & 94.7 & 88.4 & 84.5 & 6.43 & 0.07 \\
\bottomrule
\end{tabular}
}
\vspace{-4mm}
\label{tab:results-motion-main-f1}
\end{table}

\begin{table}[t]
\centering
\caption{Motion prediction evaluation on \acd.
Overall performance is low, showing the challenge of predicting articulations for realistic container objects as in \acd. However, we note that \gammagroup and \ogroup with heuristic are the best options.
}
\vspace{-5pt}
\resizebox{\linewidth}{!}
{
\begin{tabular}{@{} llc rrr rr @{}}
\toprule 
& & & \multicolumn{3}{c}{F1 \% $\uparrow$} & \multicolumn{2}{c}{Error $\downarrow$} \\
\cmidrule(l{0pt}r{2pt}){4-6} 
\cmidrule(l{0pt}r{2pt}){7-8}
Motion & Segmentation  & \# & {+}M & {+}MA & {+}MAO & AE & OE \\
\midrule
Learned & \stom & 10 & 1.2 & 1.2 & 0.8 & 9.0 & 0.24 \\
Learned & \gammagroup & 106 & 10.8 & 9.4 & 5.2 & \textbf{15.8} & 0.25 \\ 
Heur & \maskthreed & 66 & 6.9 & 6.1 & 3.2 & 27.3 & 0.58 \\
Heur  & \gammagroup & 106 & 10.9 & 7.7 & 6.0 & 28.5 & \textbf{0.23} \\
Heur & \ogroup & \textbf{151} & \textbf{17.8} & \textbf{11.6} & \textbf{7.6} & 27.5 & 0.26 \\
\midrule
Heur & GT & 1350 & 92.5 & 81.3 & 72.3 & 12.8 & 0.18 \\
\bottomrule
\end{tabular}
}
\vspace{-2mm}
\label{tab:results-motion-main-hssd-f1}
\end{table}

\begin{table}[t]
\caption{
Motion prediction with our heuristic baseline using ground truth part segmentation, broken down by openable part category.
These results show which part types are challenging even with GT part segmentation.
}
\vspace{-5pt}
\resizebox{\linewidth}{!}
{
\label{tab:results-motion-heur-gt-cat}
\begin{tabular}{@{} l rrr rrr rrr @{}}
\toprule
& \multicolumn{3}{c}{Drawer $\% \uparrow$} & \multicolumn{3}{c}{Door $\% \uparrow$} & \multicolumn{3}{c}{Lid $\% \uparrow$} \\
\cmidrule(l{0pt}r{2pt}){2-4} \cmidrule(l{2pt}r{0pt}){5-7} \cmidrule(l{2pt}r{0pt}){8-10}
Dataset & M & MA & MAO & M & MA & MAO & M & MA & MAO \\
\midrule
\pmopenshort & 100 & 100 & 100 & 93.6 & 84.0 & 74.5 & 100 & 100 & 88.9 \\
\ourdatashort & 100 & 92.2 & 92.2 & 88.1 & 66.5 & 30.0 & 100 & 83.3 & 75.0 \\
\bottomrule
\end{tabular}
}
\vspace{-4mm}
\end{table}

\mypara{Results.}
We compare our \heurmot against two learned methods (our \gammagroup and \stom~\cite{wang2019shape2motion}) on \pmopen  (\cref{tab:results-motion-main-f1}) and \acd (\cref{tab:results-motion-main-hssd-f1}).

\textit{How well does heuristic motion prediction work?}
To verify our heuristic motion prediction, we conduct experiments using ground truth segmentations.
\Cref{tab:results-motion-main-f1,tab:results-motion-main-hssd-f1} (bottom rows) show \heurmot works quite well.
We examine the performance for different part types in \cref{tab:results-motion-heur-gt-cat}. 
On \pmopen, \heurmot works perfectly for drawers and almost perfectly for lids, with the exception of axis origin, but has lower performance on doors.
Doors are more challenging as they exhibit various mobility arrangements.
We see a similar trend for \acd where \heurmot also works well for drawers and lids, but again less so for doors.

\textit{How does our heuristic compare to learned methods on predicted segmentations?}
In \cref{tab:results-motion-main-f1,tab:results-motion-main-hssd-f1}, we compare the performance of our heuristic-based motion prediction with different segmentations. With the best-performing segmentation method \ogroup, our \heurmot outperforms learned \stom across all metrics except origin error (OE), where \stom achieves slightly better performance.
However, the number of matched parts is very small for \stom. We note that \heurmot outperforms \gammagroup when given \ogroup segmentation. To disentangle the impact of better segmentation, we also compare using \gammagroup segmentation with our motion heuristic. Even in this setting, the heuristic outperforms the learned baseline. 
\cref{tab:results-motion-main-hssd-f1} shows that on \acd our heuristic still manages to generalize on par with \gammagroup. \gammagroup, however, achieves significantly lower axis error. Overall, performance is lower due to the dataset complexity.

Examples of motion prediction on \pmopen and \acd-val are in \cref{fig:qual-motion-combined}.
The quality of motion prediction depends largely on the accuracy of the detected parts.
The heuristic approach outperforms the learned predictions of \gammagroup.
For \ourdatashort, since identifying openable parts is challenging, the quality of motion predictions is also lower.
This is particularly the case for objects with many openable parts, and with complex arrangements of the openable parts (e.g., storage units with both drawers and doors, and bookcases with some sections having doors).

\mypara{Discussion.} It is surprising that our heuristic method can beat or match more complex and conceptually powerful learning-based methods.
We hypothesize that this trend would change once the data gets even more complex. With more data, learned methods such as \gammagroup may start to benefit from data scale. Our heuristic is useful when data is limited, the target data is simple, or as a strong baseline.
This highlights avenues for future work: development of better segmentation or motion prediction methods, and construction of larger articulated 3D object datasets.

\begin{figure}
\vspace{-2mm}
\centering
\setkeys{Gin}{width=\linewidth}
\begin{tabularx}{\linewidth}{Y Y Y}
\includegraphics{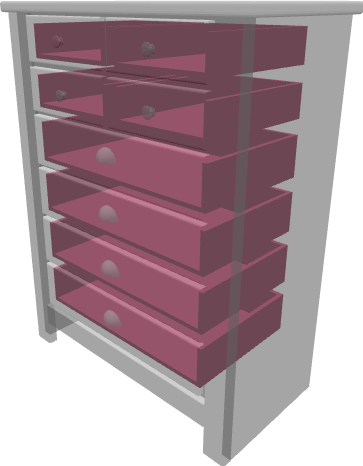} &
\includegraphics{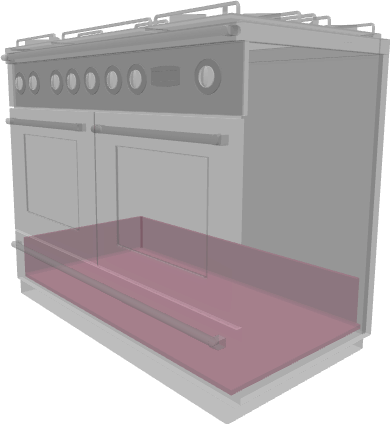} &
\includegraphics{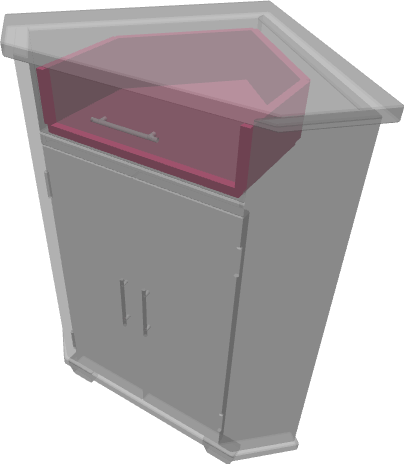} \\
\includegraphics{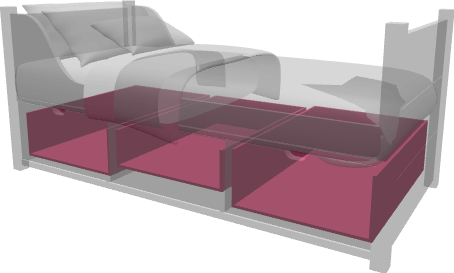} &
\includegraphics{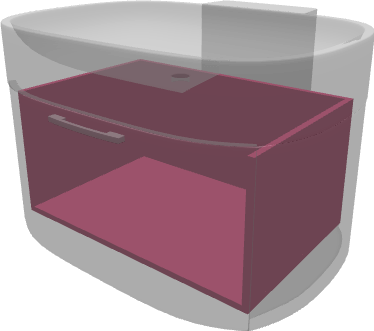} &
\includegraphics{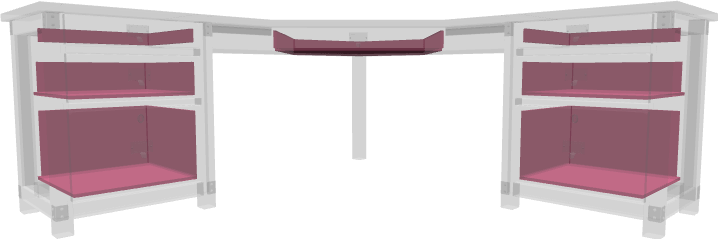} 
\end{tabularx}
\vspace{-10pt}
\caption{Interior geometry completions for drawers in various \acd objects.
From left to right and top to bottom: dresser, oven, corner cabinet, bed, sink, L-shaped desk.
The interiors of diverse openable parts are realistically completed using our approach.
}
\label{fig:interior-completion-heur}
\vspace{-4mm}
\end{figure}
\begin{figure}[t]
\includegraphics[width=\linewidth]{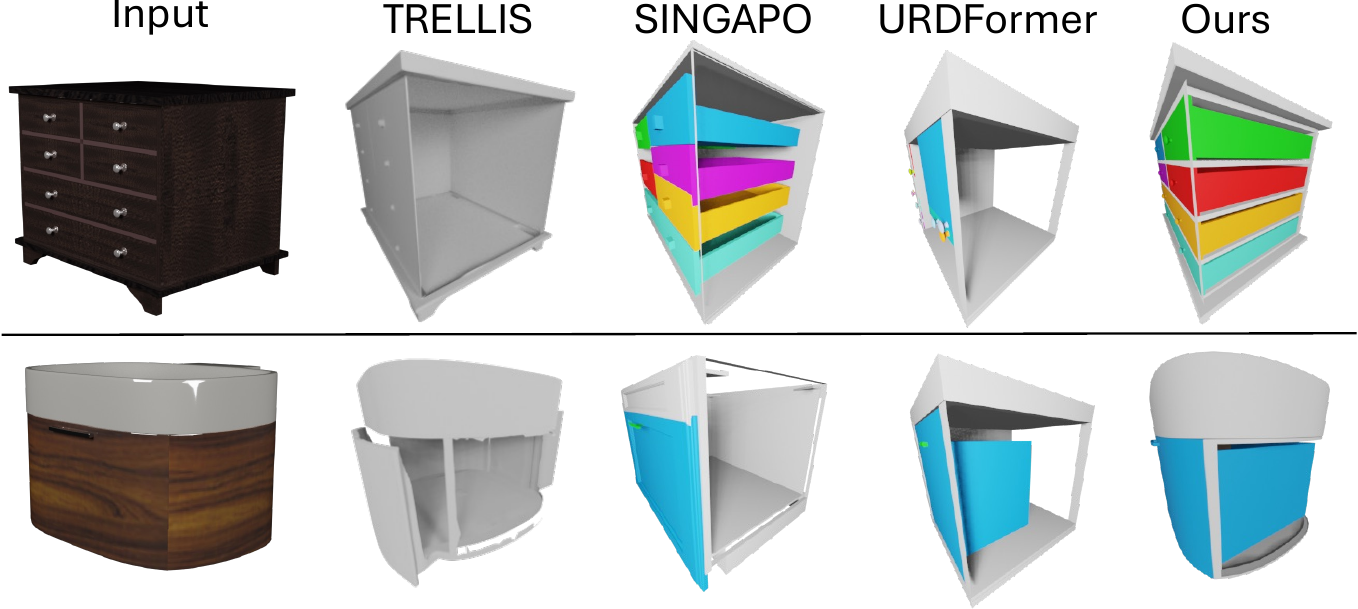}
\caption{Comparison to SOTA generative models (cutouts show interiors). SINGAPO and URDFormer produce some interiors, but miss parts and change the input object. TRELLIS ~\cite{xiang2024trellis} does not reconstruct interiors even if the input is modified with an open drawer (bottom). Our approach preserves input geometry (e.g., round sink cabinet front) and produces complete interiors.}
\label{fig:interior-completion}
\end{figure}

\subsection{Interior completion}

We compare our completion algorithm against recent methods: image-conditioned generation for static 3D meshes (TRELLIS~\cite{xiang2024trellis}) and for articulated objects (SINGAPO~\cite{liu2025singapo}), and an articulated object reconstruction (URDFormer~\cite{chen2023urdformer}).
In \cref{fig:interior-completion-heur}, we show examples of drawers completed using our approach on \acd shapes, where we handle a variety of object categories and part geometries. 

We provide a quantitative evaluation in the supplement.  In \cref{fig:interior-completion}, we show comparisons of output from our heuristic drawer completion. 
We find that the retrieval-based nature of SINGAPO becomes a weakness when applied to out-of-distribution data as the input geometry is not respected.

\textit{Application.} We apply our interior geometry completion to our annotated \acd objects and create a set of articulated containers that can be used in downstream applications (see \cref{fig:ood-supp} in supplement). Completion of the base part (e.g., oven interior) is a promising extension, but does not influence the kinematic plausibility of the shape.

\section{Conclusion}
We proposed the \ourtasklong (\ourtask) task: enhancing static 3D meshes with openable parts in an end-to-end framework.
We curated two datasets and systematically evaluated methods to tackle the task in our framework.
Our experiments highlight the deficiencies of existing datasets and methods, and suggest open challenges for future work.
Finally, we find that our pipeline generalizes to shapes outside of the datasets we evaluate on, and demonstrates automated annotation capabilities, as shown in the supplement.

\mypara{Limitations.}
We focused solely on openable parts.
However, real objects possess a broader range of articulated parts.
Generalization of the task and methods to a broader set of motions is an interesting direction.
Our work was also limited by the sparsity of available datasets with part and motion annotations, as well as biases due to lack of diversity of objects.
Extension of our work by curating larger openable 3D object datasets can improve the predictiveness of benchmarking to practical settings.

Despite these limitations, we believe our work establishes a unified framework and systematic benchmark for future work on large-scale articulated 3D object datasets.

\mypara{Acknowledgments.}
This work was funded in part by a CIFAR AI Chair, a Canada Research Chair, NSERC Discovery Grant, and enabled by support from the \href{https://alliancecan.ca/}{Digital Research Alliance of Canada}.
We would like to thank Hou In Derek Pun, Isabelle Kwan, Hou In Ivan Tam, Jiayi Liu, Kian Hosseinkhani, Mrinal Goshalia, Samuel Antunes Miranda and Xingguang Yan for their help annotating the data.
We also thank Armin Kavian, Austin T. Wang, Han-Hung Lee, Jiayi Liu, Qirui Wu and ZeMing Gong for helpful discussions and feedback.

{
\small
\bibliographystyle{ieeenat_fullname}
\bibliography{main}
}

\clearpage
\setcounter{page}{1}
\maketitlesupplementary
\appendix

In this supplement, we provide additional details and analysis for the two datasets we use in our experiments (\Cref{sec:dataset-details}), method details (\Cref{sec:supp-segmentation-details,sec:supp-motion-details}), and additional qualitative and quantitative results (\Cref{sec:additional-results}). See our paper summary video and additional results video (corresponding to \cref{sec:supp-exp-acd-train}) for more details and results.

\section{Dataset Details}
\label{sec:dataset-details}

We compare our new \ourdata (\ourdatashort) to \pmopen (\cref{sec:supp-data-acd}), provide information on how we split ACD into train and val (\cref{sec:supp-exp-acd-train}), show T-SNE visualization of similarity of shapes for \pmopen across the train/val/test splits (\cref{sec:supp-data-pm}), and provide details on the construction of \pmopenext (\cref{sec:supp-data-pmext}).

\begin{table}
\centering
\caption{Comparison of \pmopen and \acd, with the number of objects and average number of openable parts for each object type.  
The colors highlight differences between the categories in \pmopen and \ourdata.  We show object categories within the broad category \storage{StorageFurniture} (light blue), categories within the broad category \tablecat{Table} (dark blue), and lastly categories that are \newcat{newly introduced} (green) in \ourdata to add more diversity.
Note that the PM-Openable categories for \storage{StorageFurniture} and \tablecat{Table} encompass several more fine-grained categories such as bookcases and wardrobes.  Coarse category totals in italics.
}
\resizebox{\linewidth}{!}
{
\begin{tabular}{@{} ll rr rr @{}}
\toprule
 & & \multicolumn{2}{c}{\pmopen} & \multicolumn{2}{c}{\acd} \\
\cmidrule(l{0pt}r{2pt}){3-4} \cmidrule(l{0pt}r{2pt}){5-6} 
coarse category & category & obj & part/obj & obj & part/obj \\ 
\midrule
\storage{StorageFurniture} & & \textit{335} & 2.4 & \textit{72} & \textit{3.1} \\
& \storage{CabinetUnit} & 233 & 2.1 & 10 & 2.0 \\
& \storage{Cabinet} & 83 & 3.1 & 39 & 3.9 \\
& \storage{Wardrobe} & 8 & 3.3 & 66 & 4.0 \\
& \storage{ChestOfDrawers} & 4 & 6.0 & 44 & 5.8 \\
& \newcat{Sideboard} &  &  & 24 & 4.1 \\ 
& \storage{Bookcase} & 5 & 2.0 & 16 & 5.3 \\ 
& \newcat{TvStand} & & & 17 & 4.2 \\
& \newcat{WallUnit} & & & 8 & 8.2 \\
& \newcat{SinkCabinet} & & & 11 & 2.5 \\
& \storage{StorageBench} & 2 & 2.0 & 10 & 2.3 \\
\tablecat{Table} & & \textit{77} & \textit{2.6} & \textit{72} & \textit{3.1} \\ 
& \tablecat{Table} & 31 & 2.1 & 13 & 3.8 \\
& \tablecat{Sidetable} & 12 & 2.7 & 1 & 1.0 \\
& \tablecat{Desk} & 31 & 3.1 & 22 & 4.7 \\
& \tablecat{Nightstand} & 3 & 1.3 & 36 & 1.9 \\
\newcat{Bed} & & & & 6 & 2.8 \\
Appliance & & \textit{137} & \textit{1.3} & \textit{31} & \textit{1.7} \\ 
& Refrigerator & 42 & 1.62 & 5 & 2.5 \\
 & Dishwasher & 41 & 1 & 2 & 1 \\
 & Oven & 24 & 1.5 & 9 & 2.3 \\ 
 & WashingMachine & 17 & 1 & 8 & 1 \\
 & Microwave & 13 & 1 & 6 & 1.5 \\
Other & & \textit{99} & \textit{1.1} & \textit{11} & \textit{1.6} \\
 & \newcat{Barbecue} & & & 3 & 3.3 \\
 & Safe & 30 & 1 & 3 & 1 \\
 & Trashcan & 69 & 1.1 & 5 & 1.0 \\
 \midrule
All & & \textbf{648} & \textbf{2.0} & \textbf{354} & \textbf{3.8} \\
\bottomrule
\end{tabular}
}
\vspace{-2mm}
\label{tab:supp-obj-categories}
\end{table}

\subsection{\acd statistics}
\label{sec:supp-data-acd}

\begin{figure*}
\centering
\setkeys{Gin}{width=\linewidth}
\begin{tabularx}{\textwidth}{Y Y Y Y Y Y Y Y Y Y}
\toprule
\includegraphics[trim=0 0 0 0,clip,width=\linewidth]{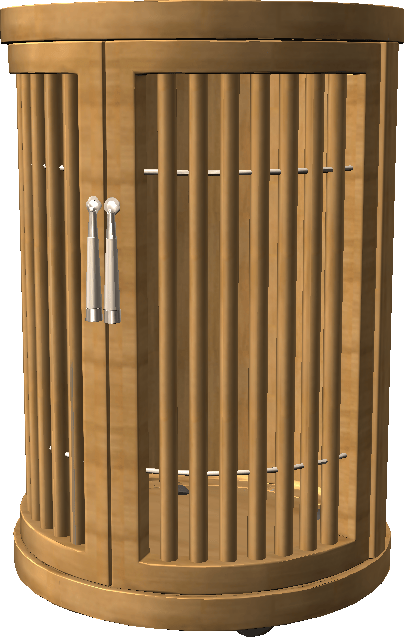} & 
\includegraphics[trim=0 0 0 0,clip,width=\linewidth]{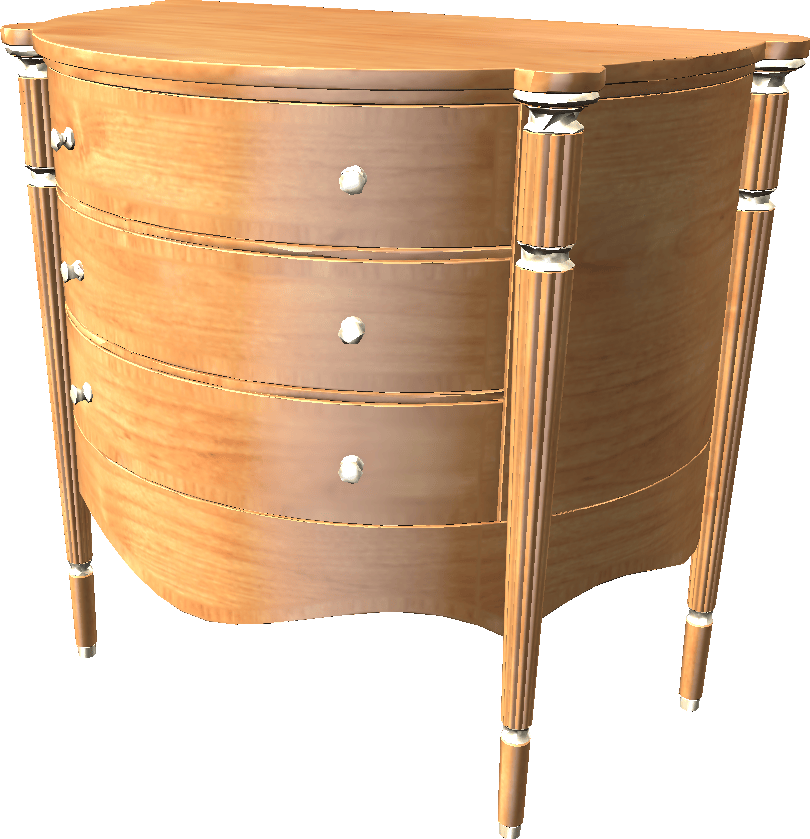} & 
\includegraphics[trim=0 0 0 0,clip,width=\linewidth]{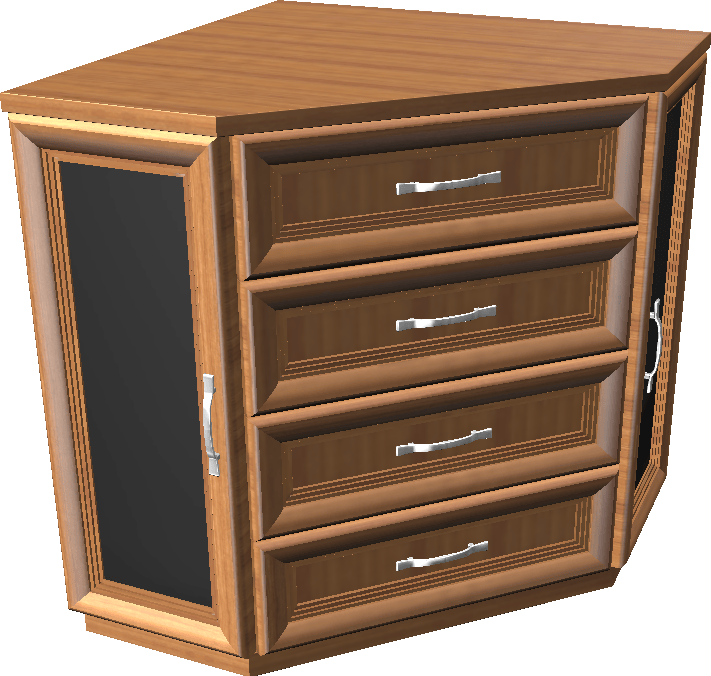} & 
\includegraphics[trim=0 0 0 0,clip,width=\linewidth]{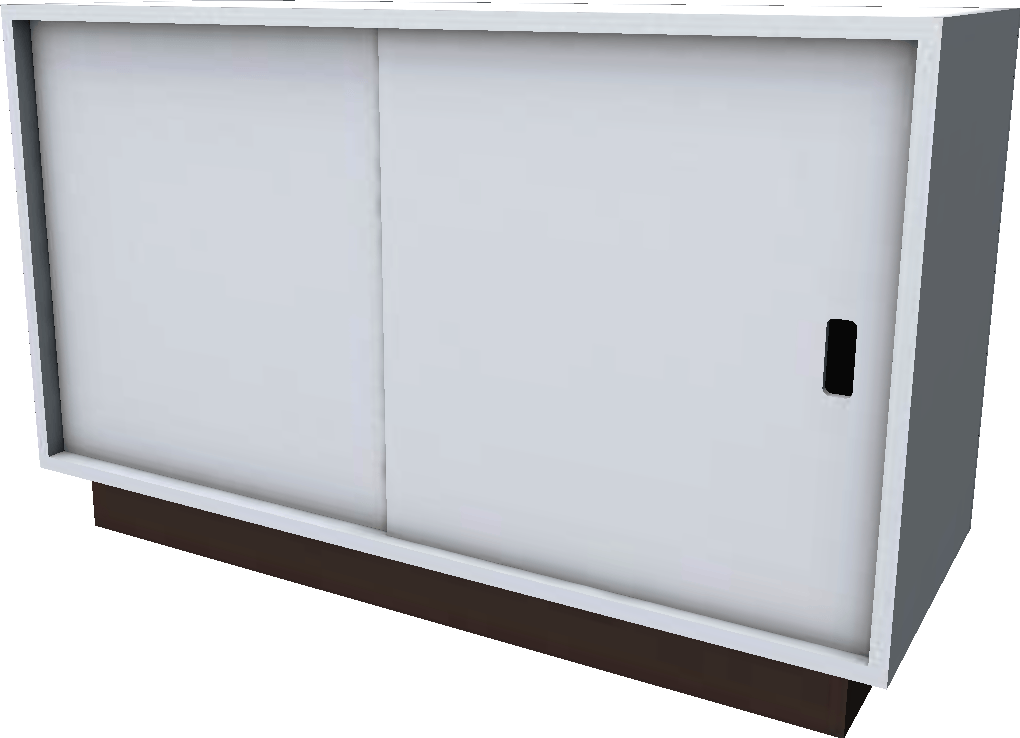} & 
\includegraphics[trim=0 0 0 0,clip,width=\linewidth]{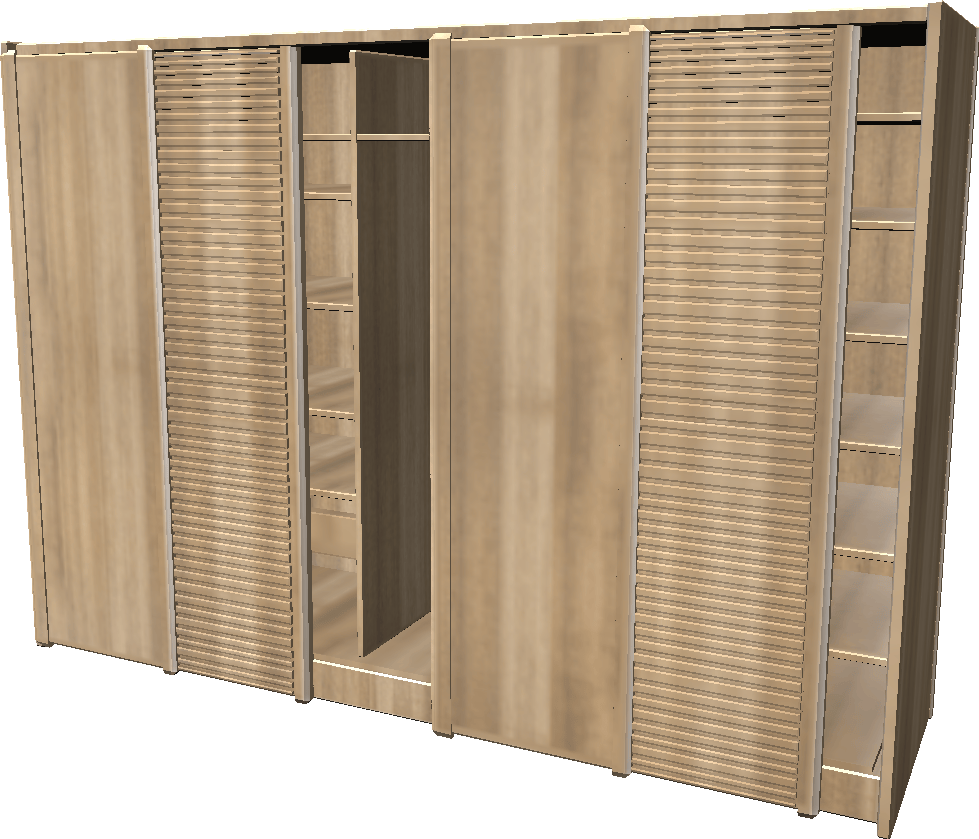} & 
\includegraphics[trim=0 0 0 0,clip,width=\linewidth]{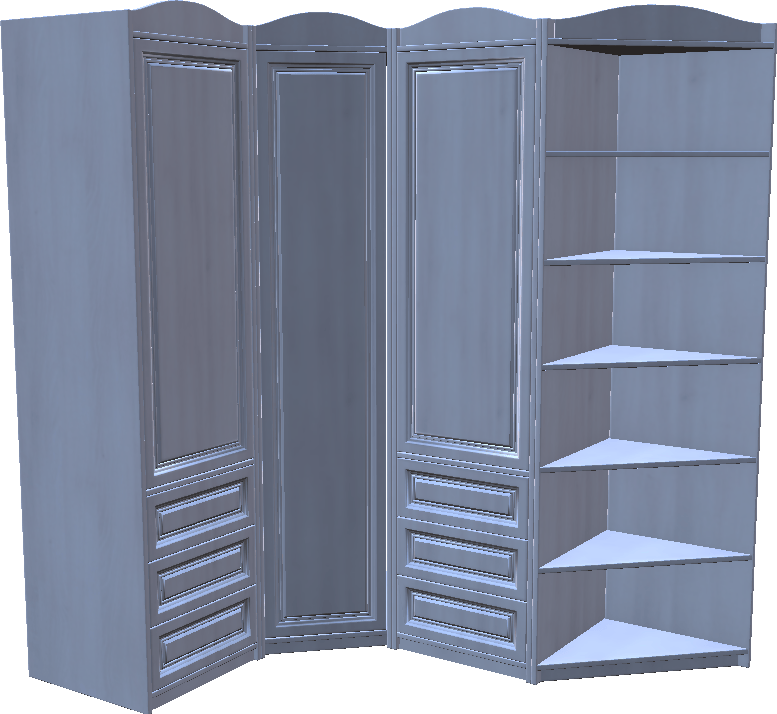} & 
\includegraphics[trim=0 0 0 0,clip,width=\linewidth]{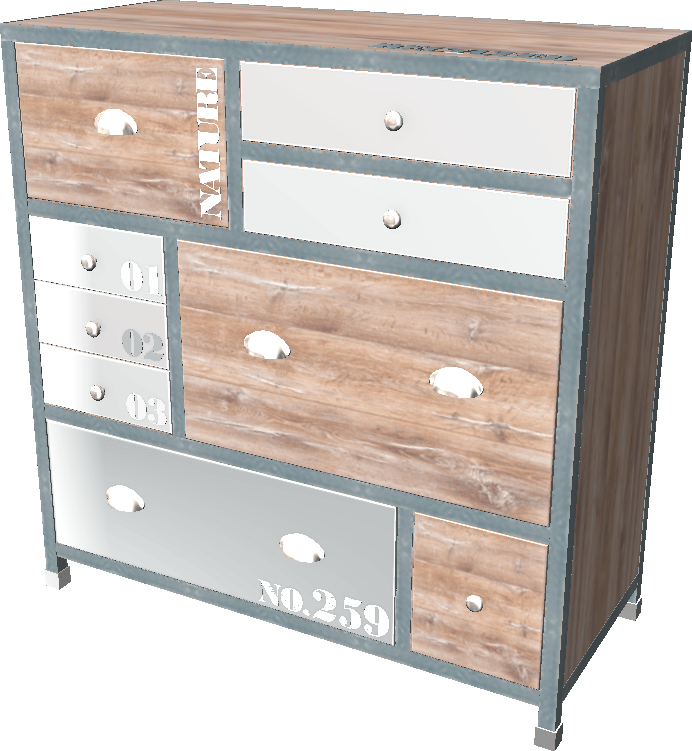} & 
\includegraphics[trim=0 0 0 0,clip,width=\linewidth]{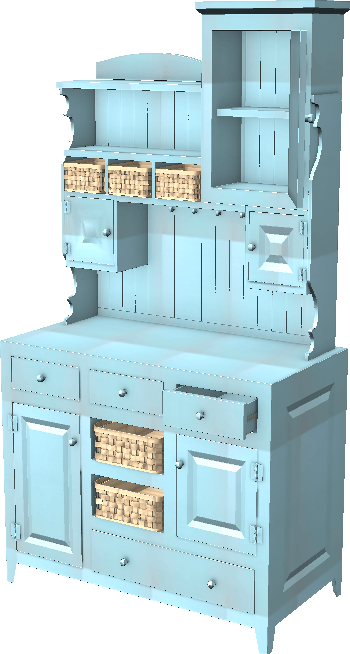} & 
\includegraphics[trim=0 0 0 0,clip,width=\linewidth]{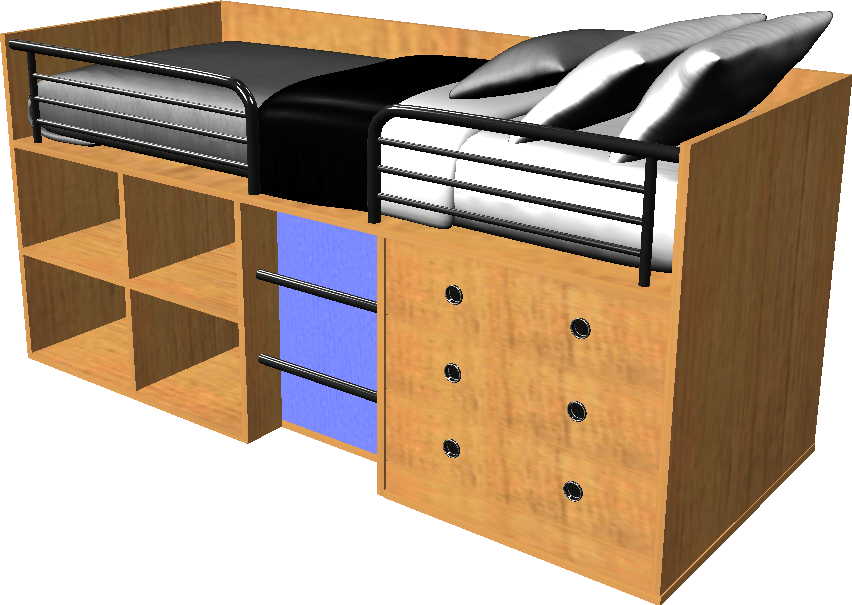} \\
\includegraphics[trim=0 70 0 9,clip,width=\linewidth]{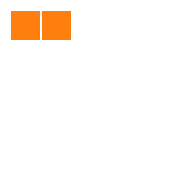} &
\includegraphics[trim=0 70 0 9,clip,width=\linewidth]{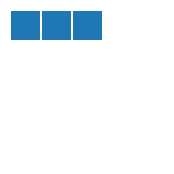} &
\includegraphics[trim=0 70 0 9,clip,width=\linewidth]{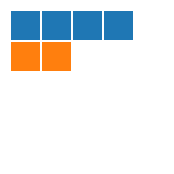} &
\includegraphics[trim=0 70 0 9,clip,width=\linewidth]{figure/icons/parts/0_2_0.png} &
\includegraphics[trim=0 70 0 9,clip,width=\linewidth]{figure/icons/parts/0_2_0.png} &
\includegraphics[trim=0 70 0 9,clip,width=\linewidth]{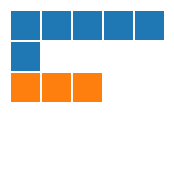} &
\includegraphics[trim=0 70 0 9,clip,width=\linewidth]{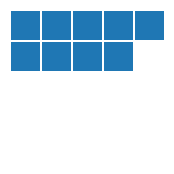} &
\includegraphics[trim=0 70 0 9,clip,width=\linewidth]{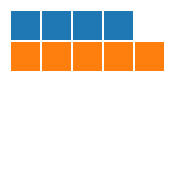} &
\includegraphics[trim=0 70 0 9,clip,width=\linewidth]{figure/icons/parts/3_0_0.png} 
\\
a & b & c & d & e & f & g & h & i 
\\
\bottomrule
\end{tabularx}
\vspace{-0.5em}
\caption
{Examples from \ourdatashort showing a diversity of part and object shapes -- round doors (a), curved drawers (b), corner cabinet (c), different motion types -- translational doors (d,e), large objects (e,f), complex part arrangements (f,g,h), and openable parts in non-standard objects -- drawers in beds (i).
For each object, we also indicate the number of openable drawers (\textcolor{tblblue}{blue}) and doors (\textcolor{tblorange}{orange}).
}
\label{fig:supp-acd-examples}
\end{figure*}

\begin{figure*}
\centering
\includegraphics[width=\linewidth]{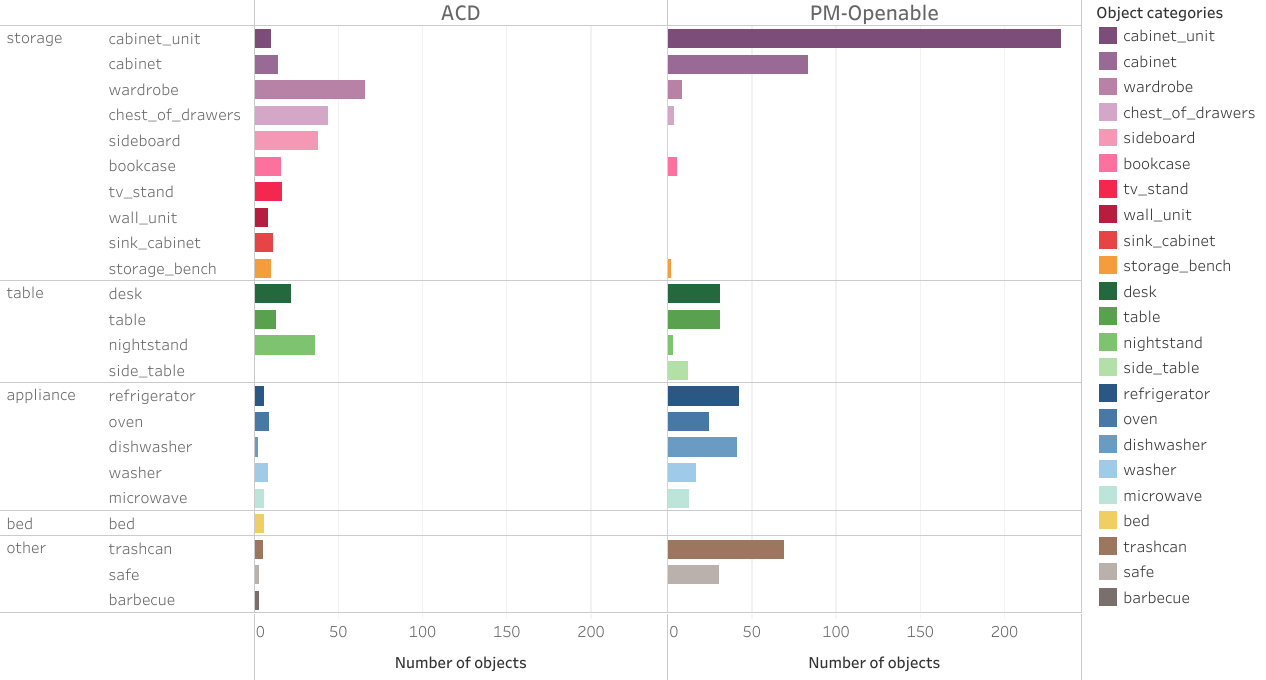}
\vspace{-1.5em}
\caption{
Comparison of the distribution of objects categories in our \ourdata (\acd) vs \pmopen. 
}
\label{fig:supp-object-compare}
\end{figure*}

We show examples from \ourdatashort in \cref{fig:supp-acd-examples}.
In \cref{tab:supp-obj-categories} and \cref{fig:supp-object-compare}, we provide statistics of the object categories.  Note that \pmopen provides only coarse level categories of StorageFurniture and Table. To obtain finer object categories, we map the PartNetMobility~\cite{xiang2020sapien} asset ids to original ShapeNet~\cite{chang2015shapenet} asset ids to determine the finer classification in \cref{tab:supp-obj-categories} and \cref{fig:supp-object-compare}. 

\begin{figure*}
\centering
\includegraphics[width=\linewidth]{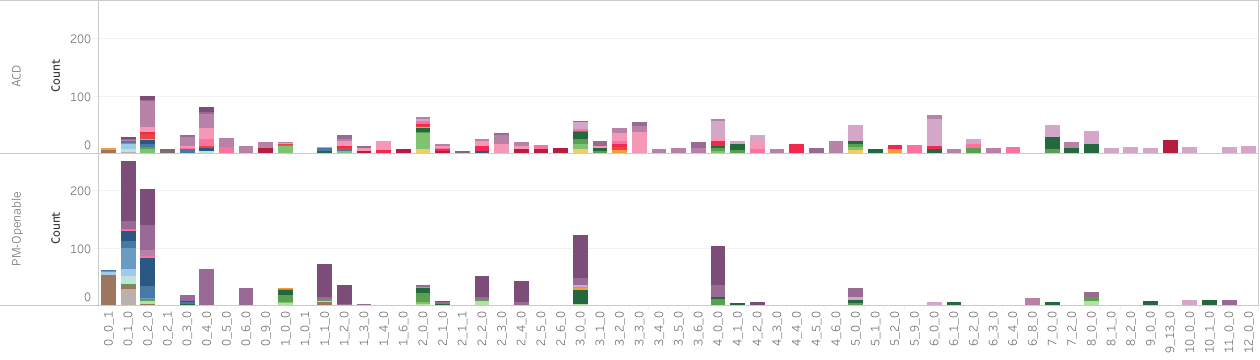}
\vspace{-1.5em}
\caption{
Comparison of the distribution of \textbf{part configuration} (drawers, doors, lids) of objects categories in our \ourdata (\acd) vs \pmopen.  For each object, we compute a \textit{part code} that indicates the number of drawers, doors, and lids.  For instance, \texttt{2\_1\_0} indicates the object has 2 drawers, 1 door, and 0 lids.  In this figure, we plot the number of objects with each different part configuration colored by their object category. Storage furniture is in purple/pink/red, appliances in blue, and tables in greens (\cref{fig:supp-object-compare} for full legend).  While \pmopen is dominated by objects with just a few openable parts (often just drawers or doors), \acd shows a broader distribution of openable part configurations.  
}
\label{fig:supp-part-structure-distribution}
\end{figure*}

\begin{figure*}
\centering
\includegraphics[trim=0 0 0 0,clip,width=\linewidth]{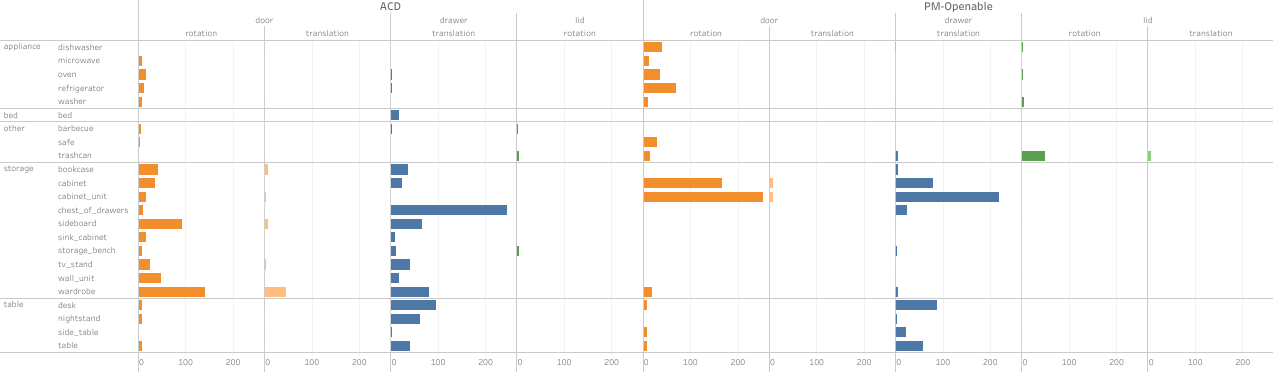}
\vspace{-1.5em}
\caption{
Comparison of the distribution of part \textbf{motion type} in objects categories in our \ourdata (\acd) vs \pmopen.  \acd has more diverse motion types for doors (more doors that translate) across a broader set of object categories.   
}
\label{fig:supp-part-motion}
\end{figure*}

\begin{figure*}
\centering
\setkeys{Gin}{width=\linewidth}
\begin{tabularx}{0.9\textwidth}{@{} p{1.5cm} Y Y Y Y Y | Y Y Y Y Y }
\toprule
& \multicolumn{5}{c}{\ourdatashort} & \multicolumn{5}{c}{\pmopen} \\
\toprule
\multirow{11}{*}{storage}  & \shortstack{4\\\includegraphics[trim=0 90 0 9,clip,width=\linewidth]{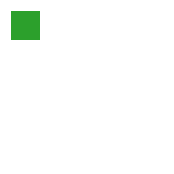}} & \shortstack{8\\\includegraphics[trim=0 90 0 9,clip,width=\linewidth]{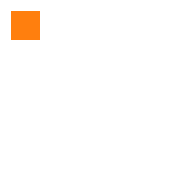}} & \shortstack{39\\\includegraphics[trim=0 90 0 9,clip,width=\linewidth]{figure/icons/parts/0_2_0.png}} & \shortstack{8\\\includegraphics[trim=0 90 0 9,clip,width=\linewidth]{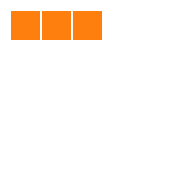}} & \shortstack{18\\\includegraphics[trim=0 90 0 9,clip,width=\linewidth]{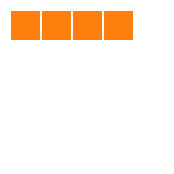}} & \shortstack{121\\\includegraphics[trim=0 90 0 9,clip,width=\linewidth]{figure/icons/parts/0_1_0.png}} & \shortstack{60\\\includegraphics[trim=0 90 0 9,clip,width=\linewidth]{figure/icons/parts/0_2_0.png}} & \shortstack{3\\\includegraphics[trim=0 90 0 9,clip,width=\linewidth]{figure/icons/parts/0_3_0.png}} & \shortstack{16\\\includegraphics[trim=0 90 0 9,clip,width=\linewidth]{figure/icons/parts/0_4_0.png}} & \shortstack{5\\\includegraphics[trim=0 90 0 9,clip,width=\linewidth]{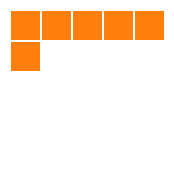}} \\ 
 & \shortstack{5\\\includegraphics[trim=0 90 0 9,clip,width=\linewidth]{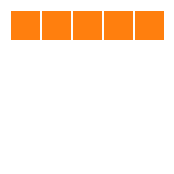}} & \shortstack{2\\\includegraphics[trim=0 90 0 9,clip,width=\linewidth]{figure/icons/parts/0_6_0.png}} & \shortstack{2\\\includegraphics[trim=0 90 0 9,clip,width=\linewidth]{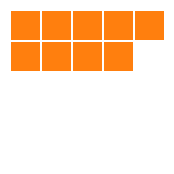}} & \shortstack{4\\\includegraphics[trim=0 90 0 9,clip,width=\linewidth]{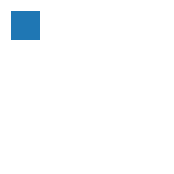}} & \shortstack{1\\\includegraphics[trim=0 90 0 9,clip,width=\linewidth]{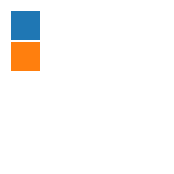}} & \shortstack{1\\\includegraphics[trim=0 90 0 9,clip,width=\linewidth]{figure/icons/parts/1_0_0.png}} & \shortstack{30\\\includegraphics[trim=0 90 0 9,clip,width=\linewidth]{figure/icons/parts/1_1_0.png}} & \shortstack{12\\\includegraphics[trim=0 90 0 9,clip,width=\linewidth]{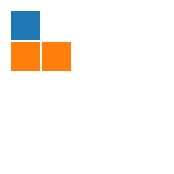}} & \shortstack{1\\\includegraphics[trim=0 90 0 9,clip,width=\linewidth]{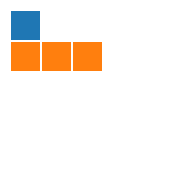}} & \shortstack{3\\\includegraphics[trim=0 90 0 9,clip,width=\linewidth]{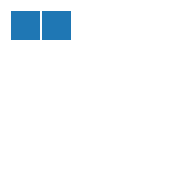}} \\ 

 & \shortstack{8\\\includegraphics[trim=0 70 0 9,clip,width=\linewidth]{figure/icons/parts/1_2_0.png}} & \shortstack{3\\\includegraphics[trim=0 70 0 9,clip,width=\linewidth]{figure/icons/parts/1_3_0.png}} & \shortstack{4\\\includegraphics[trim=0 70 0 9,clip,width=\linewidth]{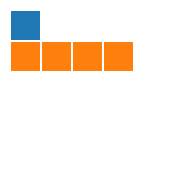}} & \shortstack{1\\\includegraphics[trim=0 70 0 9,clip,width=\linewidth]{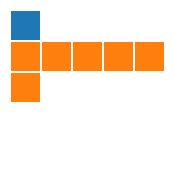}} & \shortstack{10\\\includegraphics[trim=0 70 0 9,clip,width=\linewidth]{figure/icons/parts/2_0_0.png}} & \shortstack{2\\\includegraphics[trim=0 70 0 9,clip,width=\linewidth]{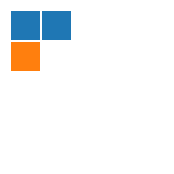}} & \shortstack{10\\\includegraphics[trim=0 70 0 9,clip,width=\linewidth]{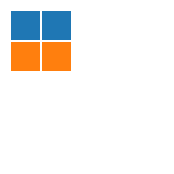}} & \shortstack{7\\\includegraphics[trim=0 70 0 9,clip,width=\linewidth]{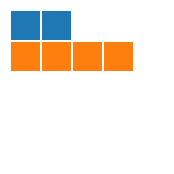}} & \shortstack{32\\\includegraphics[trim=0 70 0 9,clip,width=\linewidth]{figure/icons/parts/3_0_0.png}} & \shortstack{22\\\includegraphics[trim=0 70 0 9,clip,width=\linewidth]{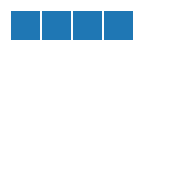}} \\ 

 & \shortstack{5\\\includegraphics[trim=0 40 0 9,clip,width=\linewidth]{figure/icons/parts/2_1_0.png}} & \shortstack{5\\\includegraphics[trim=0 40 0 9,clip,width=\linewidth]{figure/icons/parts/2_2_0.png}} & \shortstack{7\\\includegraphics[trim=0 40 0 9,clip,width=\linewidth]{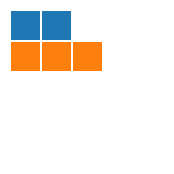}} & \shortstack{3\\\includegraphics[trim=0 40 0 9,clip,width=\linewidth]{figure/icons/parts/2_4_0.png}} & \shortstack{2\\\includegraphics[trim=0 40 0 9,clip,width=\linewidth]{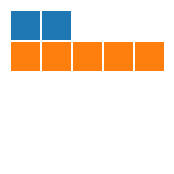}} & \shortstack{1\\\includegraphics[trim=0 40 0 9,clip,width=\linewidth]{figure/icons/parts/4_2_0.png}} & \shortstack{4\\\includegraphics[trim=0 40 0 9,clip,width=\linewidth]{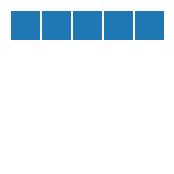}} & \shortstack{1\\\includegraphics[trim=0 40 0 9,clip,width=\linewidth]{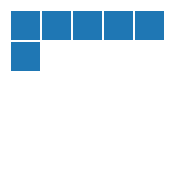}} & \shortstack{1\\\includegraphics[trim=0 40 0 9,clip,width=\linewidth]{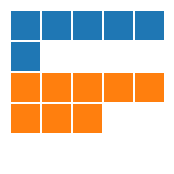}} & \shortstack{1\\\includegraphics[trim=0 40 0 9,clip,width=\linewidth]{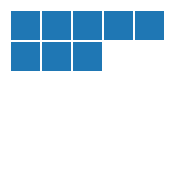}} \\ 
 & \shortstack{1\\\includegraphics[trim=0 50 0 9,clip,width=\linewidth]{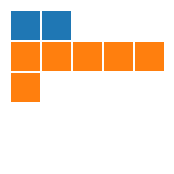}} & \shortstack{7\\\includegraphics[trim=0 50 0 9,clip,width=\linewidth]{figure/icons/parts/3_0_0.png}} & \shortstack{3\\\includegraphics[trim=0 50 0 9,clip,width=\linewidth]{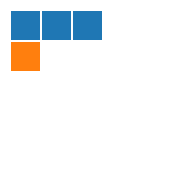}} & \shortstack{9\\\includegraphics[trim=0 50 0 9,clip,width=\linewidth]{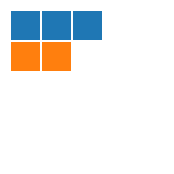}} & \shortstack{9\\\includegraphics[trim=0 50 0 9,clip,width=\linewidth]{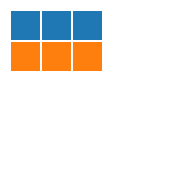}} & \shortstack{1\\\includegraphics[trim=0 50 0 9,clip,width=\linewidth]{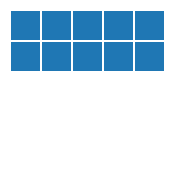}} & \shortstack{1\\\includegraphics[trim=0 50 0 9,clip,width=\linewidth]{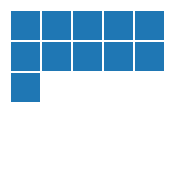}} &  &  &  \\ 
 & \shortstack{1\\\includegraphics[trim=0 70 0 9,clip,width=\linewidth]{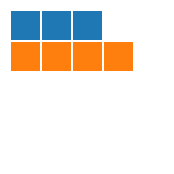}} & \shortstack{1\\\includegraphics[trim=0 70 0 9,clip,width=\linewidth]{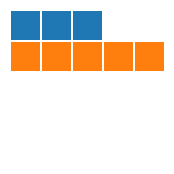}} & \shortstack{2\\\includegraphics[trim=0 70 0 9,clip,width=\linewidth]{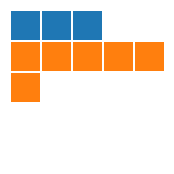}} & \shortstack{12\\\includegraphics[trim=0 70 0 9,clip,width=\linewidth]{figure/icons/parts/4_0_0.png}} & \shortstack{1\\\includegraphics[trim=0 70 0 9,clip,width=\linewidth]{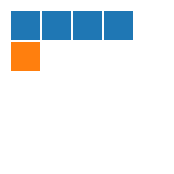}} &  &  &  &  &  \\ 
 & \shortstack{5\\\includegraphics[trim=0 70 0 9,clip,width=\linewidth]{figure/icons/parts/4_2_0.png}} & \shortstack{1\\\includegraphics[trim=0 70 0 9,clip,width=\linewidth]{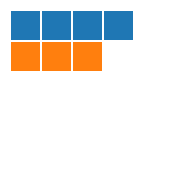}} & \shortstack{2\\\includegraphics[trim=0 70 0 9,clip,width=\linewidth]{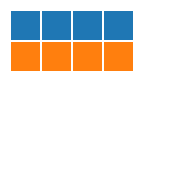}} & \shortstack{1\\\includegraphics[trim=0 70 0 9,clip,width=\linewidth]{figure/icons/parts/4_5_0.png}} & \shortstack{2\\\includegraphics[trim=0 70 0 9,clip,width=\linewidth]{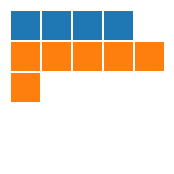}} &  &  &  &  &  \\ 
 & \shortstack{6\\\includegraphics[trim=0 70 0 9,clip,width=\linewidth]{figure/icons/parts/5_0_0.png}} & \shortstack{2\\\includegraphics[trim=0 70 0 9,clip,width=\linewidth]{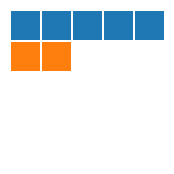}} & \shortstack{1\\\includegraphics[trim=0 70 0 9,clip,width=\linewidth]{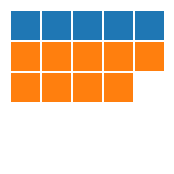}} & \shortstack{10\\\includegraphics[trim=0 70 0 9,clip,width=\linewidth]{figure/icons/parts/6_0_0.png}} & \shortstack{1\\\includegraphics[trim=0 70 0 9,clip,width=\linewidth]{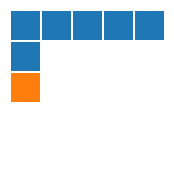}} &  &  &  &  &  \\ 
 & \shortstack{2\\\includegraphics[trim=0 50 0 9,clip,width=\linewidth]{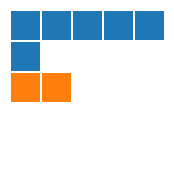}} & \shortstack{1\\\includegraphics[trim=0 50 0 9,clip,width=\linewidth]{figure/icons/parts/6_3_0.png}} & \shortstack{1\\\includegraphics[trim=0 50 0 9,clip,width=\linewidth]{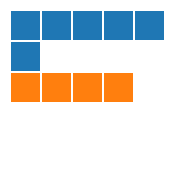}} & \shortstack{3\\\includegraphics[trim=0 50 0 9,clip,width=\linewidth]{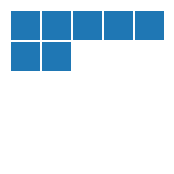}} & \shortstack{1\\\includegraphics[trim=0 50 0 9,clip,width=\linewidth]{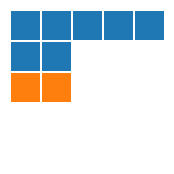}} &  &  &  &  &  \\ 
 & \shortstack{3\\\includegraphics[trim=0 10 0 9,clip,width=\linewidth]{figure/icons/parts/8_0_0.png}} & \shortstack{1\\\includegraphics[trim=0 10 0 9,clip,width=\linewidth]{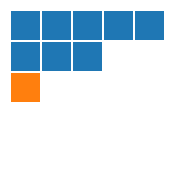}} & \shortstack{1\\\includegraphics[trim=0 10 0 9,clip,width=\linewidth]{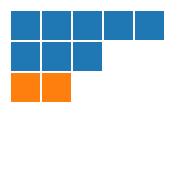}} & \shortstack{1\\\includegraphics[trim=0 10 0 9,clip,width=\linewidth]{figure/icons/parts/9_0_0.png}} & \shortstack{1\\\includegraphics[trim=0 10 0 9,clip,width=\linewidth]{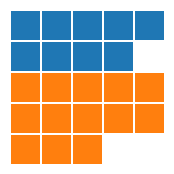}} &  &  &  &  &  \\ 
 & \shortstack{1\\\includegraphics[trim=0 70 0 9,clip,width=\linewidth]{figure/icons/parts/10_0_0.png}} & \shortstack{1\\\includegraphics[trim=0 70 0 9,clip,width=\linewidth]{figure/icons/parts/11_0_0.png}} & \shortstack{1\\\includegraphics[trim=0 70 0 9,clip,width=\linewidth]{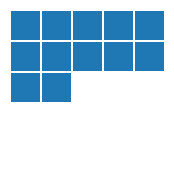}} &  &  &  &  &  &  &  \\ 
 
 \midrule 
\multirow{4}{*}{table}  & \shortstack{5\\\includegraphics[trim=0 100 0 9,clip,width=\linewidth]{figure/icons/parts/0_2_0.png}} & \shortstack{14\\\includegraphics[trim=0 100 0 9,clip,width=\linewidth]{figure/icons/parts/1_0_0.png}} & \shortstack{2\\\includegraphics[trim=0 100 0 9,clip,width=\linewidth]{figure/icons/parts/1_1_0.png}} & \shortstack{1\\\includegraphics[trim=0 100 0 9,clip,width=\linewidth]{figure/icons/parts/1_2_0.png}} & \shortstack{19\\\includegraphics[trim=0 100 0 9,clip,width=\linewidth]{figure/icons/parts/2_0_0.png}} & \shortstack{2\\\includegraphics[trim=0 100 0 9,clip,width=\linewidth]{figure/icons/parts/0_1_0.png}} & \shortstack{4\\\includegraphics[trim=0 100 0 9,clip,width=\linewidth]{figure/icons/parts/0_2_0.png}} & \shortstack{1\\\includegraphics[trim=0 100 0 9,clip,width=\linewidth]{figure/icons/parts/0_3_0.png}} & \shortstack{26\\\includegraphics[trim=0 100 0 9,clip,width=\linewidth]{figure/icons/parts/1_0_0.png}} & \shortstack{3\\\includegraphics[trim=0 100 0 9,clip,width=\linewidth]{figure/icons/parts/1_1_0.png}} \\ 
 & \shortstack{10\\\includegraphics[trim=0 100 0 9,clip,width=\linewidth]{figure/icons/parts/3_0_0.png}} & \shortstack{2\\\includegraphics[trim=0 100 0 9,clip,width=\linewidth]{figure/icons/parts/3_1_0.png}} & \shortstack{3\\\includegraphics[trim=0 100 0 9,clip,width=\linewidth]{figure/icons/parts/4_0_0.png}} & \shortstack{3\\\includegraphics[trim=0 100 0 9,clip,width=\linewidth]{figure/icons/parts/4_1_0.png}} & \shortstack{3\\\includegraphics[trim=0 100 0 9,clip,width=\linewidth]{figure/icons/parts/5_0_0.png}} & \shortstack{15\\\includegraphics[trim=0 100 0 9,clip,width=\linewidth]{figure/icons/parts/2_0_0.png}} & \shortstack{1\\\includegraphics[trim=0 100 0 9,clip,width=\linewidth]{figure/icons/parts/2_1_0.png}} & \shortstack{3\\\includegraphics[trim=0 100 0 9,clip,width=\linewidth]{figure/icons/parts/2_2_0.png}} & \shortstack{9\\\includegraphics[trim=0 100 0 9,clip,width=\linewidth]{figure/icons/parts/3_0_0.png}} & \shortstack{4\\\includegraphics[trim=0 100 0 9,clip,width=\linewidth]{figure/icons/parts/4_0_0.png}} \\ 
 & \shortstack{1\\\includegraphics[trim=0 70 0 9,clip,width=\linewidth]{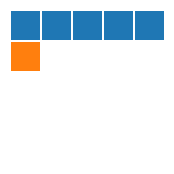}} & \shortstack{1\\\includegraphics[trim=0 70 0 9,clip,width=\linewidth]{figure/icons/parts/6_0_0.png}} & \shortstack{1\\\includegraphics[trim=0 70 0 9,clip,width=\linewidth]{figure/icons/parts/6_2_0.png}} & \shortstack{4\\\includegraphics[trim=0 70 0 9,clip,width=\linewidth]{figure/icons/parts/7_0_0.png}} & \shortstack{1\\\includegraphics[trim=0 70 0 9,clip,width=\linewidth]{figure/icons/parts/7_2_0.png}} & \shortstack{1\\\includegraphics[trim=0 70 0 9,clip,width=\linewidth]{figure/icons/parts/4_1_0.png}} & \shortstack{2\\\includegraphics[trim=0 70 0 9,clip,width=\linewidth]{figure/icons/parts/5_0_0.png}} & \shortstack{1\\\includegraphics[trim=0 70 0 9,clip,width=\linewidth]{figure/icons/parts/6_1_0.png}} & \shortstack{1\\\includegraphics[trim=0 70 0 9,clip,width=\linewidth]{figure/icons/parts/7_0_0.png}} & \shortstack{2\\\includegraphics[trim=0 70 0 9,clip,width=\linewidth]{figure/icons/parts/8_0_0.png}} \\ 
 & \shortstack{2\\\includegraphics[trim=0 70 0 9,clip,width=\linewidth]{figure/icons/parts/8_0_0.png}} &  &  &  &  & \shortstack{1\\\includegraphics[trim=0 70 0 9,clip,width=\linewidth]{figure/icons/parts/9_0_0.png}} & \shortstack{1\\\includegraphics[trim=0 70 0 9,clip,width=\linewidth]{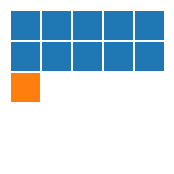}} &  &  &  \\ 
 
\midrule
\multirow{2}{*}{appliance}  & \shortstack{17\\\includegraphics[trim=0 100 0 9,clip,width=\linewidth]{figure/icons/parts/0_1_0.png}} & \shortstack{6\\\includegraphics[trim=0 100 0 9,clip,width=\linewidth]{figure/icons/parts/0_2_0.png}} & \shortstack{2\\\includegraphics[trim=0 100 0 9,clip,width=\linewidth]{figure/icons/parts/0_3_0.png}} & \shortstack{2\\\includegraphics[trim=0 100 0 9,clip,width=\linewidth]{figure/icons/parts/0_4_0.png}} & \shortstack{2\\\includegraphics[trim=0 100 0 9,clip,width=\linewidth]{figure/icons/parts/1_1_0.png}} & \shortstack{7\\\includegraphics[trim=0 100 0 9,clip,width=\linewidth]{figure/icons/parts/0_0_1.png}} & \shortstack{92\\\includegraphics[trim=0 100 0 9,clip,width=\linewidth]{figure/icons/parts/0_1_0.png}} & \shortstack{35\\\includegraphics[trim=0 100 0 9,clip,width=\linewidth]{figure/icons/parts/0_2_0.png}} & \shortstack{2\\\includegraphics[trim=0 100 0 9,clip,width=\linewidth]{figure/icons/parts/0_3_0.png}} & \shortstack{1\\\includegraphics[trim=0 100 0 9,clip,width=\linewidth]{figure/icons/parts/1_0_0.png}} \\ 
 & \shortstack{1\\\includegraphics[trim=0 100 0 9,clip,width=\linewidth]{figure/icons/parts/1_2_0.png}} & \shortstack{1\\\includegraphics[trim=0 100 0 9,clip,width=\linewidth]{figure/icons/parts/2_2_0.png}} &  &  &  &  &  &  &  &  \\ 
 
 \midrule 
\multirow{1}{*}{bed}  & \shortstack{3\\\includegraphics[trim=0 130 0 9,clip,width=\linewidth]{figure/icons/parts/2_0_0.png}} & \shortstack{2\\\includegraphics[trim=0 130 0 9,clip,width=\linewidth]{figure/icons/parts/3_0_0.png}} & \shortstack{1\\\includegraphics[trim=0 130 0 9,clip,width=\linewidth]{figure/icons/parts/5_0_0.png}} &  &  &  &  &  &  &  \\ 
 
 \midrule 
\multirow{2}{*}{other}  & \shortstack{5\\\includegraphics[trim=0 70 0 9,clip,width=\linewidth]{figure/icons/parts/0_0_1.png}} & \shortstack{3\\\includegraphics[trim=0 70 0 9,clip,width=\linewidth]{figure/icons/parts/0_1_0.png}} & \shortstack{2\\\includegraphics[trim=0 70 0 9,clip,width=\linewidth]{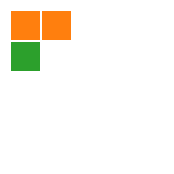}} & \shortstack{1\\\includegraphics[trim=0 70 0 9,clip,width=\linewidth]{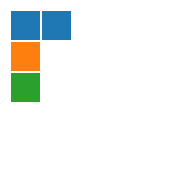}} &  & \shortstack{54\\\includegraphics[trim=0 70 0 9,clip,width=\linewidth]{figure/icons/parts/0_0_1.png}} & \shortstack{37\\\includegraphics[trim=0 70 0 9,clip,width=\linewidth]{figure/icons/parts/0_1_0.png}} & \shortstack{2\\\includegraphics[trim=0 70 0 9,clip,width=\linewidth]{figure/icons/parts/0_2_0.png}} & \shortstack{2\\\includegraphics[trim=0 70 0 9,clip,width=\linewidth]{figure/icons/parts/1_0_0.png}} & \shortstack{1\\\includegraphics[trim=0 70 0 9,clip,width=\linewidth]{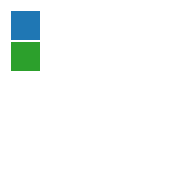}} \\ 
 &  &  &  &  &  & \shortstack{3\\\includegraphics[trim=0 100 0 9,clip,width=\linewidth]{figure/icons/parts/1_1_0.png}} &  &  &  &  \\  
 
\bottomrule
\end{tabularx}
\caption{
Summary visualization of part configurations for different object categories in the two datasets for our experiments (\ourdatashort on left and \pmopen on right).
Each colored block represents one openable part: blue are drawers, orange are doors, and green are lids.
The numbers above each icon are the counts of objects with that part configuration.
}
\label{fig:supp-obj-part-stats}
\end{figure*}

\ourdatashort focused on a diverse collection of objects with a variety of part configurations and part shapes, with subclasses of storage furniture and tables which have more part variation than appliances. \Cref{fig:supp-part-structure-distribution} shows the distribution of part configurations based on the number of drawers, doors, and lids that a object has.  Compared to \pmopen, \ourdatashort has a wider range of different number and type of parts for each object.  Many \pmopen objects have just 1 or 2 parts of the same type (71\% compared to 39\% for \ourdatashort).  \ourdatashort also has more variety in motion types in different object categories such as objects with doors that translate (see \cref{fig:supp-part-motion}).
We show counts per part configuration in \cref{fig:supp-obj-part-stats}.

\ourdatashort has diverse object and part shapes such as non-rectangular or curved drawers (22), L-shaped desks (14), angled corner cabinets (10), and objects with drawer-like parts that articulate as doors (10).  The complex part arrangements and diversity in shapes all make \ourdatashort a better dataset for benchmarking generalization of openable part segmentation and motion prediction.

\subsection{\acd-train and \acd-val}
\label{sec:supp-exp-acd-train}
To study the distribution gap between \pmopen and \acd as well as its potential for training, we split \acd into train and val by selecting shapes by their original data source (to ensure that the val set is not too similar to the training set). 
We select all the shapes from 3D-FUTURE \cite{fu20213d} (185) to use for training, and keep shapes from HSSD \cite{khanna2023habitat} (147) and ABO \cite{collins2022abo} (22) for evaluation. 
This results in a total of 645 training shapes (PM + 3DF), 169 validation shapes in \acd-val and 95 in \pmopen-val. 
The additional data allows us to compare training with PM only, with addition of \pmopenext, with ACD 3DF models, or with both added.


\clearpage
\clearpage

\subsection{Analysis of PM-Openable}
\label{sec:supp-data-pm}

To create PM-Openable, we select from PartNet-Mobility openable objects.  In total, we obtain 648 objects (out of a total of 2346 objects).  These openable containers make up roughly 28\% of objects in PartNet-Mobility. 

\begin{figure}
\includegraphics[trim=20 10 0 20,clip,width=\linewidth]{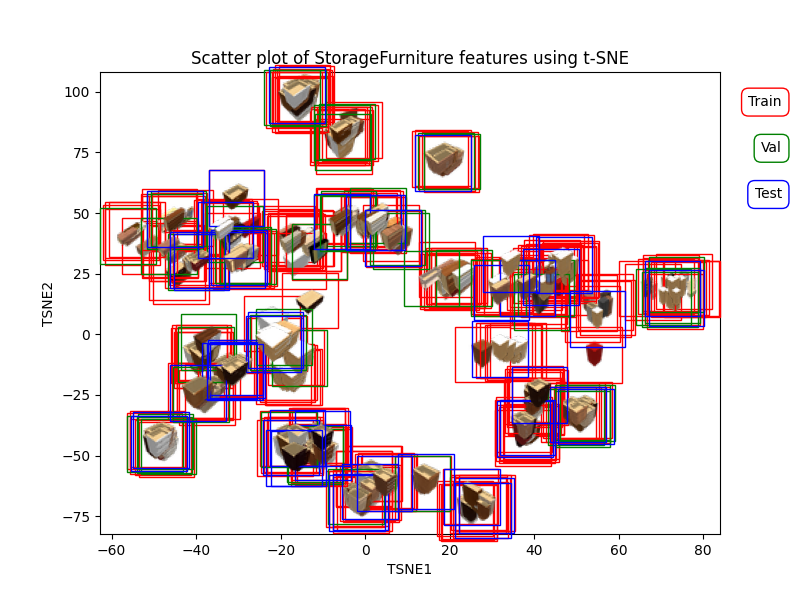}
\vspace{-2em}
\caption{Projection of in \pmopen storage furniture object embeddings using \tsne. Note strong clustering of objects spanning train, val, and test splits.  This is indicative of the lack of a general lack of diversity in object geometry, and some degree of data leakage between the splits.}
\label{fig:supp-tsne-storage}
\end{figure}

While PM-Openable contains a large number of objects, we find that objects in PM-Openable are highly similar (even across the train/val/test splits).  This is especially true for category with the largest number of objects (Storage Furniture).  In the main paper Fig. 3, we show examples from PM-Openable that are visually very similar.  Here in \Cref{fig:supp-tsne-storage},  we visualize \imnet~\cite{chen2019learning} shape embeddings for storage furniture in \pmopen across the train/val/test sets.   We project the embeddings using \tsne~\cite{van2008visualizing} with the perplexity set to 10.
As the figure shows, storage furniture in \pmopen is highly repetitive within and across the train, val, and test splits.
This clustering of highly similar objects across splits is indicative of both a lack of diversity and data leakage between the splits.
Observation of these trends was one of our motivations for the construction of \ourdatashort.
The above statistics and the object similarity analysis from Section 4 of the main paper show that \ourdatashort is more diverse and exhibits significantly less similarity between splits.

\begin{figure}
\centering
\setkeys{Gin}{width=\linewidth}
\begin{tabularx}{\linewidth}{Y Y}
\toprule
original & original interior\\
\includegraphics[trim=40 40 40 40,clip,width=\linewidth]{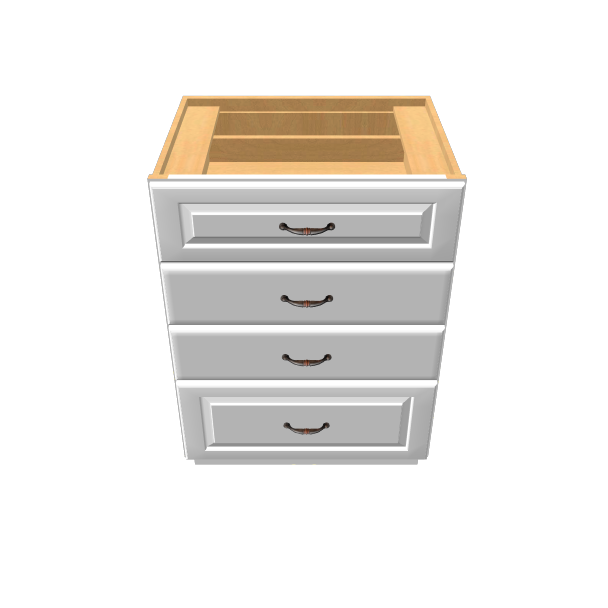} & \includegraphics[trim=40 40 40 40,clip,width=\linewidth]{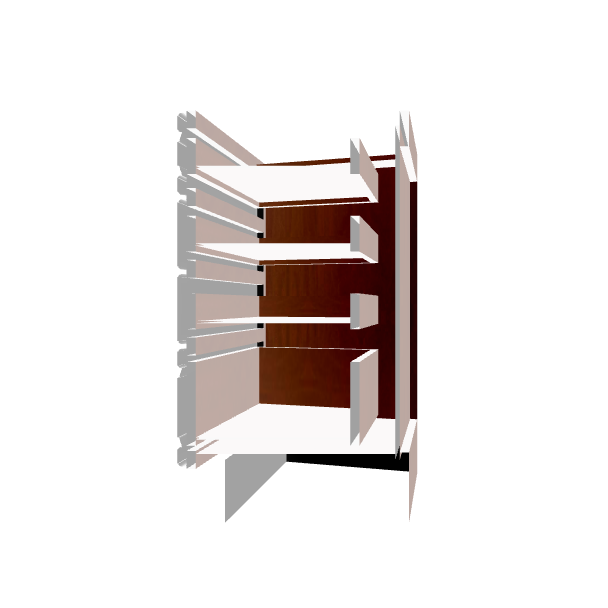}\\
+countertop &  +countertop --interior\\
\includegraphics[trim=40 40 40 40,clip,width=\linewidth]{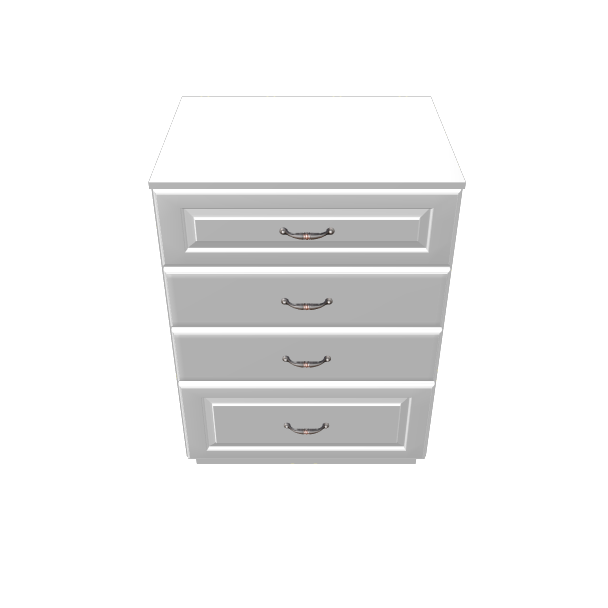} &
\includegraphics[trim=40 40 40 40,clip,width=\linewidth]{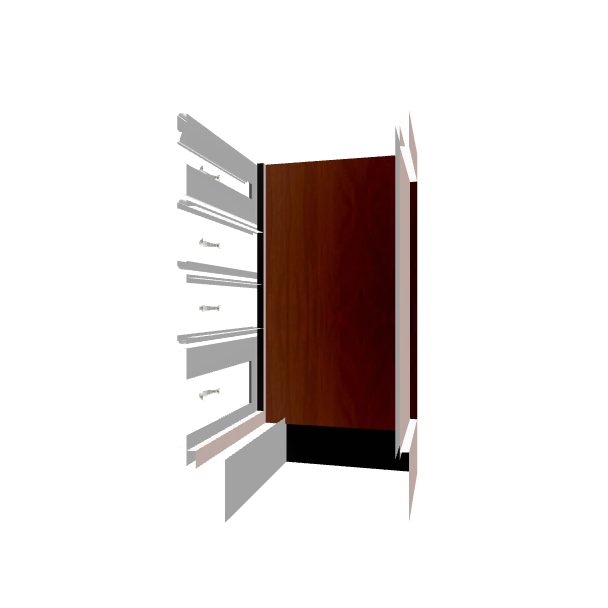}
\\
\bottomrule
\end{tabularx}
\caption{Example of an object in \pmopen (top) with corresponding object from \pmopenext (bottom) after addition of counterop surface and removal of interior geometry.}
\label{fig:supp-pm-ext}
\end{figure}

\subsection{Construction of \pmopenext}
\label{sec:supp-data-pmext}

The first step required to remove interiors from \pmopen shapes is to perform an over-segmentation.
As \pmopen objects are originally human-modelled URDFs, they already come segmented in semantic parts.
In order to segment them further, we employ connectivity-based segmentation which works relatively well.
This is because the original 3D assets from which \pmopen is constructed were authored by human designers.

Then, we proceed to render the indices of triangles visible from the views sampled uniformly around the shape and keep only the precomputed segments that have at least one triangle visible.
This way, we eliminate the interiors while keeping the triangles connected to the outer layer of the shape which preserves some structure rather than having a single triangle layer equivalent to the scan.
\pmopen contains a significant number of shapes with missing countertops, which would exhibit artifacts with a straightforward application of the above algorithm.
Therefore, we identify all such shapes and add a countertop surface via a simple algorithm as a pre-processing step.
An example of an object before and after the full procedure can be found in \cref{fig:supp-pm-ext}.

\clearpage
\clearpage

\begin{figure*}
\includegraphics[width=\linewidth]{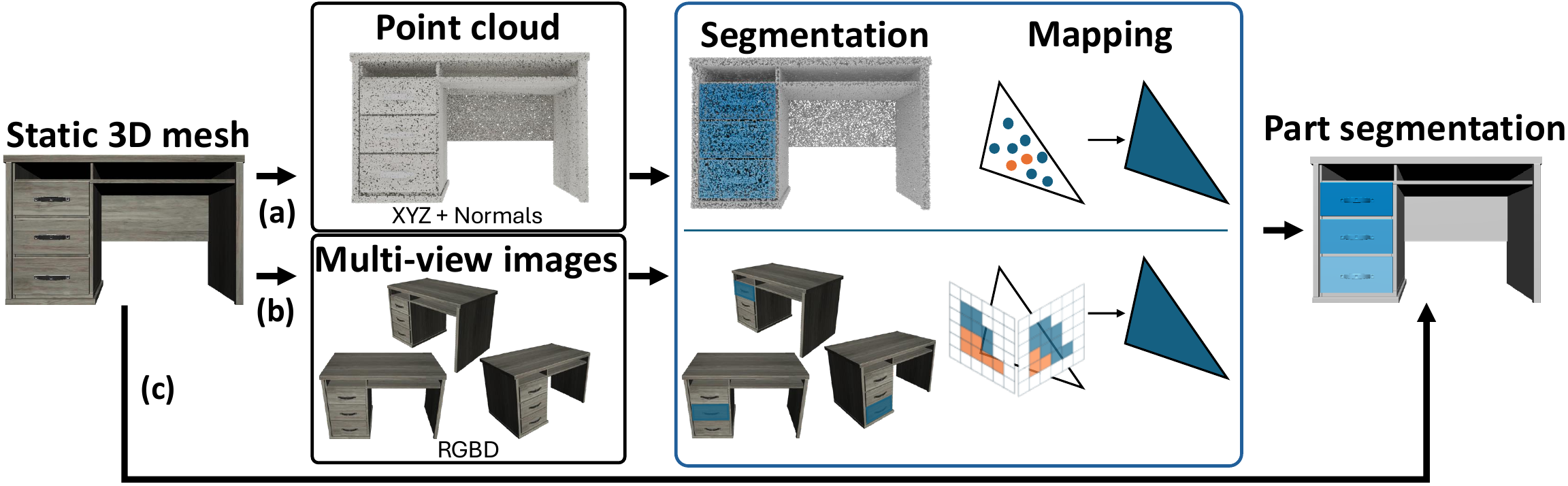}
\caption{
Illustration of our part segmentation experiments on different modalities: a) point cloud-based; b) image-based; and c) 3D mesh-based segmentation methods. The three modalities are all produced from the input mesh.
}
\vspace{-8pt}
\label{fig:seg-modalities}
\end{figure*}

\section{Part segmentation details}
\label{sec:supp-segmentation-details}

Here we describe baseline methods (\cref{sec:supp-part-seg-baselines}) and implementation details for the part segmentation stage of our framework.  We provide details on our FPN for improving PointGroup (\cref{sec:supp-seg-fpn}), the point-cloud sampling (\cref{sec:supp-pc-sampling}), projecting predictions to the mesh from rendered images (\cref{sec:supp-image-mapping}) and sampled point-clouds (\cref{sec:supp-pc-mapping}),  implementation and training details (\cref{sec:supp-experiment-impl-details}).

\subsection{Part segmentation baselines}
\label{sec:supp-part-seg-baselines}

We consider obtaining part segmentation from different types of input: rendered images, sampled-point cloud, or directly on the mesh.  \Cref{fig:seg-modalities} illustrates how segmentation can be done for different modalities.

\mypara{Image.}
We use \opdformer~\cite{sun2023opdmulti} for view-based segmentation as it is the state-of-the-art for openable part detection in images.
We sample views from the training viewpoint distribution used by \opdformer, and render three RGB and depth views per object.
To ensure that we can project the predicted segmentation back onto the mesh, we also render triangle indices for each view.
See \cref{sec:supp-image-mapping} for image to mesh projection details.

\mypara{Point-cloud.}
For point-cloud, we considered several baselines including \stom~\cite{wang2019shape2motion} which is designed for segmenting articulated objects and motion prediction, and two commonly used object segmentation methods: \pointgroup~\cite{jiang2020pointgroup} and Mask3D~\cite{schult2023mask3d}. See \cref{sec:supp-pc-sampling} for point sampling details and \cref{sec:supp-pc-mapping} for image to mesh projection details.

\mypara{Mesh.}
We use \meshwalker~\cite{lahav2020meshwalker}, an RNN-based method that predicts semantic segmentation using random walks on the mesh vertices.
This method makes far fewer assumptions such as manifoldness and watertightness on the input mesh compared to other mesh segmentation methods.
Since the input 3D object meshes typically do not conform to such requirements, this method is ideal for practical application.
Since MeshWalker only outputs semantic categories (i.e. it does not predict instances), we treat each semantic category as one instance.

\subsection{FPN details}
\label{sec:supp-seg-fpn}

We introduce a simple Feature Pyramid Network (FPN) as a feature adapter module after the initial PointNeXt feature extraction in PointGroup.
Our FPN consists of: pre-FPN convolution, FPN with 3 bottleneck blocks with the first one halving the feature dimension, followed by top-down convolutions per bottleneck block. Features are concatenated afterwards and passed to a post-FPN convolution that reduces them back to the original hidden dimension.

\subsection{Point cloud sampling details}
\label{sec:supp-pc-sampling}

For training, we sample 200K points for each annotated part and apply farthest point sampling (FPS) to downsample to 20K points total for each object.
During inference, to ensure every triangle is represented, we sample a point cloud with 1M points from the mesh, and add all triangle vertices as points for the methods leveraging triangle-based propagation so that all triangles are covered.
To match training, we downsample to  20K points for inference, and predictions are mapped back to the high-res point cloud using $k$ nearest neighbor lookup.
We leverage the computed connectivity segmentation to guide the initial sampling of 1M points by distributing the number of samples for each connected component based on the area of the component, with a threshold if the area is too small. This allows to obtain detailed geometry even for small parts (e.g., handles). Such point clouds are, however, non-uniform. One would expect that non-uniformness would disappear after applying farthest point downsampling from 1M to 20k points, however we find that our models are not robust to the changes in sampling.

Then, to propagate predictions to the original point cloud from 20K subset, we apply a $k$-nearest neighbors lookup to map predictions to all points in the point cloud.
We set $k = 3$ in our experiments.

\subsection{Image to mesh projection}
\label{sec:supp-image-mapping}
Predictions on the images are projected to the mesh using triangle indices at each pixel.
We keep instance predictions with confidence greater than a threshold of 0.9.
It is possible that predicted part masks overlap in triangles.
To reconcile different predictions, we consider all predictions in order of confidence, from highest to lowest.
Overlapping masks in a triangle with triangle-area weighted IoU greater than 0.8 are merged into one part and we take the label from the prediction with higher confidence.
Otherwise, we retain two separate parts and assign the triangle to the higher confidence prediction.
By reasoning with projected masks (vs individual triangles), we ensure that part predictions on the mesh are not broken into smaller interleaving parts.
Note that this approach assumes confidence scores are comparable across views.
Triangles not observed in any view are assigned to the \textit{base} part.

\subsection{Point cloud to mesh projection}
\label{sec:supp-pc-mapping}
\mypara{Default propagation (voting for triangles).} We employ voting for triangles for our point cloud based baselines. The point cloud segmentation is projected to the original mesh, and overlapping predictions are handled with a heuristic similar to the one for images.
If the IoU of two masks is larger than 0.8, we keep only the mask with higher confidence, otherwise we assign overlapping points to the predicted instance with higher confidence and keep both.
Semantic instance labels are assigned by majority voting per triangle, which guarantees at least three votes since vertices are included in our point clouds during inference.

\mypara{Over-segmentation voting for topology-aware propagation.}
We also design a topology-aware propagation procedure using voting on over-segmentations. The over-segmentation is obtained by computing a connectivity segmentation of the mesh. Two triangles are considered connected if they share at least a single identical vertex index. We find that such over-segmentation often yields semantically-meaningful components, though at a finer level (i.e., drawer face, separate sides of drawer boxes, parts of drawer handles). Thus, propagating the labels to such segments yields much more complete segmentation than using separate triangles. This, however, improves the results only for a model that is sufficiently good, while incorporating such procedure for the bad models tends to propagate the errors further. It also eliminates the need of including the vertices in the point clouds during inference as there is no concern of zero points being sampled from a segment. Other details of the mapping procedure are identical to the default voting on triangles strategy.

\subsection{Training and implementation details}
\label{sec:supp-experiment-impl-details}

\mypara{Implementation details.} All point cloud methods use only point coordinates and normals as input features.
We use a \pointgroup (\pg) re-implementation\footnote{\url{https://github.com/3dlg-hcvc/minsu3d}} for the \unet backbone, and adapt it to work with \pointnext \footnote{\url{https://github.com/guochengqian/PointNeXt}}.
For \pg with Swin3D, we use the Pointcept implementation~\cite{pointcept2023}.
For \stom, we use the re-implementation from \citet{mao2022multiscan}.
We follow the suggested hyperparameters from the original work and train until convergence (see supplement).
For \maskthreed, we query the transformer with 10 queries, setting the confidence threshold on predictions to 0.7 during inference.
We use a pretrained \opdformer-P checkpoint that is trained on RGB-D data.
This model takes images of size $256\times256$ as an input. 

\mypara{Training details.}
\begin{table}[t]
\centering
\caption{Training parameters for part segmentation.}
\label{tab:training-parameters}
\resizebox{\linewidth}{!}
{
\begin{tabular}{@{} l rrrrr @{}}
\toprule
Method & dim & optimizer & learning rate & batch size & epochs \\ 
\midrule
\pg + \unet & 16 & \adam & 0.002 & 4 & 500 \\
\pg + \swinthreed & 48 & \adamw & 0.006 & 8 & 600 \\
\pg + \pointnext & 48 & \adamw & 0.002 & 8 & 500 \\
\maskthreed & 128 & \adamw & 0.0001 & 32 & 600 \\
\bottomrule
\end{tabular}
}
\end{table}

We train all models (except for \opdformer, which is pre-trained) on Nvidia RTX 2080Ti, A5000 and A40 GPUs.
We train to convergence based ontrain set evaluation metrics, and find that training runs take from 5-12 hours for \pointgroup variants, 12 hours for \maskthreed, 17 hours for \stom, and roughly four and a half days for \meshwalker.
\Cref{tab:training-parameters} summarizes key training parameters for most of the methods we report in the main paper.  Below, we provide training details for the other methods.

\meshwalker is trained using \adam optimizer with learning rate of 0.00001 for 600K iterations.
All other parameters are set to defaults according to the original paper.

\stom consists of three modules that are trained in stages.  We train the Motion Part Proposal and Motion Attribute Proposal Modules for 500 epochs with batch size 8, then the Proposal Matching Module for 100 epochs with batch size 16, and finally the Motion Optimization Network for 100 epochs with batch size 8. We use learning rate 0.001 and \adam optimizer for all stages, with other parameters (e.g., loss weights) set according to the original paper.

\section{Motion prediction details}
\label{sec:supp-motion-details}
Here we provided information about the motion prediction baselines (\cref{sec:supp-motion-baselines}) and details of our heuristic based motion prediction (\cref{sec:supp-motion-heuristics}).

\subsection{Motion prediction baselines}
\label{sec:supp-motion-baselines}
\stom (\stomacro) predicts instance segmentations along with motions.
It does not produce part semantic predictions off the shelf.
We heuristically infer the part semantic label based on the predicted instance and corresponding mobility parameters.
All parts with prismatic motion are labeled as drawers.
Parts with revolute motion are split into:
1) vertical axis - labeled as door;
2) non-vertical axis with non-vertical average part normal (computed from normals of part points) - labeled as door;
3) other cases (horizontal axis and vertical average normal) - labeled as lid.
For \stomacro, the input point clouds are downsampled to 4096 points, so we map its predictions first to our subset and then to the full inference point clouds using $k$ nearest neighbor lookups.

\subsection{Motion prediction heuristic}
\label{sec:supp-motion-heuristics}

For motion prediction, we designed a heuristic-based method (introduced in \Cref{sec:motion-methods} of the main paper).  Here we provide more details about the heuristics we used.

For the motion type, we utilize the most common motion type for each part semantic category following the statistics from the train set. Specifically, prismatic motion is assigned to drawers, while revolute motion is assigned to doors and lids.
For the prismatic joint, the heuristic is simple: we use the given front direction of the openable part as the motion axis direction.
For the revolute joint, we assume the motion axis should be on one of the edges of the bounding box of the openable part (four edges on the front face and four edges on the back face).

To determine if the axis is on the front face or the back face, we check its alignment with the base part.
We observe that in most cases, the motion axis should also be very close to the edges of the base part.
Leveraging this observation, our heuristic determines whether the front face or the back face of the openable part is closer to the edges in the bounding box of the base part.

After selecting the face, we still have four edges from which to choose.
To handle this choice, we follow a geometric heuristic that finds the handle position first, and sets the axis line on the opposite edge of the handle position.
To determine the handle position, we leverage the assumption that the handle geometry is more complex compared to the whole openable part.
There are mainly two cases: the handle is raised (e.g., the door of the cabinet) or concave (e.g., the door of the washing machine). In both cases, there is some asymmetry in the geometry.
We detect this asymmetry by binning vertex densities from front to back and then locate the handle position accordingly.
We check the number of points in each bin to detect the handle and infer the edge that serves as the motion axis, and applying symmetry-based reasoning to select direction and origin for revolute joints.

\section{Additional Experiments}
\label{sec:additional-results}

We provide additional results including ablation experiments (\cref{sec:supp-exp-ablations}) for informing the design of our models, evaluation of point-cloud segmentation directly on point-clouds (\cref{sec:supp-exp-pc-seg}), evaluation of mesh segmentation using additional metrics (\cref{sec:supp-exp-mesh-seg}). 

\subsection{Ablation Experiments}
\label{sec:supp-exp-ablations}

\subsubsection{\ogroup ablations}
\label{sec:pg-backbone-ablate}
To construct \ogroup, we ablated a number of backbones in \Cref{tab:results-pg-backbone-ablate}. The losses used follow \pointgroup. Overall, we find that recent backbones provide a significant improvement compared to the original \unet. We find that using \pointnext leads to superior results on \pmopen but lags behind the \swinthreed when it comes to generalizing on \acd. We find that adding our FPN feature adapter leads to a very significant performance improvement, especially when it comes to generalizing on \acd. Employing our topology-aware voting procedure pushes results on \pmopen further, while the results on \acd deteriorate slightly. This is expected as voting for over-segmentation while the predictions are not very good could lead to propagating these errors further. This, however, changes with the additional data added into the training split as can be seen in the version trained on \pmopenext in addition. Any further improvements in training data would benefit from such propagation procedure even further. Thus, we select these modifications to be \ogroup. 

\begin{table}
\centering
\caption{
Ablation of choices made for \gammagroup construction and their effects on segmentation. OO stands for predicting offsets to origin rather than axis. OSV stands for voting for over-segmentation.
}
\vspace{-5pt}
{
\scriptsize
\begin{tabular}{@{} cc rrr rrr @{}}
\toprule
& & \multicolumn{3}{c}{\pmopen} & \multicolumn{3}{c}{\ourdatashort} \\
\cmidrule(l{0pt}r{2pt}){3-5} \cmidrule(l{0pt}r{2pt}){6-8}
OO & OSV & P & R & F1 & P & R & F1 \\
\midrule
\myxmark & \myxmark & 69.7 & 62.7 & 65.8 & 21.8 & 5.3 & 8.5\\
\checkmark & \myxmark & 75.9 & 65.9 & 70.4 & \textbf{24.8} & \textbf{5.1} & \textbf{8.4}\\
\checkmark & \checkmark& \textbf{77.3} & \textbf{66.9} & \textbf{71.6} & 24.5 & 4.5 & 7.5\\
\bottomrule
\end{tabular}
}
\label{tab:results-gammagroup-ablate}
\end{table}

\begin{table}[t]
\centering
\caption{Ablation of \gammagroup design choices and their effects on motion prediction, evaluated on \pmopen. OO stands for predicting offsets to origin rather than axis. OSV stands for voting for over-segmentation. EA stands for edge-aware motion predictions post-processing. 
}
\vspace{-2pt}
{
\scriptsize
\begin{tabular}{@{} ccc c rrr rr @{}}
\toprule 
& & & & \multicolumn{3}{c}{F1 \% $\uparrow$} & \multicolumn{2}{c}{Error $\downarrow$} \\
\cmidrule(l{0pt}r{2pt}){5-7} 
\cmidrule(l{0pt}r{2pt}){8-9}
OO & OSV & EA & \# & {+}M & {+}MA & {+}MAO & AE & OE \\
\midrule
\myxmark & \myxmark & \myxmark & 131 & 66.2 & 57.0 & 26.5 & 9.7 & 0.27 \\
\checkmark & \myxmark & \myxmark & 131 & 70.2 & 59.9 & 34.2 & 9.8 & 0.25 \\
\checkmark & \checkmark & \myxmark & \textbf{133} & \textbf{71.2} & 60.9 & 35.1 & 9.7 & 0.25 \\
\checkmark & \checkmark & \checkmark & \textbf{133} & \textbf{71.2} & \textbf{61.9} & \textbf{50.6} & \textbf{9.5} & \textbf{0.18} \\
\bottomrule
\end{tabular}
}
\label{tab:fpngroupmot-ablate-motion}
\end{table}

\subsubsection{\gammagroup ablations}
In \cref{tab:results-gammagroup-ablate}, we provide an ablation of our \gammagroup.

\mypara{Add motion prediction to \ogroup.}
We extend our \ogroup for motion prediction by adding motion prediction support similar to those used in \gammamotion~\cite{yu2024gamma}, which uses a submodule for segmentation and a separate submodule for motion prediction.
Following our \ogroup, we use two FPNs for preparing features for the two submodules.  This is akin to taking \gammamotion and replacing the convolutional projection layers with FPNs after the backbone and before each of the submodules. 

We also propose a number of additional changes to the initial \gammamotion formulation. We simplify the motion type losses from combination of dice and focal loss to simpler cross-entropy. Axis directions and offsets to origin are supervised using the \pointgroup losses - L1 on vector norm and negative cosine similarity on vector direction.  Moreover, we remove the use of combination of both offsets to part centroid from segmentation submodule and offsets to motion axis from motion submodule that were used for segmentation in \gammamotion and stick with only the latter as in the original formulation of \pointgroup. 

The next modification we add is predicting offsets not to motion axis but to motion origin in motion submodule. While this modification targets improving motion predictions, we find that it is also quite beneficial for segmentation. Since both task are learned together, it is hard to decouple their effects but our hypothesis is that it is easier to learn offsets to the origin which benefits the overall training stability. Finally, we employ topology-aware voting procedure which further pushes the segmentation results. We see minor reduction in performance on \acd, similar and as discussed by \cref{sec:pg-backbone-ablate}. 

We see that these improvements subsequently enhance the motion prediction capabilities. We see a large improvement in MAO F1 score when predicting offsets to origin directly rather than axis from 26.5 to 34.2 points. Voting for over-segments further improves the results a little. 

\mypara{Post-processing.} Finally, we propose a post-processing step inspired by our mobility heuristic. As the motion axis is extracted by majority voting, we decide to straighten it as in our data, aligning to world coordinate axes. Thus, we select the most dominant direction by absolute value, set it as 1 with corresponding sign and all other values are set to 0. We see that this results in minor improvement in MA F1 score from 60.9 to 61.9 points. Finally, we also post-process the predicted origin. As we voted for over-segmentations, we now likely have much better part bounds than with the initial point cloud predictions. We note that outputs of \gammamotion are actually not guaranteed to respect segmentation bounds as motion and segmentation predictions are done independently. We improve on it by computing oriented bounding boxes for the predicted doors and lids and finding the closest corner to the predicted motion origin. Then, we set that corner to be a new origin. This results in a significant improvement in origin quality, as can be seen from MAO F1 score improvement of 15.5 points as well as reduction in origin error.

\begin{table}
\centering
\caption{Point cloud segmentation evaluation.
Overall, \ourdata (\acd) is much more challenging than \pmopen.
Note that \stomacro does not output semantic labels, therefore evaluation considers two classes: base vs openable part.
}
\resizebox{\linewidth}{!}
{
\begin{tabular}{@{} l rrrrrr @{}}
\toprule
 & \multicolumn{3}{c}{\pmopen} & \multicolumn{3}{c}{\ourdatashort}\\
\cmidrule(l{0pt}r{2pt}){2-4} \cmidrule(l{0pt}r{2pt}){5-7}
Method & mAP$\uparrow$ & mAR$\uparrow$ & OC$\downarrow$ & mAP$\uparrow$ & mAR$\uparrow$ & OC$\downarrow$\\
\midrule

\pg + \unet & 38.0 & 44.6 & 0.236 & 17.4 & 22.3 & 0.357\\
\pg + \swinthreed & 41.0 & 51.2 & 0.209 & 17.6 & 25.5 & 0.339\\
\pg + \pointnextshort & 51.3 & 55.7 & 0.198 & 14.9 & 22.8 & 0.357\\
\ogroup & \textbf{69.3} & \textbf{75.3} & \textbf{0.099} & 21.0 & 27.8 & 0.324\\
\ogroup$\!^{*}$ & 67.8 & 72.4 & 0.117 & \textbf{28.2} & \textbf{33.6} & \textbf{0.303}\\
\gammagroup & 56.5 & 60.0 & 0.142 & 19.2 & 24.6 & 0.338 \\
\gammagroup$\!^{*}$ & 51.7 & 55.5 & 0.168 & 18.5 & 23.0 & 0.344 \\
\maskthreed & 34.9 & 41.5 & 0.250 & 15.1 & 18.3 & 0.370\\
$\stomacro$ & 11.2 & 13.6 & 0.323 & 15.2 & 16.9 & 0.375\\

\bottomrule

\end{tabular}
}
\label{tab:results-pc-seg}
\end{table}
\begin{table}[t]
\centering
\caption{Breakdown per part category for evaluation of point cloud segmentation methods. Segmentation of openable parts is challenging, especially for drawers in the \ourdata data dataset where interior geometry behind the drawer is not typically available as a useful signal. 
}
\resizebox{\linewidth}{!}
{
\begin{tabular}{@{} ll rrrr rrrr @{}}
\toprule
 & & \multicolumn{2}{c}{Drawer} & \multicolumn{2}{c}{Door} & \multicolumn{2}{c}{Lid} & \multicolumn{2}{c}{Base}\\
\cmidrule(l{0pt}r{2pt}){3-4} \cmidrule(l{0pt}r{2pt}){5-6} \cmidrule(l{0pt}r{2pt}){7-8} \cmidrule(l{0pt}r{2pt}){9-10} 
  & Method & AP$\uparrow$ & AR$\uparrow$ & AP$\uparrow$ & AR$\uparrow$ & AP$\uparrow$ & AR$\uparrow$ & AP$\uparrow$ & AR$\uparrow$\\
\midrule
\multirow{7}{*}{\rotatebox[origin=c]{90}{\pmopen}}
& \pg + \unet & 15.6 & 18.4 & 40.9 & 43.9  & 19.6 & 29.6 & 76.0 & 86.7\\
& \pg + \swinthreed & 12.6 & 20.1 & 51.3  & 61.1 & 9.8 & 30.9 & 90.1 & 92.5\\
& \pg + \pointnextshort & 11.4 & 13.2 & 52.5 & 59.6  & 55.2 & 58.0 & 85.9 & 91.9\\
& \ogroup & 55.2 & \textbf{63.9} & 69.7 & \textbf{77.2} & \textbf{64.9} & \textbf{66.7} & 87.4 & 93.5\\
& \ogroup$\!^{*}$ & \textbf{59.9} & 63.6 & \textbf{74.5} & 76.5 & 49.0 & 55.6 & \textbf{87.9} & \textbf{94.2}\\
& \gammagroup & 52.0 & 54.4 & 54.5 & 58.4 & 31.4 & 35.8 & 87.4 & 91.2 \\
& \gammagroup$\!^{*}$ & 55.1 & 55.4 & 49.0 & 51.2 & 18.6 & 24.7 & 84.1 & 90.9 \\
& \maskthreed & 30.7 & 36.7 & 22.7 & 27.4  & 18.4 & 27.2 & 67.8 & 74.6\\
& \stomacro & 0.0 & 0.4 & 0.0 & 0.4 & 0.0 & 0.0 & 44.6 & 53.7 \\

\midrule

\multirow{7}{*}{\rotatebox[origin=c]{90}{\ourdatashort}}

& \pg + \unet & 0.1 & 0.4 & 1.5 & 4.1 & 12.5 & 15.7 & 55.5 & 69.2\\
& \pg + \swinthreed & 0.0 & 0.2 & 2.2 & 8.6 & 2.5 & 15.7 & 65.7 & 77.3\\
& \pg + \pointnextshort & 0.0 & 0.1 & 2.4 & 4.6 & 6.0 & 20.4 & 51.3 & 65.9\\
& \ogroup & 0.0 & 0.2 & 8.3 & 13.3 & 14.1 & 25.0 & 61.6 & 72.9\\
& \ogroup$\!^{*}$ & \textbf{0.9} & \textbf{3.6} & \textbf{11.5} & \textbf{18.0} & \textbf{30.5} & \textbf{31.5} & \textbf{69.9} & \textbf{81.5}\\
& \gammagroup & 0.0 & 0.2 & 4.1 & 8.8 & 20.2 & 23.1 & 52.5 & 66.2 \\
& \gammagroup$\!^{*}$ & 0.0 & 0.5 & 2.7 & 5.7 & 15.5 & 17.6 & 55.7 & 68.4 \\
& \maskthreed & 0.0 & 0.0 & 0.9 & 3.5  & 0.9 & 0.9 & 58.4 & 68.9\\
& \stomacro & 0.0 & 0.0 & 0.0 & 0.1 & 0.0 & 0.0 & 60.9 & 67.4\\
\bottomrule
\end{tabular}
}
\label{tab:results-pc-seg-category}
\end{table}

\subsection{Part segmentation}
We provide additional results for part segmentation, including evaluation directly on the point-cloud (\cref{sec:supp-exp-pc-seg}, as well as a comparison of different approaches (image based, point-cloud based, mesh based) on mesh segmentation (\cref{sec:supp-exp-mesh-seg}).  
For point-cloud based methods, we compare 
\stom~\cite{wang2019shape2motion}, \maskthreed~\cite{schult2023mask3d}, \pointgroup~\cite{jiang2020pointgroup} with our proposed \ogroup.  For \pointgroup, we also compare the performance of different backbones: the original \unet, \swinthreed~\cite{yang2023swin3d}, and \pointnext~\cite{qian2022pointnext} which is the basis of our \ogroup.  
We also report the segmentation performance of our \gammagroup which includes a submodule for motion prediction.
In these experiments, we use $^*$ as an shorthand to indicate that the model was trained using both \pmopen and \pmopenext.  By default, only \pmopen was used in training.

\textit{Results consistently show that our \ogroup outperforms other methods}, with the model trained with additional \pmopenext data (\ogroup$\!^*$) having better performance on \acd.  We provide more details on these additional experiments and results below.

\subsubsection{Point-cloud segmentation}
\label{sec:supp-exp-pc-seg}
\mypara{Metrics.}
We report the \textbf{mAP} and \textbf{mAR} metrics for point cloud segmentation following prior work~\cite{jiang2020pointgroup}. Although these metrics are commonly used, we find that they have flaws:
1) they do not take false positives into account if their confidence is lower than the confidence of the true positive prediction; and
2) mAP and mAR are aggregated across classes and therefore do not weigh each shape equally. Therefore, we report another metric that does not have these issues: \textbf{\occost} by \citet{otani2022optimal}.
\occost (OC) evaluates the part segmentation quality for each shape by measuring the cost of correcting the predicted segmentation to match the ground truth segmentation. 
The metric was originally proposed for images and uses GIoU~\cite{rezatofighi2019generalized}.
We adapt it to shapes by computing GIoU on 3D bounding boxes.

\begin{figure*}
\centering
\setkeys{Gin}{width=\linewidth}
\begin{tabularx}{\textwidth}{p{2.5cm}  Y @{\hskip 8pt} Y @{\hskip 8pt} Y @{\hskip 8pt} Y @{\hskip 8pt} Y @{\hskip 8pt} Y @{\hskip 8pt} Y @{\hskip 8pt} Y}
\toprule
& \multicolumn{8}{c}{\textbf{\ourdatashort-val}}\\

GT &
\imgclip{0}{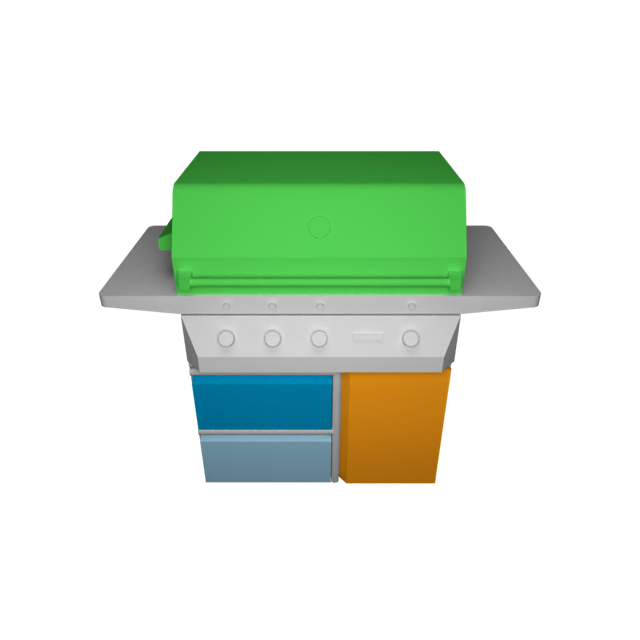} & 
\imgclip{0}{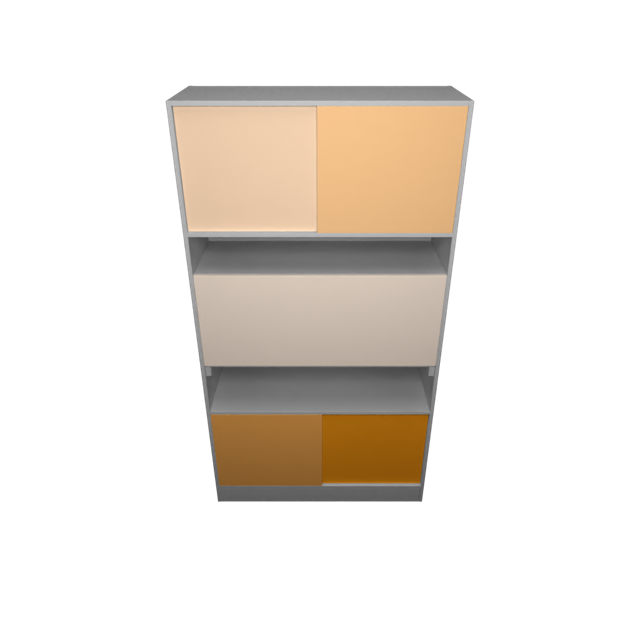} &
\imgclip{0}{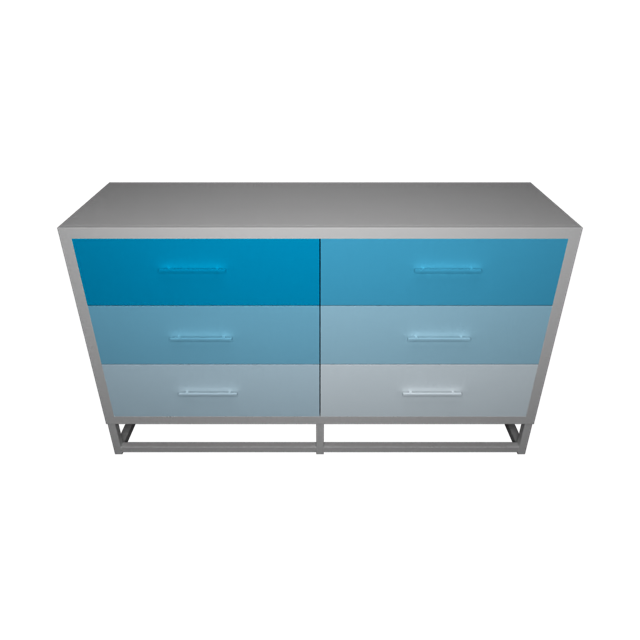} &
\imgclip{0}{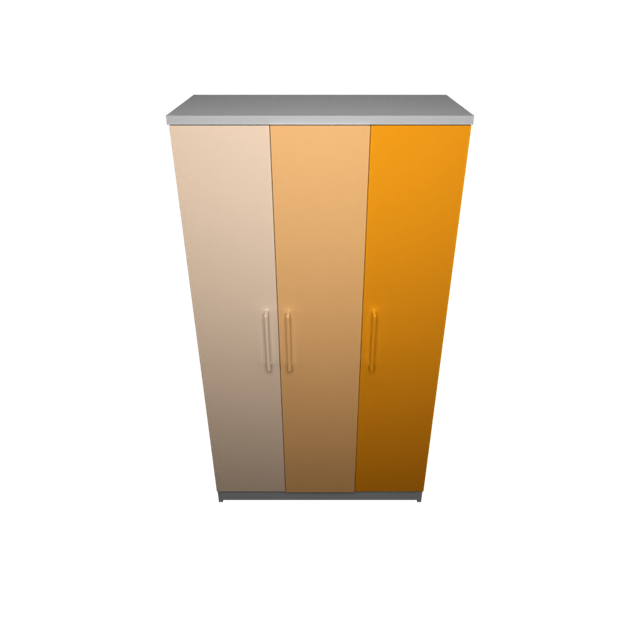} &
\imgclip{0}{figure/blender_supp/seg/gt/6ef98b7750ff80c5225b0f7c733f23d196998038} &
\imgclip{0}{figure/blender_supp/seg/gt/112dc87e26450400941c6eaff60866bc19badc64} &
\imgclip{0}{figure/blender_supp/seg/gt/d5ba163ba97f94c7aa4a4a625eb0547b8894e1a5} &
\imgclip{0}{figure/blender_supp/seg/gt/807955fd4dcb59b67789b24f2e7bc167027c870a} \\
\midrule

\ogroup &
\imgclip{0}{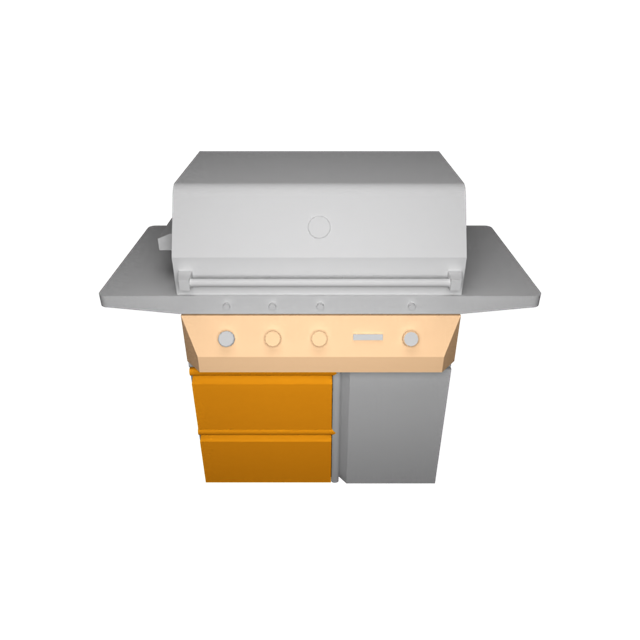} & 
\imgclip{0}{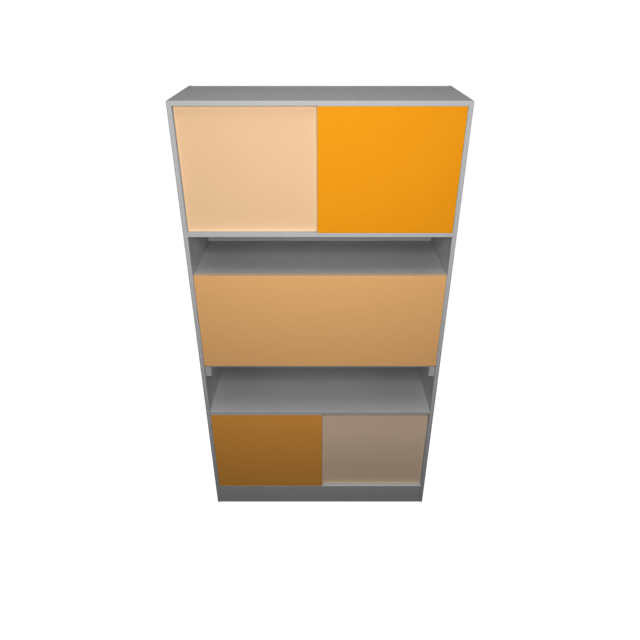} &
\imgclip{0}{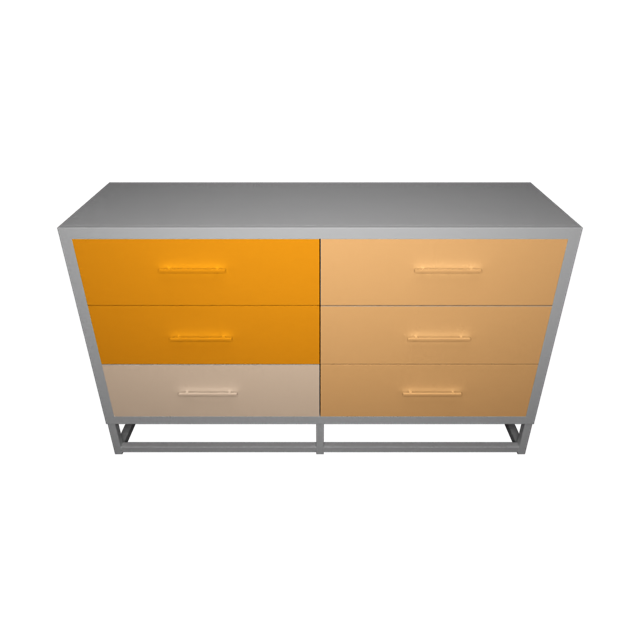} &
\imgclip{0}{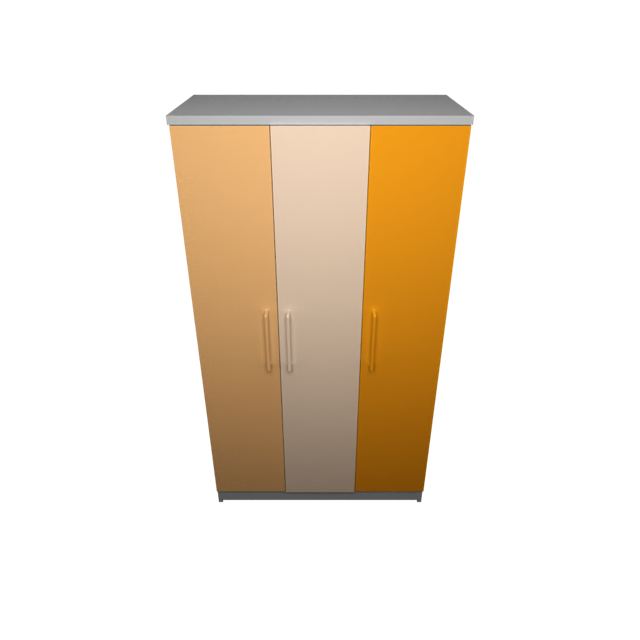} &
\imgclip{0}{figure/blender_supp/seg/fpngroup/6ef98b7750ff80c5225b0f7c733f23d196998038} &
\imgclip{0}{figure/blender_supp/seg/fpngroup/112dc87e26450400941c6eaff60866bc19badc64} &
\imgclip{0}{figure/blender_supp/seg/fpngroup/d5ba163ba97f94c7aa4a4a625eb0547b8894e1a5} &
\imgclip{0}{figure/blender_supp/seg/fpngroup/807955fd4dcb59b67789b24f2e7bc167027c870a} \\

\gammagroup &
\imgclip{0}{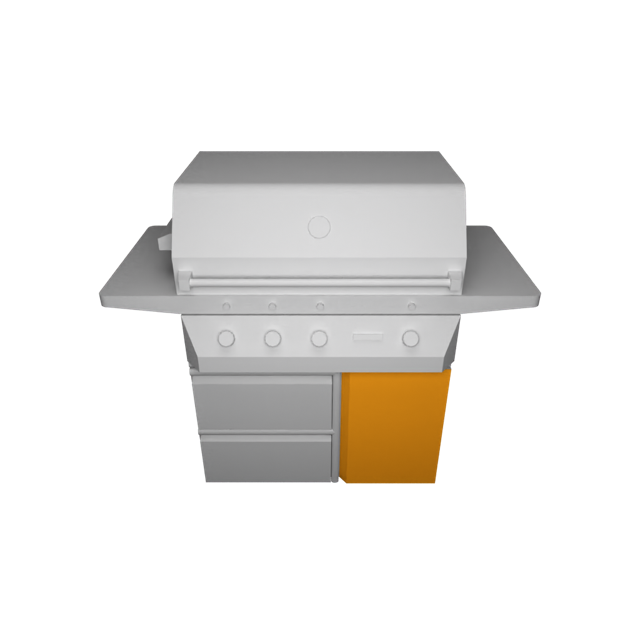} & 
\imgclip{0}{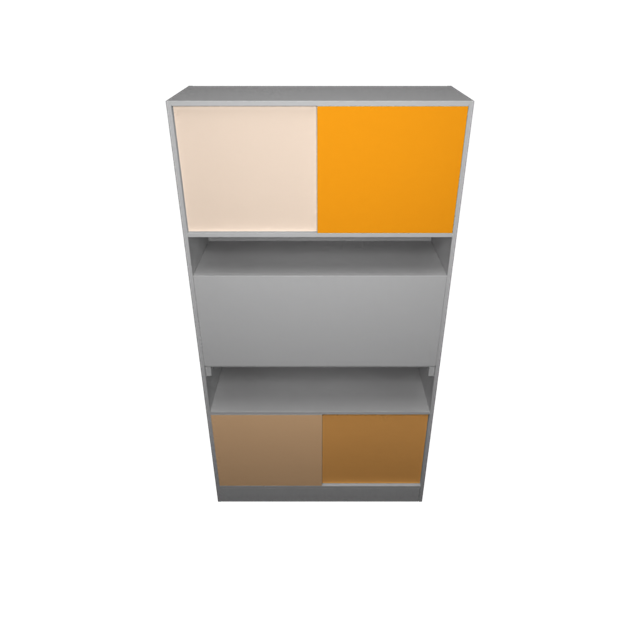} &
\imgclip{0}{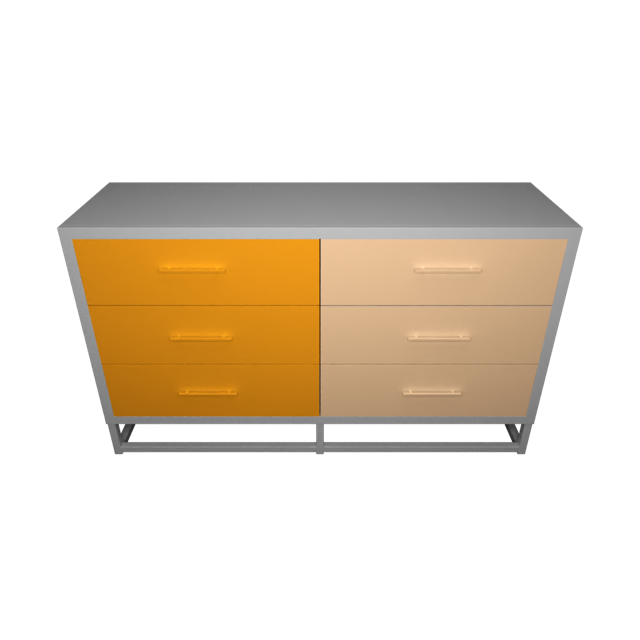} &
\imgclip{0}{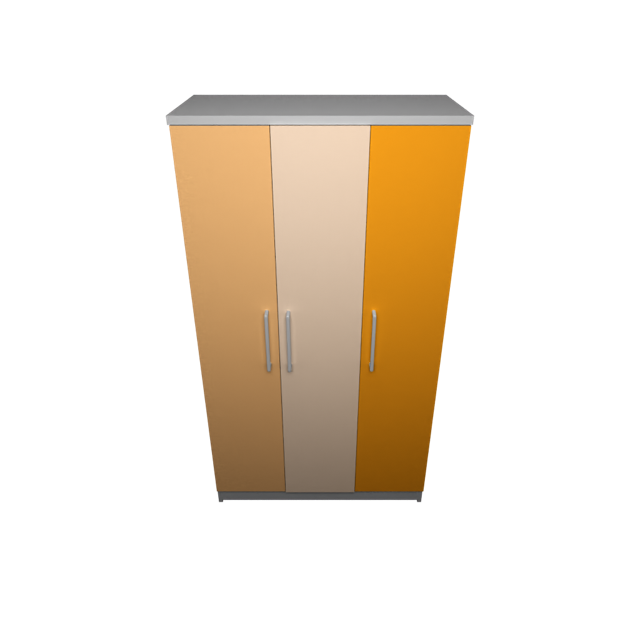} &
\imgclip{0}{figure/blender_supp/seg/fpngroupmot/6ef98b7750ff80c5225b0f7c733f23d196998038} &
\imgclip{0}{figure/blender_supp/seg/fpngroupmot/112dc87e26450400941c6eaff60866bc19badc64} &
\imgclip{0}{figure/blender_supp/seg/fpngroupmot/d5ba163ba97f94c7aa4a4a625eb0547b8894e1a5} &
\imgclip{0}{figure/blender_supp/seg/fpngroupmot/807955fd4dcb59b67789b24f2e7bc167027c870a} \\

\ogroup $^{\dagger}$ &
\imgclip{0}{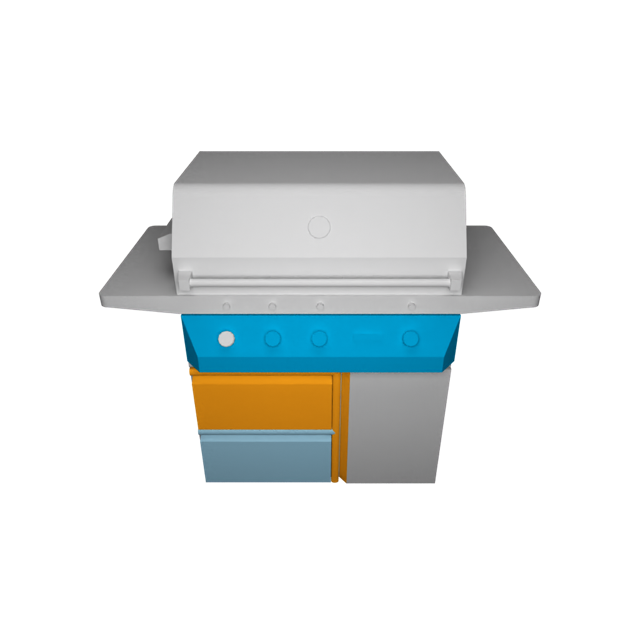} & 
\imgclip{0}{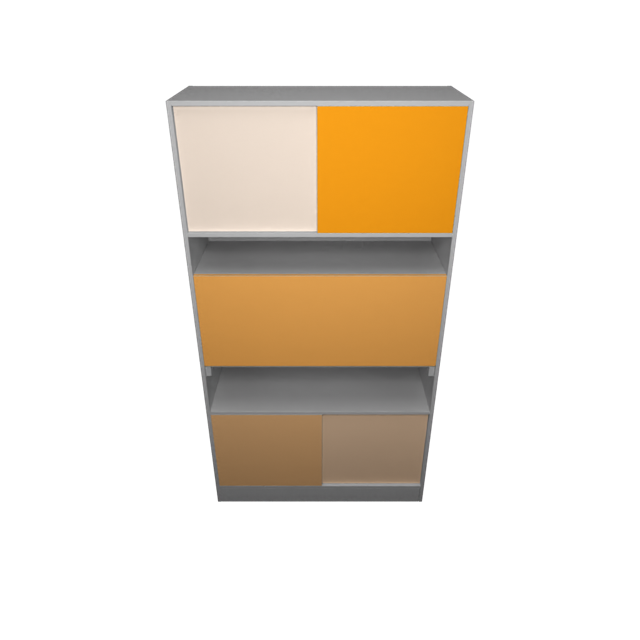} &
\imgclip{0}{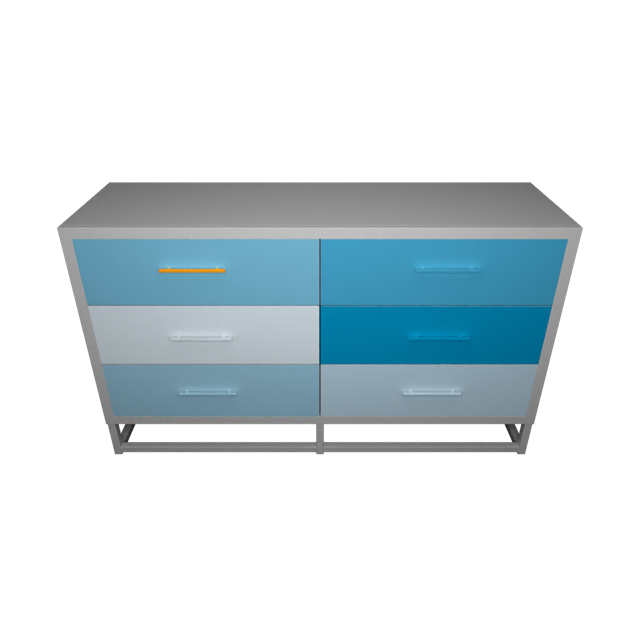} &
\imgclip{0}{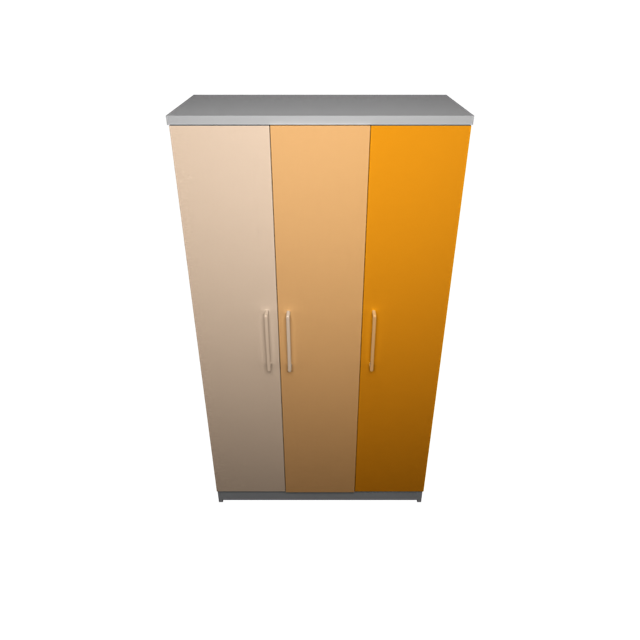} &
\imgclip{0}{figure/blender_supp/seg/fpngroup_w3df/6ef98b7750ff80c5225b0f7c733f23d196998038} &
\imgclip{0}{figure/blender_supp/seg/fpngroup_w3df/112dc87e26450400941c6eaff60866bc19badc64} &
\imgclip{0}{figure/blender_supp/seg/fpngroup_w3df/d5ba163ba97f94c7aa4a4a625eb0547b8894e1a5} &
\imgclip{0}{figure/blender_supp/seg/fpngroup_w3df/807955fd4dcb59b67789b24f2e7bc167027c870a} \\

\resizebox{!}{0.9em}{\gammagroup $^{\dagger}$} &
\imgclip{0}{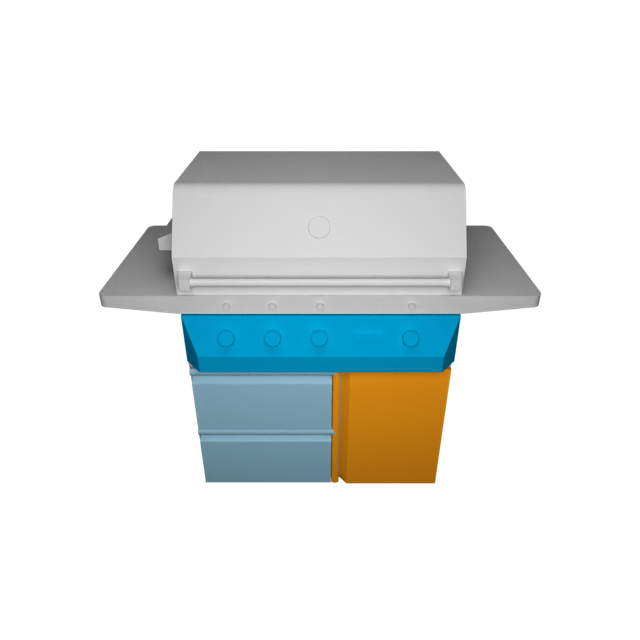} & 
\imgclip{0}{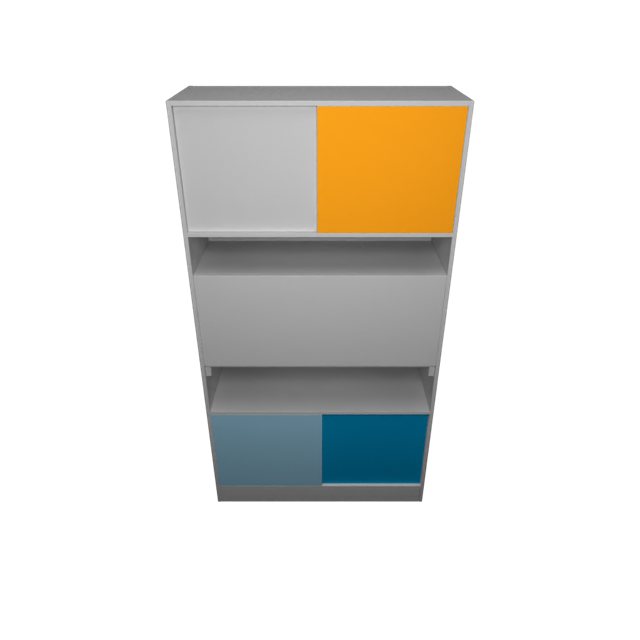} &
\imgclip{0}{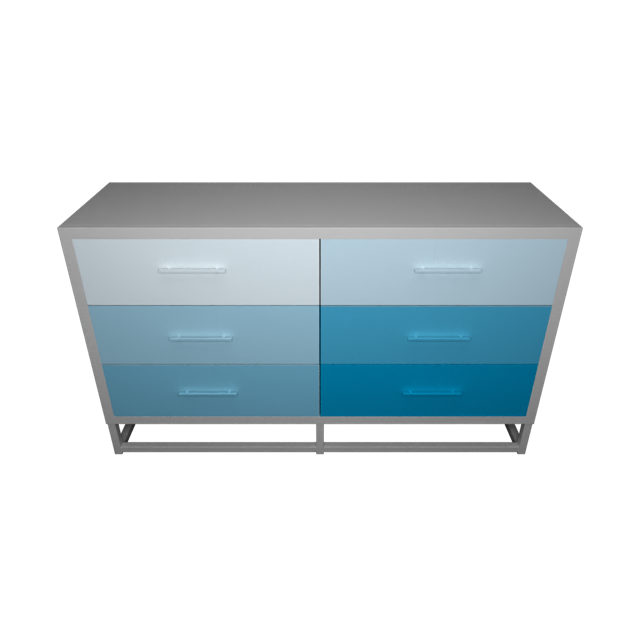} &
\imgclip{0}{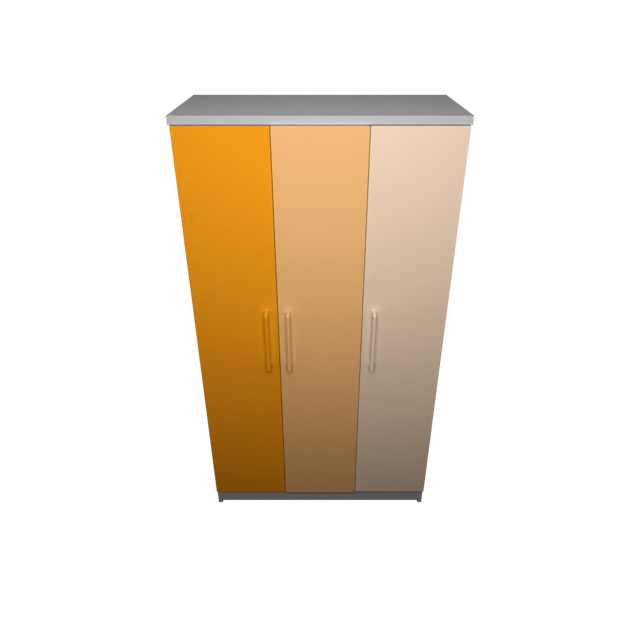} &
\imgclip{0}{figure/blender_supp/seg/fpngroupmot_w3df/6ef98b7750ff80c5225b0f7c733f23d196998038} &
\imgclip{0}{figure/blender_supp/seg/fpngroupmot_w3df/112dc87e26450400941c6eaff60866bc19badc64} &
\imgclip{0}{figure/blender_supp/seg/fpngroupmot_w3df/d5ba163ba97f94c7aa4a4a625eb0547b8894e1a5} &
\imgclip{0}{figure/blender_supp/seg/fpngroupmot_w3df/807955fd4dcb59b67789b24f2e7bc167027c870a} \\

& a & b & c & d & e & f & g & h \\

\bottomrule
\end{tabularx}
\caption{
Comparison of our methods trained on \pmopenshort and \pmopenshort + \acd-train ($^{\dagger}$) variants.  We show drawers in blue, doors in orange, and lids in green. As \pmopenshort contains interiors for drawers, models trained on it struggle to generalize on ACD as it lacks  interiors. WE notice that the models tend to be predicting doors instead of drawers (c, g, h). Adding \acd-train to the training split, helps both \ogroup and \gammagroup to both separate the instances and identify the drawers (c, g, h). We find that lids are still hard to segment as the data is scarce (a, e). Models also struggle to generalize on the objects that are large in the real-world, as after resizing the openable parts become smaller (g).
}
\label{fig:qual-results-pg-px-comp-supp}
\end{figure*}

\mypara{Results.}
Our \ogroup has considerably higher performance than the baseline without FPN, and the best overall performance on \pmopen (\Cref{tab:results-pc-seg}) with the highest mAP (69.3\%) and mAR (75.3\%) and lowest \occost.
As expected, training on the mixed \pmopen and \pmopenext data gives better performance on \ourdatashort, as \pmopenext provides a closer distribution to \ourdatashort.
At the same time, \ourdatashort remains a challenging benchmark due to more diverse and challenging shapes.
\ourdatashort shapes typically do not have complete interior geometry thus the useful signal of a drawer box attached behind the front is lost.
In fact, we see that point cloud based methods mistake drawers for doors on multiple occasions (see \cref{fig:qual-results-pg-px-comp-supp}). However, after some data exhibiting missing interiors is added, \ogroup and \gammagroup learn to generalize better at segmenting drawers without interiors.
Similarly, shelves behind the doors that are present in \pmopen are mostly missing in \ourdatashort which also explains lower performance on this part category. Additional challenges come from more diverse categories (i.e. BBQ grills) and large differences in real-world scale of the objects as, once resized during pre-processing, the openable parts become too small. 

\mypara{Breakdown by part-category.} In \cref{tab:results-pc-seg-category}, we examine the performance by part category.
Not surprisingly, the base part is relatively easy (with the highest performance).  There is a significant drop in performance for drawers and doors when going from \pmopen to \acd.

\subsubsection{Mesh segmentation}
\label{sec:supp-exp-mesh-seg}
\begin{table*}
\centering
\caption{Mesh segmentation of openable parts. We report the precision (P), recall (R), and F1 for openable parts (base part is excluded) at IoU=0.5. Note that a subset of macro-averaged metrics was reported in the main paper.
}
\vspace{-5pt}
{
\begin{tabular}{@{} ll rrr rrr rrr rrr @{}}
\toprule
 & &  \multicolumn{6}{c}{\pmopen} & \multicolumn{6}{c}{\ourdatashort} \\
\cmidrule(l{0pt}r{2pt}){3-8} \cmidrule(l{0pt}r{2pt}){9-14}
& & \multicolumn{3}{c}{Micro} & \multicolumn{3}{c}{Macro} & \multicolumn{3}{c}{Micro} & \multicolumn{3}{c}{Macro} \\
\cmidrule(l{0pt}r{2pt}){3-5} \cmidrule(l{0pt}r{2pt}){6-8} \cmidrule(l{0pt}r{2pt}){9-11} \cmidrule(l{0pt}r{2pt}){12-14}
Type & Method & P & R & F1 & P & R & F1 & P & R & F1 & P & R & F1 \\
\midrule
\multirow{7}{*}{PC} 
& \stomacro & 3.2 & 1.6 & 2.2 & 1.2 & 0.5 & 0.7 & 3.7 & 0.8 & 1.3 & 2.2 & 0.5 & 0.8\\
& \maskthreed & 60.0 & 42.9 & 50.0 & 52.6 & 33.9 & 41.1 & 15.5 & 4.9 & 7.5 & 19.2 & 5.0 & 9.2\\
& \ogroup & \textbf{90.8} & \textbf{81.3} & \textbf{85.8} & \textbf{92.2} & \textbf{83.0} & \textbf{87.2} & 42.3 & 11.3 & 17.8 & 30.0 & 7.0 & 11.2\\
& \ogroup$\!^{*}$ & 88.0 & 76.4 & 81.8 & 85.2 & 74.8 & 79.4 & \textbf{43.1} & \textbf{19.0} & \textbf{26.4} & \textbf{37.0} & \textbf{15.5} & \textbf{21.9}\\
& \gammagroup & 82.6 & 73.1 & 77.6 & 77.3 & 66.9 & 71.6 & 35.9 & 7.9 & 13.0 & 24.5 & 4.5 & 7.5\\
& \gammagroup$\!^{*}$ & 81.3 & 62.1 & 70.4 & 74.5 & 55.1 & 63.2 & 33.2 & 6.4 & 10.8 & 24.9 & 3.7 & 6.5\\
Mesh & \meshwalker & 1.7 & 1.6 & 1.7 & 1.0 & 1.0 & 1.0 & 1.1 & 0.7 & 0.8 & 1.2 & 0.7 & 0.9 \\
View & \opdformer & 0.3 & 0.5 & 0.4 & 0.7 & 1.2 & 0.9 & 1.6 & 1.0 & 1.2 & 1.8 & 1.0 & 1.3 \\
\bottomrule
\end{tabular}
}
\label{tab:results-mesh-precision-no-base}
\end{table*}
\begin{table*}
\centering
\caption{
Mesh segmentation results on \pmopen and \acd-val, including \ogroup and \gammagroup trained on a subset of \acd.
Training on \pmopenextshort+3DF helps \ogroup with performance on \pmopen, also providing an improvement on \acd-val. However, having \pmopenext additionally in the training split pushes the performance on \acd-val even further. \gammagroup, on the other hand, benefits the most from having only 3DF added to the training split, likely due to need to predict the motion parameters as well. We note that macro averages have been reported in the main paper.
}
\vspace{-5pt}
\resizebox{\linewidth}{!}
{
\begin{tabular}{@{} l cc r rrr rrr rrr rrr @{}}
\toprule
& & & & \multicolumn{6}{c}{\pmopen} & \multicolumn{6}{c}{\acd-val (HSSD + ABO)} \\
\cmidrule(l{0pt}r{2pt}){5-10} \cmidrule(l{0pt}r{2pt}){11-16}
& \multicolumn{2}{c}{Data}  & & \multicolumn{3}{c}{Micro} & \multicolumn{3}{c}{Macro} & \multicolumn{3}{c}{Micro} & \multicolumn{3}{c}{Macro} \\
\cmidrule(l{0pt}r{2pt}){2-3} 
\cmidrule(l{0pt}r{2pt}){5-7} \cmidrule(l{0pt}r{2pt}){8-10} \cmidrule(l{0pt}r{2pt}){11-13} \cmidrule(l{0pt}r{2pt}){14-16}
Method & \pmopenextshort & 3DF & Size & P & R & F1 & P & R & F1 & P & R & F1 & P & R & F1 \\
\midrule
\ogroup & \checkmark & \myxmark & 920 & 88.0 & 76.4 & 81.8 & 85.2 & 74.8 & 79.4 & 47.2 & 25.3 & 33.0 & 47.1 & 28.0 & 34.8\\
\ogroup & \myxmark & \checkmark & 645 & \textbf{91.8} & \textbf{86.3} & \textbf{89.0} & \textbf{89.8} & \textbf{85.8} & \textbf{87.6} & 63.5 & 47.3 & 54.2 & 66.6 & 52.8 & 58.8\\
\ogroup & \checkmark & \checkmark & 1105 & 84.9 & 77.5 & 81.0 & 75.4 & 69.2 & 71.8 & \textbf{65.8} & \textbf{49.0} & \textbf{56.1} & \textbf{75.3} & \textbf{58.5} & \textbf{65.7}\\
\gammagroup & \checkmark & \myxmark & 920 & 81.3 & 62.1 & 70.4 & 74.5 & 55.1 & 63.2 & 40.0 & 11.5 & 17.8 & 49.3 & 14.2 & 21.9\\
\gammagroup & \myxmark & \checkmark  & 645 & 82.2 & 68.7 & 74.9 & 72.4 & 60.4 & 65.6 & 58.7 & 41.6 & 48.7 & 62.6 & 44.2 & 51.8\\
\gammagroup & \checkmark & \checkmark & 1105 & 86.7 & 71.4 & 78.3 & 79.3 & 64.7 & 71.1 & 47.0 & 29.9 & 36.6 & 55.0 & 35.3 & 42.9\\
\bottomrule
\end{tabular}
}
\label{tab:results-mesh-precision-no-base-acd-train-supp}
\end{table*}

Standard mesh segmentation metrics typically measure the overall class accuracy of the segmentation~\cite{lahav2020meshwalker}.
As the base part dominates the mesh surface (i.e. most of the surface is not an openable part), it is possible to achieve a high accuracy by only predicting the base.  
Thus in the main paper, we reported the precision (P), recall (R), and F1 metrics  macro-averaged over object instances, measuring how well openable parts are identified at the object level.  Here, we report the P/R/F1 micro-averaged over part instances as well as additional accuracy based mesh metrics used in prior work. 

\mypara{Full results with part-level averages}
In \cref{tab:results-mesh-precision-no-base} and \cref{tab:results-mesh-precision-no-base-acd-train-supp}, we report the P/R/F1 micro-averaged over part instances.  We see the micro-average P/R/F1 tend to trend with the macro-averages, and that the methods follow the same ordering in performance. \textit{Comparison of predicting on point-cloud vs image vs mesh.} In our experiments, we find point-cloud based methods to be the most effective.  The mesh-based method (\meshwalker) cannot reliably identify openable parts.  This is likely because these openable parts do not have geometry that protrudes from the object base body to provide a signal that they can be opened (i.e. door handles are often not represented in enough geometric detail).  In this work, we only considered one view-based method, \opdformer. While it did not do as well as the point cloud methods, a stronger base model (e.g. SAM~\cite{kirillov2023segment}) and improved aggregation schemes may improve the performance of view-based models.

\mypara{Additional metrics.} 
In \cref{tab:results-mesh-seg-with-base} we report mesh segmentation metrics following prior work in mesh segmentation~\cite{kalogerakis2010learning} that focuses on the area of the shape that is semantically accurately labeled and compares the precise instance segmentation to the ground-truth.
However, we find the original naming\footnote{Metrics were originally named `Classification Error' and `Segment-Weighted Error'} to be misleading and therefore propose the following names: Classification Accuracy (\textbf{CA}) for Equation 12 from the paper and Normalized Classification Accuracy (\textbf{NCA}) for Equation 13.
The metrics are defined as below:
\[
\text{CA} = \frac{\sum_i \indicator[c_i, \hat{c}_{i}] a_i}{\sum_i a_i}, \quad
\text{NCA} = \sum_i \indicator[c_i, \hat{c}_{i}] \frac{a_i}{A_{c_i}}
\] where $a_i$ is the area of the $i$th face, $c_i$ is the ground truth and $\hat{c}_i$ the predicted label, $\indicator(c, c')$ is the standard indicator function which equals to 1 if the two labels match and 0 otherwise; $A_{c_i}$ in the NCA formulation is the total area of a segment with ground truth label $c_i$.
NCA is effectively the average of the per-label accuracy while CA is the face-area weighted segmentation accuracy.
As both NCA and CA are semantic accuracies (e.g. do not account for instances), we follow \citet{kalogerakis2010learning} and also report the Adjusted Rand Index (\textbf{ARI}) which measures agreement between the ground truth and predicted instance segmentations (adjusted for chance groupings).

As the CA and NCA metrics take the base part into account, the accuracy numbers may be inflated since the base part constitutes the largest part of the object.
For instance, even when no openable parts are predicted, the CA and NCA metrics will still be high due to high accuracy on predicting the base.
Thus, we introduce \CAnobase, a variant of CA which only considers openable parts (no base).
\CAnobase\xspace aggregates matching predictions over the union of GT openable parts and predicted openable parts only. This way, both cases of over- and under-segmentation are considered.

\begin{table*}
\centering
\caption{Evaluation of mesh segmentation following metrics from \citet{kalogerakis2010learning}.
CA is the face-area weighted semantic accuracy, NCA is the average per-label accuracy, and \CAnobase is a variant of CA that excludes the base part.
The Adjusted Rand Index (ARI) measures the quality of the instance segmentation.
For CA and NCA, the classification accuracy is dominated by the base part, these results overestimate the ability of the different methods to detect and identify the openable parts.
As P/R/F1 metrics from the main paper use matches based on ixed area threshold, they do not assess how much of the area gets a correct label assigned. We note that \ogroup $^{*}$ is dominant on \pmopen with these metrics rather than expected \ogroup. This might indicate that \ogroup $^{*}$ produces more partial segmentations less than the area threshold.
On \acd, we see that \ogroup $^{*}$ is dominant by quite a large margin across all metrics. Overall, results deteriorate significantly due to the dataset complexity.
The * suffix denotes training on a mix of \pmopen and \pmopenext.
}
{
\begin{tabular}{@{} ll rrrr rrrr @{}}
\toprule
& &  \multicolumn{4}{c}{\pmopen} & \multicolumn{4}{c}{\ourdatashort}\\
\cmidrule(l{0pt}r{2pt}){3-6} \cmidrule(l{0pt}r{2pt}){7-10}
Type & Method &  CA$\uparrow$ & NCA$\uparrow$ & ARI$\uparrow$ & 
\CAnobase$\uparrow$ &
CA$\uparrow$ & NCA$\uparrow$ & ARI$\uparrow$ &
\CAnobase $\uparrow$ \\ 
\midrule

\multirow{9}{*}{PC} & \pg + \unet & 91.7 & 87.1  & 0.473 & 69.7  & 81.4 & 53.2 & 0.143 & 19.3 \\
& \pg + \swinthreed & 94.0 & 90.7  & 0.549 & 77.1 & 84.7 & 59.4 & 0.242 & 30.3 \\
& \pg + \pointnextshort & 95.2 & 91.3  & 0.511 & 81.3 & 80.7 & 52.0 & 0.112 & 17.3 \\
& \ogroup & \textbf{96.9} & 93.4 & 0.617 & 85.8 & 84.2 & 54.1 & 0.136 & 18.7 \\
& \ogroup$\!^{*}$ & 96.5 & \textbf{94.4} & \textbf{0.754} & \textbf{86.4} & \textbf{87.7} & \textbf{67.1} & \textbf{0.358} & \textbf{41.4} \\
& \gammagroup & 94.4 & 89.2 & 0.566 & 77.6 & 83.0 & 51.2 & 0.096 & 12.3 \\
& \gammagroup$\!^{*}$ & 94.2 & 86.9 & 0.574 & 72.9 & 84.3 & 53.7 & 0.133 & 16.9 \\
& \maskthreed & 88.3 & 75.3  & 0.360 & 51.4 & 81.9 & 52.9 & 0.150 & 17.6 \\
& \stomacro & 78.4 & 55.3 & 0.030 & 13.7 & 82.5 & 46.0 & 0.008 & 3.8 \\
\midrule
Mesh & \meshwalker & 65.8 & 52.6  & 0.185 & 17.5 & 72.4 & 44.3 & 0.044 & 6.7 \\
\midrule
View & \opdformer & 75.6 & 56.5  & 0.058 & 17.8 & 77.0 & 54.7 & 0.070 & 21.0  \\
\bottomrule
\end{tabular}
}
\label{tab:results-mesh-seg-with-base}
\end{table*}

\mypara{Results.}
\Cref{tab:results-mesh-seg-with-base} shows results of mesh segmentation predictions, suggesting that point cloud-based methods outperform all other methods.
We note that as base parts are typically the biggest parts in all of the objects, they significantly influence the metric results.
The performance based on CA and NCA are significantly higher than when we focus on just the openable parts.
We see with \CAnobase, the impact of the base part is limited.
But as CA, NCA, and \CAnobase\xspace do not account for instance segmentation, we also report ARI (a perfect segmentation will achieve an ARI of 1).
We find that compared to our metrics (precision, recall, F1 over detected parts), these metrics from prior work are less interpretable (ARI) or do not properly evaluate the instance segmentation and are too heavily influenced by the base part (CA, NCA).

\mypara{Discussion.}  Despite these flaws, results from these metrics show mostly the same trend as we report in the main paper, with the \ogroup variants being the top performers for point-cloud methods and a large drop in performance as we go from \pmopen to \ourdatashort, showing the challenge of \ourdatashort. As well as \stom being the worst-performing PC-based method.
We see that \opdformer under-performs compared to other methods (except for \meshwalker).
This is likely due to overprediction by \opdformer and these metrics being area-weighted vs considering detected instances.
The overprediction is due to the fact that there can be multiple predictions from different viewpoints that may predict either the same or different part instances.
While our heuristic for dealing with overlapping masks tries to solve some of these issues, there might still be some ``leftover'' masks after splitting.
The ARI metric gives a good measure of how well the predicted segmentation matches the ground truth (\opdformer also performs poorly here), but it is less easy to determine why the match is poor, while the separation into precision and recall allows us to determine that \opdformer is overpredicting.

We also find that \meshwalker under-performs significantly compared to other methods.
We hypothesize this is due to \pmopen and our mesh data in general having highly non-uniform vertex distribution with curved parts such as handles being far denser than flat surfaces, severely impacting the random walks on which the method is based.

In terms of performance on \ourdatashort, while CA may be not that susceptible to the errors, NCA and ARI show significant performance degradation.

\begin{table}[t]
\centering
\caption{Motion prediction precision and recall on \pmopen. \ogroup with our heuristic works the best. When comparing \gammagroup and heuristic prediction given the same (\gammagroup) segmentation, we find precision and recall comparable for M and MA but our heuristic clearly outperforming when it comes to MAO. Other baselines perform poorly.
}
\vspace{-5pt}
\resizebox{\linewidth}{!}
{
\begin{tabular}{@{} llc rrr rrr @{}}
\toprule 
& & & \multicolumn{3}{c}{Precision \% $\uparrow$} & \multicolumn{3}{c}{Recall \% $\uparrow$} \\
\cmidrule(l{0pt}r{2pt}){4-6} 
\cmidrule(l{0pt}r{2pt}){7-9}
Motion & Segmentation  & \# & {+}M & {+}MA & {+}MAO & {+}M & {+}MA & {+}MAO \\
\midrule
Learned & \stom & 3 & 1.1 & 1.1 & 1.1 & 0.8 & 0.8 & 0.8 \\
Learned & \gammagroup & 133 & 71.1 & 61.7 & 50.7 & 71.3 & 62.2 & 50.5 \\
Heur & \maskthreed & 78 & 42.4 & 37.5 & 30.5 & 36.4 & 31.2 & 24.5 \\
Heur  & \gammagroup & 133 & 71.0 & 62.4 & 58.0 & 71.3 & 61.7 & 57.0\\
Heur & \ogroup & \textbf{148} & \textbf{79.6} & \textbf{75.4} & \textbf{60.9} & \textbf{78.4} & \textbf{70.3} & \textbf{59.1} \\
\midrule
Heur & GT  & 182 & 94.7 & 88.4 & 84.5 & 94.7 & 88.4 & 84.5\\
\bottomrule
\end{tabular}
}
\vspace{-4mm}
\label{tab:results-motion-main}
\end{table}

\begin{table}[t]
\centering
\caption{Motion prediction precision and recall on \acd.
Performance overall is quite low, as segmentation remains being the bottleneck. We find \ogroup, providing the best segmentation, with our heuristic is the best option in the setup with no additional training. 
}
\vspace{-5pt}
\resizebox{\linewidth}{!}
{
\begin{tabular}{@{} llc rrr rrr @{}}
\toprule 
& & & \multicolumn{3}{c}{Precision \% $\uparrow$} & \multicolumn{3}{c}{Recall \% $\uparrow$}\\
\cmidrule(l{0pt}r{2pt}){4-6} 
\cmidrule(l{0pt}r{2pt}){7-9}
Motion & Segmentation  & \# & {+}M & {+}MA & {+}MAO & {+}M & {+}MA & {+}MAO\\
\midrule
Learned & \stom & 10 & 1.1 & 1.1 & 0.9 & 1.2 & 1.2 & 0.8 \\
Learned & \gammagroup & 106 & 13.6 & 11.9 & 6.4 & 0.9 & 0.8 & 4.3 \\
Heur & \maskthreed & 66 & 8.4 & 7.8 & 3.8 & 5.9 & 5.0 & 2.8 \\
Heur  & \gammagroup & 106 & 13.8 & 9.3 & 7.2 & 9.1 & 6.6 & 5.1 \\
Heur & \ogroup & \textbf{151} & \textbf{21.3} & \textbf{14.2} & \textbf{9.5} & \textbf{15.2} & \textbf{9.8} & \textbf{6.4} \\
\midrule
Heur & GT & 1350 & 94.9 & 81.0 & 65.3 & 94.9 & 81.0 & 65.3  \\
\bottomrule
\end{tabular}
}
\vspace{-2mm}
\label{tab:results-motion-main-hssd}
\end{table}

\begin{table}[t]
\centering
\caption{Motion prediction precision and recall on \acd-val. Overall, our heuristic dominates in terms of precision while recall is comparable when given the same \gammagroup segmentation. With \ogroup segmentation, heuristic-based approach is clearly dominant. We note that there is still quite a bit of room until heuristic reaches its upper-bound, meaning that stronger segmentaiton is required. 
}
\vspace{-5pt}
\resizebox{\linewidth}{!}
{
\begin{tabular}{@{} llc rrr rrr @{}}
\toprule 
& & & \multicolumn{3}{c}{Precision \% $\uparrow$} & \multicolumn{3}{c}{Recall \% $\uparrow$} \\
\cmidrule(l{0pt}r{2pt}){4-6} 
\cmidrule(l{0pt}r{2pt}){7-9}
Motion & Segmentation  & \# & {+}M & {+}MA & {+}MAO & {+}M & {+}MA & {+}MAO\\
\midrule
Learned & \gammagroup $^{\dagger}$  & 225 & 44.4 & 37.4 & 26.3 & 39.7 & 33.4 & 25.1 \\
Heur  & \gammagroup $^{\dagger}$ & 225 & 44.8 &  37.3 & 30.3 & 40.3 & 33.5 & 27.7 \\
Heur & \ogroup $^{ * \dagger}$ & \textbf{265} & \textbf{58.4} & \textbf{46.1} & \textbf{36.9} & \textbf{53.0} & \textbf{40.8} & \textbf{32.1} \\
\midrule
Heur & GT & 541 & 96.7 & 82.4 & 76.9 & 96.7 & 82.4 & 76.9\\
\bottomrule
\end{tabular}
}
\vspace{-2mm}
\label{tab:results-motion-main-acdtrain}
\end{table}

\begin{table}[t]
\centering
\caption{Motion prediction evaluation on \ourdatashort-val. We select the best checkpoints trained on additional data and benchmark learned and heuristic-based motion prediction. We find that \ogroup with heuristic outperforms \gammagroup.
}
\vspace{-5pt}
\resizebox{\linewidth}{!}
{
\begin{tabular}{@{} llc rrr rr @{}}
\toprule 
& & & \multicolumn{3}{c}{F1 \% $\uparrow$} & \multicolumn{2}{c}{Error $\downarrow$} \\
\cmidrule(l{0pt}r{2pt}){4-6} 
\cmidrule(l{0pt}r{2pt}){7-8}
Motion & Segmentation  & \# & {+}M & {+}MA & {+}MAO & AE & OE \\
\midrule
Learned & \gammagroup $^{\dagger}$ & 225 & 41.9 & 35.3 & 25.7 & \textbf{11.2} & 0.40 \\
Heur  & \gammagroup $^{\dagger}$ & 225 & 42.4 & 35.3 & 29.0 & 11.6 & \textbf{0.27} \\
Heur & \ogroup $^{* \dagger}$ & \textbf{265} & \textbf{55.6} & \textbf{43.3} & \textbf{34.3} & 15.1 & 0.30 \\
\midrule
Heur & GT & 541 & 96.7 & 82.4 & 76.9 & 13.6 & 0.18 \\
\bottomrule
\end{tabular}
}
\vspace{-2mm}
\label{tab:results-motion-main-acdtrain-f1}
\end{table}

\subsection{Motion Prediction}
\Cref{tab:results-motion-main,tab:results-motion-main-hssd,tab:results-motion-main-acdtrain} present precision and recall values for motion prediction, in addition to the F1 scores reported in the main paper. In these experiments, \pmopen was used in training by default.  We use $^*$ as an indication that the model was trained using both \pmopenext and $^\dagger$ to indicate that \acd-train was also used for training. 

\begin{table}[t]
\centering
\caption{
Interior and object reconstruction evaluation.
Given a shape from \pmopenext as an input, we evaluate against shapes from \pmopen.  We note that our pipeline allows to preserve the original geometry. We report the chamfer distance over all parts (CD), and F1 scores for drawers. 
We note that SINGAPO has seen ~83\% of these shapes during training not accounting for augmented data.
}
\label{tab:chamfer-supp}
\vspace{-5pt}
{
\begin{tabular}{@{} l rrrr @{}}
\toprule
Method & CD & F1\\
\midrule
\ogroup $^{* \dagger}$ + Heuristic & \textbf{1.7} & 37.3 \\
SINGAPO  & 3.3 & \textbf{69.8} \\
URDFormer  & 23.4 & 0.0 \\
\midrule
GT + Heuristic & 1.6 & 73.9 \\
\bottomrule
\end{tabular}
}
\vspace{-4mm}
\end{table}

We further select the best checkpoints for \ogroup and \gammagroup trained on additional data, according to their segmentation performance on \acd-val as seen in \cref{tab:results-mesh-precision-no-base-acdtrain} and study the performance on \acd-val. Overall, the metrics are noticeably improved but we still find that the same trends hold as in evaluation on full \acd. There is a lot of room until heuristic hits its upper-bound as seen in GT segmentation row. We find that segmentation is the biggest challenge of the pipeline and bottlenecks downstream motion prediction and interior completion.

\subsection{Interior completion}
For quantitative evaluation, we leverage the correspondence between \pmopenext and \pmopen: applying interior completion on \pmopenext shapes, and treating \pmopen as ground truth.

\mypara{Metrics.}
To evaluate interior completion, we compare the predicted shape with the ground-truth shape using chamfer distance (CD) across all parts based on 20K sampled points to also take in account how the baselines preserve the original geometry. As we do not target exact match of drawer interiors but rather having a plausible completion we compute macro F1 score for drawers based on bounding box matching. We select a high threshold of 0.8 for matching so that the drawers are sufficiently close to ground truth but allowing for minor variations.

\mypara{Results.} We evaluate interior completion on our validation split. We filter the categories unseen by SINGAPO (Safe, Trashcan), leaving 81 shapes total. We note that SINGAPO's training split, not considering the augmented data, has 67 overlapping shapes with our validation split. The results in \cref{tab:chamfer-supp} shows lower CD value for our approach, meaning that it excels at preserving the original geometry. In terms of drawer F1 score, we find that segmentation remains being the bottleneck and SINGAPO performs better. With the ground truth segmentation, however, our heuristic dominates all the metrics.

\begin{figure*}
\centering
\includegraphics[width=\linewidth]{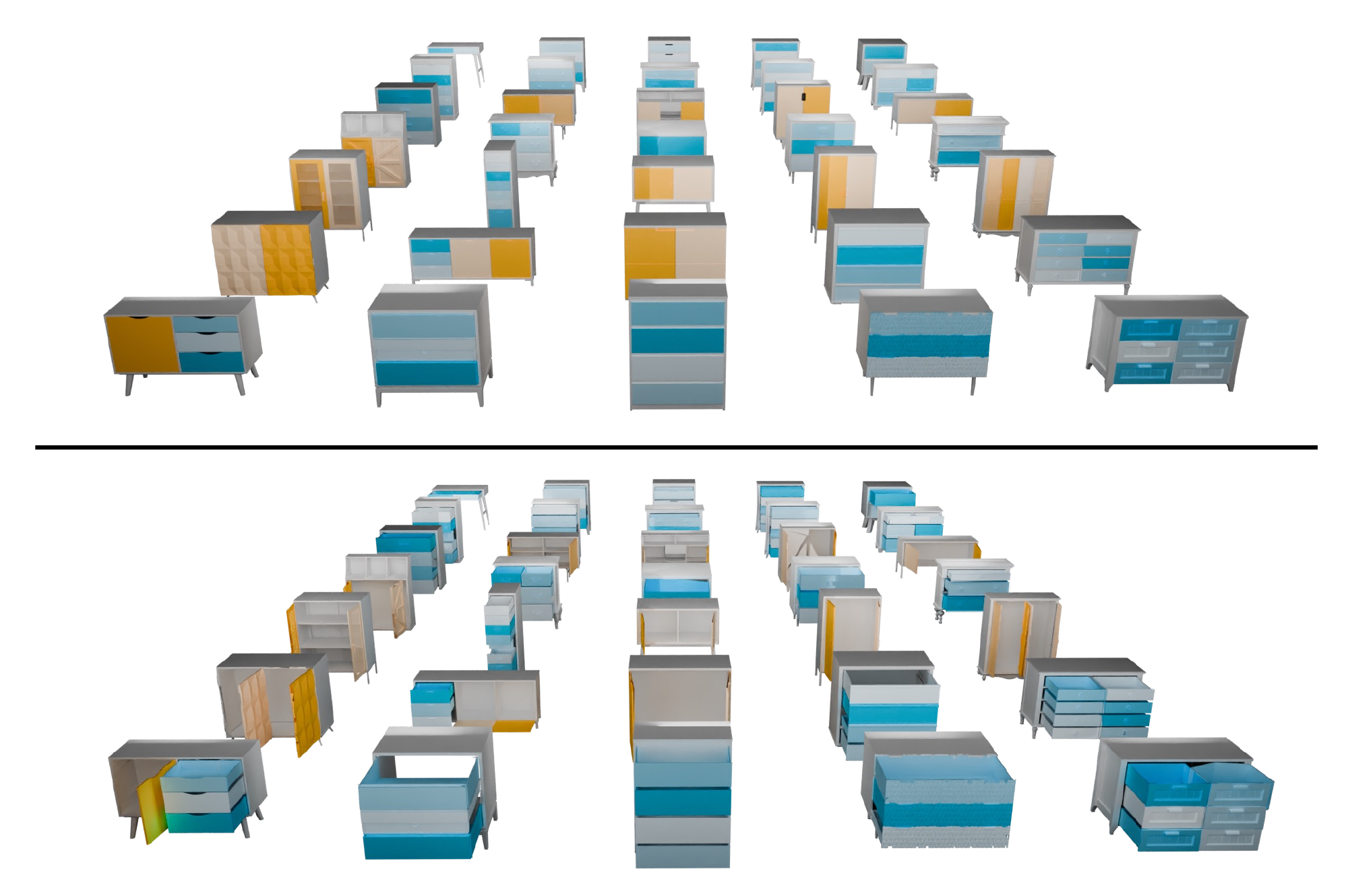}
\vspace{-2pt}
\caption{
Compilation of results after applying our pipeline on 3D-FUTURE~\cite{fu20213d} and ABO~\cite{collins2022abo} shapes, outside of annotated \ourdatashort subset.
The top shows the openable part segmentations with doors in shades of orange and drawers in shades of blue.
The bottom shows the same objects with their openable parts articulated to a more open state.
} 
\label{fig:ood-supp}
\end{figure*}

\subsection{Application}
We find that the improvements introduced by training on \acd-train and \pmopenext allow more reliably generalizing to more complex shapes. This enables automated application of S2O pipeline on shapes not included in \ourdatashort. For this matter, we apply it on some shapes from 3D-FUTURE and ABO datasets which were not previously annotated as part of \acd effort. We find that our pipeline is able to make such shapes articulated, as shown in \cref{fig:ood-supp}. This enables for automated and more efficient shape annotation. See supplemental videos for more detailed overview of examples.
\clearpage

\clearpage

\end{document}